\documentclass[journal]{IEEEtran}
\let\labelindent\relax
\usepackage{graphicx}
\usepackage{subcaption}
\usepackage[font=small]{caption}
\usepackage{eqlist}
\usepackage{amsfonts}
\usepackage{tabularx}
\usepackage{booktabs}
\usepackage{amssymb}
\usepackage{amsmath}
\usepackage{enumitem}
\usepackage{threeparttable}
\usepackage[vlined, ruled, linesnumbered, commentsnumbered]{algorithm2e}
\usepackage{lipsum}

\usepackage[usenames,dvipsnames]{xcolor}
\usepackage{comment}
\usepackage{flushend}
\usepackage{multirow}

\usepackage{empheq}

\begin{document}
\title{Planning a Sequence of Base Positions for \\ a Mobile Manipulator to Perform \\ Multiple Pick-and-Place Tasks}

\author{Jingren~Xu,~\IEEEmembership{Student~Member,~IEEE,}
        Yukiyasu~Domae,~\IEEEmembership{Member,~IEEE,}
        Toshio~Ueshiba,~\IEEEmembership{Member,~IEEE,}
        Weiwei~Wan,~\IEEEmembership{Member,~IEEE,}
        and~Kensuke~Harada,~\IEEEmembership{Senior~Member,~IEEE}
\thanks{J. Xu, W. Wan and K. Harada are with Graduate School of Engineering Science, Osaka University, Osaka 560-8531, Japan. J. Xu and K. Harada are also with Artificial Intelligence Research Center, National Institute of Advanced Industrial Science and Technology (AIST), Japan (e-mail: xu@hlab.sys.es.osaka-u.ac.jp; wan@sys.es.osaka-u.ac.jp; harada@sys.es.osaka-u.ac.jp).}
\thanks{Y. Domae and T. Ueshiba are with Artificial Intelligence Research Center, National Institute of Advanced Industrial Science and Technology (AIST), Japan (e-mail: domae.yukiyasu@aist.go.jp).}
\thanks{Manuscript received Month 00, 0000.}}

\markboth{IEEE Transactions on Automation Science and Engineering. Submission for Review, 2020}
{xu \MakeLowercase{\textit{et al.}}: Planning a Sequence of Base Positions for a Mobile Manipulator to Perform Multiple Pick-and-Place Tasks}
\maketitle

\begin{abstract}

In this paper, we present a planner that plans a sequence of base positions for a mobile manipulator to efficiently and robustly collect objects stored in distinct trays. We achieve high efficiency by exploring the common areas where a mobile manipulator can grasp objects stored in multiple trays simultaneously and move the mobile manipulator to the common areas to reduce the time needed for moving the mobile base. We ensure robustness by optimizing the base position with the best clearance to positioning uncertainty so that a mobile manipulator can complete the task even if there is a certain deviation from the planned base positions. Besides, considering different styles of object placement in the tray, we analyze feasible schemes for dynamically updating the base positions based on either the remaining objects or the target objects to be picked in one round of the tasks. In the experiment part, we examine our planner on various scenarios, including different object placement: (1) Regularly placed toy objects; (2) Randomly placed industrial parts; and different schemes for online execution: (1) Apply globally static base positions; (2) Dynamically update the base positions. The experiment results demonstrate the efficiency, robustness and feasibility of the proposed method.

\end{abstract}
\def\abstractname{Note to Practitioners}
\begin{abstract}

The presented project uses mobile manipulators to fetch and supply parts in an automotive assembly factory. Mechanical parts at these sites are usually regularly or randomly placed in supply trays. The project develops methods to explore robust mobile base positions where the objects with different placement styles and from different trays can be reached by a mobile manipulator. The mobile manipulators could perform efficient and high-quality pick-and-place operations by navigating to the explored positions. The developed methods also discuss the dynamic updates the base positions following picking process to further increase system robustness. The presented project is expected to shed light on the deployment of mobile manipulators to collect parts at large manufacturing factories.

\end{abstract}


%
\IEEEpeerreviewmaketitle

\section{Introduction}

\begin{figure}
    \centering
    \includegraphics[width=\columnwidth]{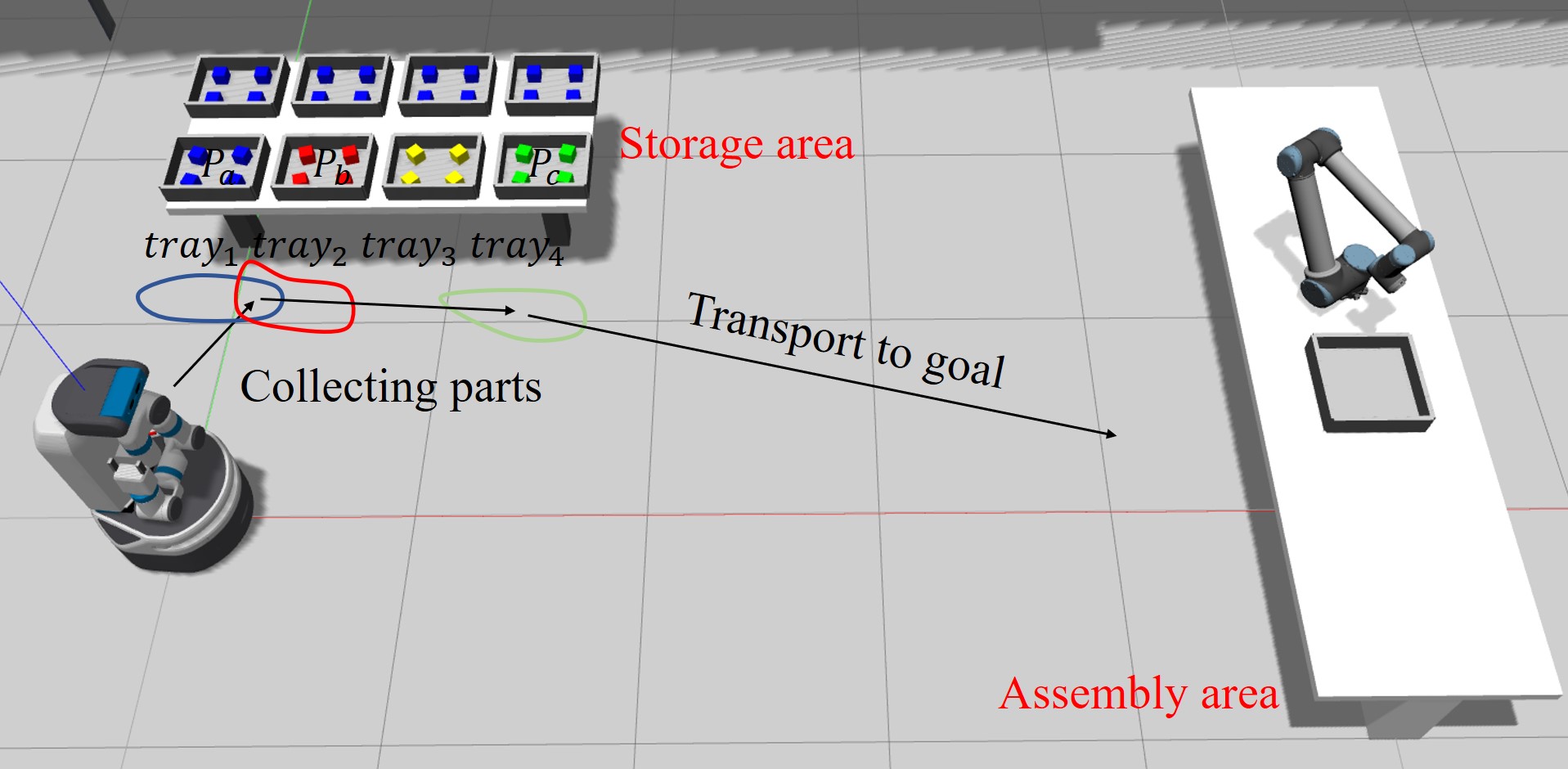}
    \caption{A schematic diagram of the storage area in an assembly factory, a mobile manipulator is used to replace human workers to pick up objects from multiple places and then transport to the assembly line for further assembly.}
    \label{fig:assembly_factory}
\end{figure}

\IEEEPARstart{T}{here} is an increasing demand on robots which are able to flexibly assist human in daily work and industrial production. A mobile manipulator, combining a mobile base and a manipulator, is able to navigate and manipulate in human environment. Therefore, mobile manipulators are promising for serving human environment and performing a variety of tasks in a large work space. Especially, the mobile manipulator is suitable for picking up objects and transporting them to the desired position. One example of such multiple pick-and-place tasks is the part-supply task in manufacturing environment, such as in an automotive assembly factory, where a huge amount of assembly components of an automobile are scattered in a large storage area (Fig. \ref{fig:assembly_factory}). Currently human workers are still highly engaged in the part-supply task and they have to go back and forth to fetch required assembly components from trays in different locations, and then carry the collected components to the assembly area. It is promising to apply mobile manipulators to replace human workers and automate this part-supply process, owning to their ability of moving and picking. 

In order to assemble a product {\it P}, which is comprised of several types of parts {\it $P_a$}, {\it $P_b$} and {\it $P_c$}, stored in different trays {\it $tray_1$}, {\it $tray_2$} and {\it $tray_4$}, respectively, the mobile manipulator may have to move to a sequence of positions to gradually pick up the required assembly parts from different trays. The base sequence are comprised of a series of base positions, in which the mobile manipulator is able to grasp the required assembly parts from the trays without self-collision and the collision with the environment. To collect the required parts from the target trays, we can simply move the mobile manipulator to the position in front of every target tray, alternatively, we can move the mobile manipulator to the position where the mobile manipulator can reach the parts stored in multiple trays, and then the mobile manipulator carries the collected assembly parts to the designated area for further assembly.

However, with increased base sequence size, the overall operation time increases significantly for the following reasons: (1) The mobile manipulator decelerates and then accelerates at every base position of the base sequence, which lowers the overall base velocity and increases the total operation time. (2) Every time the mobile manipulator experiences a "stop and manipulation", there is risk that the arrived position significantly deviates from the desired position, then the mobile manipulator has to perform time-consuming repositioning process, and the risk accumulates with expanding base sequence. Therefore, it is crucial to prune out unnecessary base movements to improve the overall efficiency. 

To simplify the base sequence for the general multiple pick-and-place tasks, it is preferable to move the mobile manipulator to the position, in which the mobile manipulator is able to pick up the assembly parts from multiple trays. As shown in Fig. \ref{fig:assembly_factory}, the mobile manipulator moves to the first base position and is able to grasp the assembly parts in both $tray_1$ and $tray_2$, thus reduces the base sequence size by one. To obtain such positions, we first calculate the base region for every target tray in the given assembly task, the base region is a set of base positions where the mobile manipulator can grasp the \textbf{target} assembly parts from that tray without collision. Then the intersections of these base regions are given high priority in the base sequence planning. Moreover, the base positioning uncertainty is considered in the base sequence planning by restricting the size of the applicable intersections, the intersections smaller than the base positioning uncertainty are discarded, otherwise the mobile manipulator is likely to move out the intersection due to the unavoidable positioning error. With the calculated base regions and robust intersections, we plan the shortest path that visit all the target trays with minimal number of base movements. Following the planned base sequence, the mobile manipulator can perform efficient and robust part-supply tasks in real world applications.

The base sequence planning can be implemented in multiple forms, depending on the definition of the "target" objects. Firstly, if the target objects are all the objects which fill up a tray, we obtain a globally static base sequence, which is always valid no matter how many objects remain in the tray, and all the base position planning can be performed offline. Secondly, the "target" objects can be the remaining objects in the tray, the base region may become larger when the mobile manipulator finishes one round of pick-and-place task, then the base region region can be updated using the remaining objects in the tray. Finally, the "target" objects can be a set of objects to be picked in a round of pick-and-place task, such that it is sufficient to calculate the base positions where the mobile manipulator can reach a specified amount of objects. In terms of the last two schemes where the base positions update, they improve the overall robustness, but the critical issue is that whether the update of base positions can be performed online, which is closed related to the pattern of object placement in the tray.

\section{Related Work}
Early work in mobile manipulation proposed rich theoretical analysis and concepts, which are extensively used as the evaluation metrics in later work. They were mainly concerned with the kinematics and dynamics modeling of mobile manipulator \cite{tchon2000, de2006kinematic}, resolution of the kinematic redundancy caused by the mobile base \cite{pin1990, white2009experimental}, coordinated control of mobile base and manipulator arm \cite{seraji1993}, and the manipulability analysis of the mobile manipulator \cite{yamamoto1999, bayle2003manipulability}.

More recently, there have been numerous work on motion planning of mobile manipulator \cite{berenson2008optimization, bodily2017motion, gochev2012planning} and optimal base position planning for achieving certain design criteria, such as manipulability, end effector poses and velocity \cite{ren2016method}, and reaching/grasping a set of targets \cite{burget2015stance, abolghasemi2016real}. 
Du et al. \cite{du2013} used the manipulability index to determine a suitable base placement. Ren et al. \cite{ren2016method} optimized the base positions for a mobile manipulator to reach a set of positions with required orientations and keep a stable velocity in local painting tasks. Berenson et al. \cite{berenson2008optimization} obtained the base placement and grasp for a mobile manipulator to move an object from one configuration to another, by optimizing a scoring function which combines the grasp quality, manipulability and distance to obstacle. OpenRAVE \cite{openrave} provides an inverse reachability module, which clusters the reachability space for a base-placement sampling distribution that can be used to find out where the robot should stand in order to perform a manipulation task. Stulp et al. \cite{stulp2012learning} proposed Action-Related Place to associate a base location with a probability of successfully performing a manipulation task, capability map was used to determine if an object was theoretically reachable. Burget et al. \cite{burget2015stance} employed the inverse reachability map to select statically stable, collision-free stance configurations for a humanoid robot to reach a given grasping target. Zacharias et al. \cite{zacharias2008positioning} took advantage of the reachability map to position a mobile manipulator to perform a linear trajectory in the workspace. Vahrenkamp \cite{vahrenkamp2013} conducted a series of research on reachability analysis and its application, the base positions with high probability of reaching a target pose can be efficiently found from the inverse reachability distribution. The reachability indicates the probability of finding an IK solution, while there is no guarantee on the completeness of obtained base positions. Some other researches that used reachability analysis are referred to \cite{leidner2013hybrid, dong2015orientation}.

Instead of planning where to place the base given the end-effector pose, there is a line of research that exploit the capability and reachability of a robot. Zacharias et al. \cite{zacharias2007capturing} developed capability map for a manipulator, using the capability map, the poses that are easy to reach can be deduced. Vahrenkamp et al. \cite{vahrenkamp2009, vahrenkamp2012} showed that the reachability space can be used to speed up the randomized IK solver and solve complex IK problems in cluttered environment. Jamone et al. \cite{jamone2014autonomous} also built up the reachable space map for a robot, but the map was described using motor representation. Malhan et al. \cite{malhan2019identifying} constructed the capability map to evaluate whether the waypoints on the surface of a workpiece satisfy position and singularity constraints.

In addition to the base placement for mobile manipulators, there is a line of research on the optimal placement for fixed-base manipulators \cite{feddema1996kinematically, hsu1999placing,zeghloul1993multi,mitsi2008determination}. Feddema et al. \cite{feddema1996kinematically} resolved the optimal position for a fixed-base manipulator to reach a set of points in the workspace, where no obstacle is assumed. Hsu. et al. \cite{hsu1999placing} considered the obstacles in the workspace, a randomized path planner and a fast path optimization routine were combined to iteratively search for the best base location. Regardless of where the manipulator is mounted, a mobile base or a fixed base, the optimization of base position shares many common criteria, such as manipulability and time-optimality of the trajectory.

Thakar et al. \cite{thakar2020} considered the manipulator motion planning problem for a mobile manipulator to pick and transport a target object, where the picking is performed while the base is moving. They considered accounting for the pose estimation of the target object in \cite{thakar2019accounting}, however, picking from a moving base is still relatively sensitive to the base position uncertainty and is not robust in real world implementation. Moreover, for multiple pick-and-place tasks considered in this paper, the mobile manipulator is expected to pick up multiple objects stored within a small region, it is very difficult for the mobile manipulator to finish the task within a short period of time, in this sense, picking while moving is not practical. Instead we employ the stop-and-pick scheme, the robot stops at the planned base positions and then picks up objects from target trays, our effort is on reducing the number of base movement.

Although the research on mobile manipulation is enormous, the problem of planning a sequence of base positions for grasping objects stored in multiple places has been hardly addressed. In this paper, we consider both the completeness of obtained base regions and the robustness with respect to base positioning uncertainty, and explore the intersections of the base regions which are used to reduce the operation time for general multiple pick-and-place tasks, the contributions are summarized as follows:
\begin{itemize}
    \item{A resolution complete method, based on precomputed reachability database, is proposed to approximate collision-free IK solutions, which is especially helpful in complex environment. And a resolution complete set of base positions can be obtained by such an IK approach. Our reachability database does not only count the number of poses in the voxel, which gives the probability of a pose being reachable, but also stores the manipulator configurations accessible to further IK queries.}
    \item{The base positioning uncertainty in real environment is considered in the base sequence planning. Following the planned base sequence, the mobile manipulator is able to efficiently and robustly collect the target parts even if there is certain deviation from the planned base positions.}
    \item{Both regularly and randomly placed objects in the tray are considered in calculating the base region, we propose a method to estimate the base region for randomly placed objects. Based on the object placement styles, we study the possible schemes for dynamically updating the base regions, which further increases the robustness. The efficiency, robustness and feasibility are comprehensively confirmed by experimental studies.}
\end{itemize}

Note that a part of this paper has been presented in \cite{xu2020planning}, it presented a special case of the current version, which planned a globally static base sequence for the mobile manipulator to collect all the objects in a tray. Different from the published conference paper, we in this journal version provide extended discussions on the implication of the result on designing tray configurations in the storage area. Moreover, we consider different object placement styles in the trays, either regularly or randomly, and discuss the possible schemes for dynamically updating the base positions as the pick-and-place tasks proceed. Significantly extended experimental results are presented to demonstrate the feasibility of our method in different scenarios, including different object placement styles and whether the base positions update.

\section{Method Overview}
\begin{figure}
    \centering
    \begin{subfigure}[h]{\columnwidth}
        \centering
        \includegraphics[width=\textwidth]{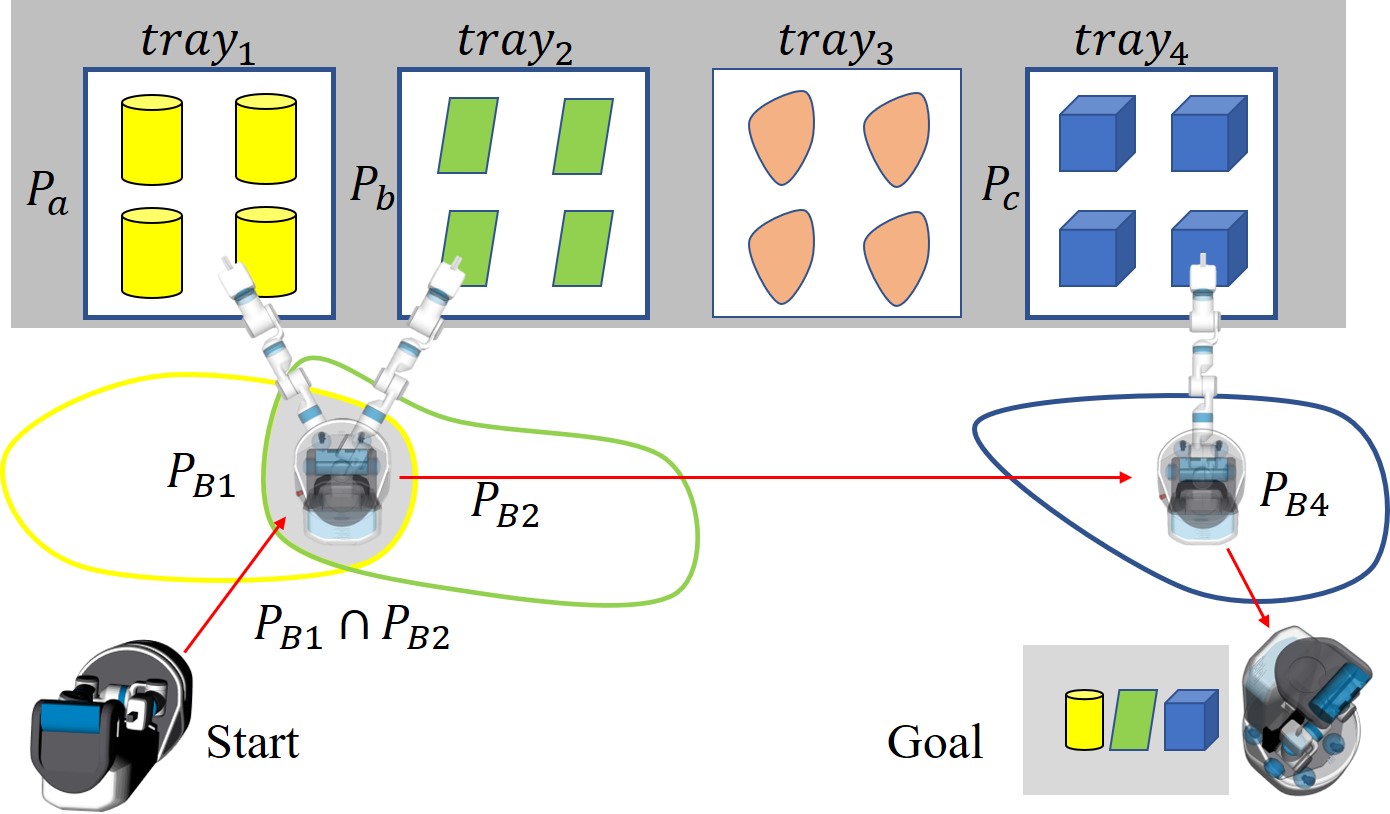}
        \caption{}
    \end{subfigure}
    \begin{subfigure}[h]{0.46\columnwidth}
        \centering
        \includegraphics[width=\textwidth]{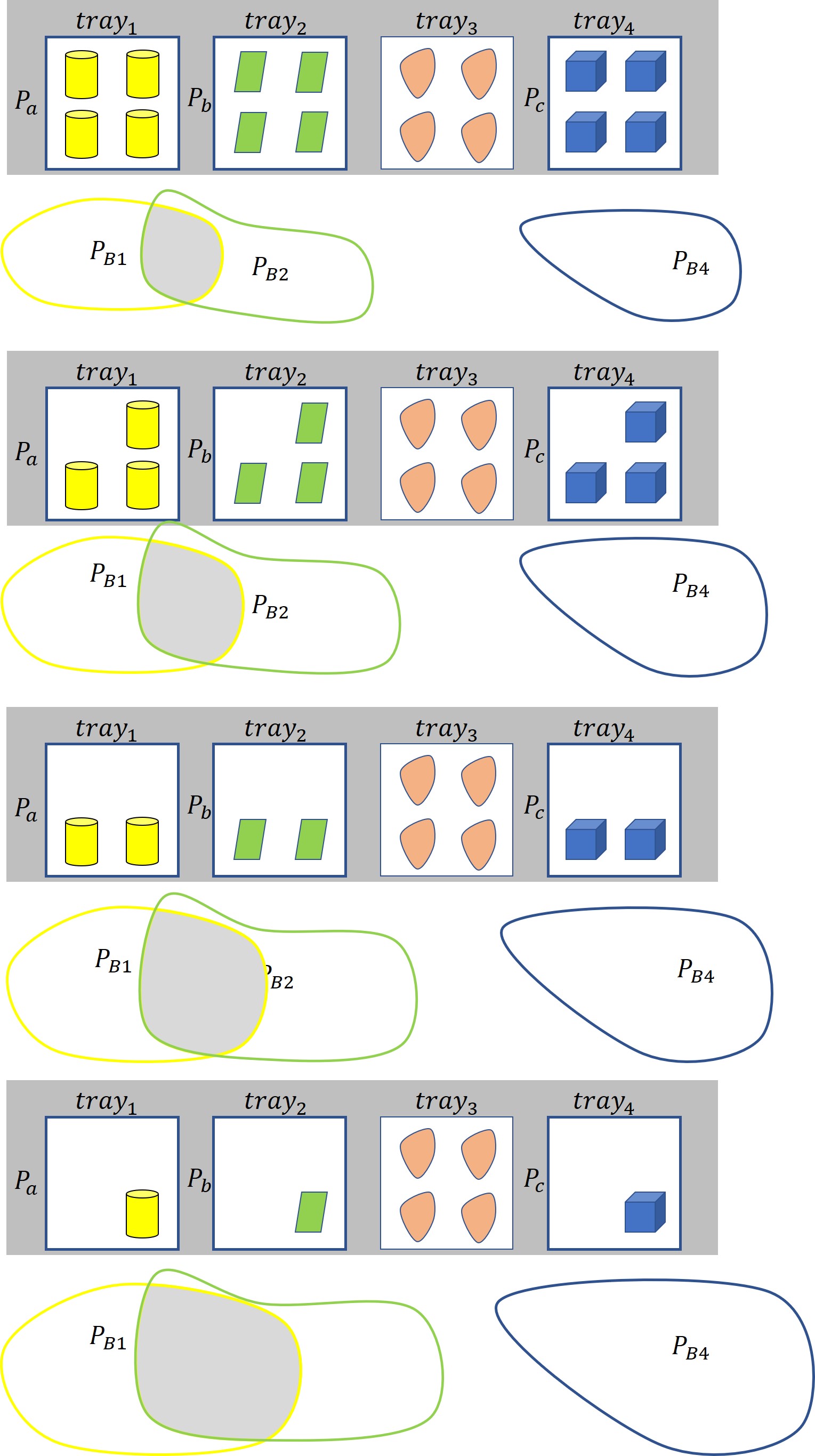}
        \caption{}
    \end{subfigure}
    \begin{subfigure}[h]{0.52\columnwidth}
        \centering
        \includegraphics[width=\textwidth]{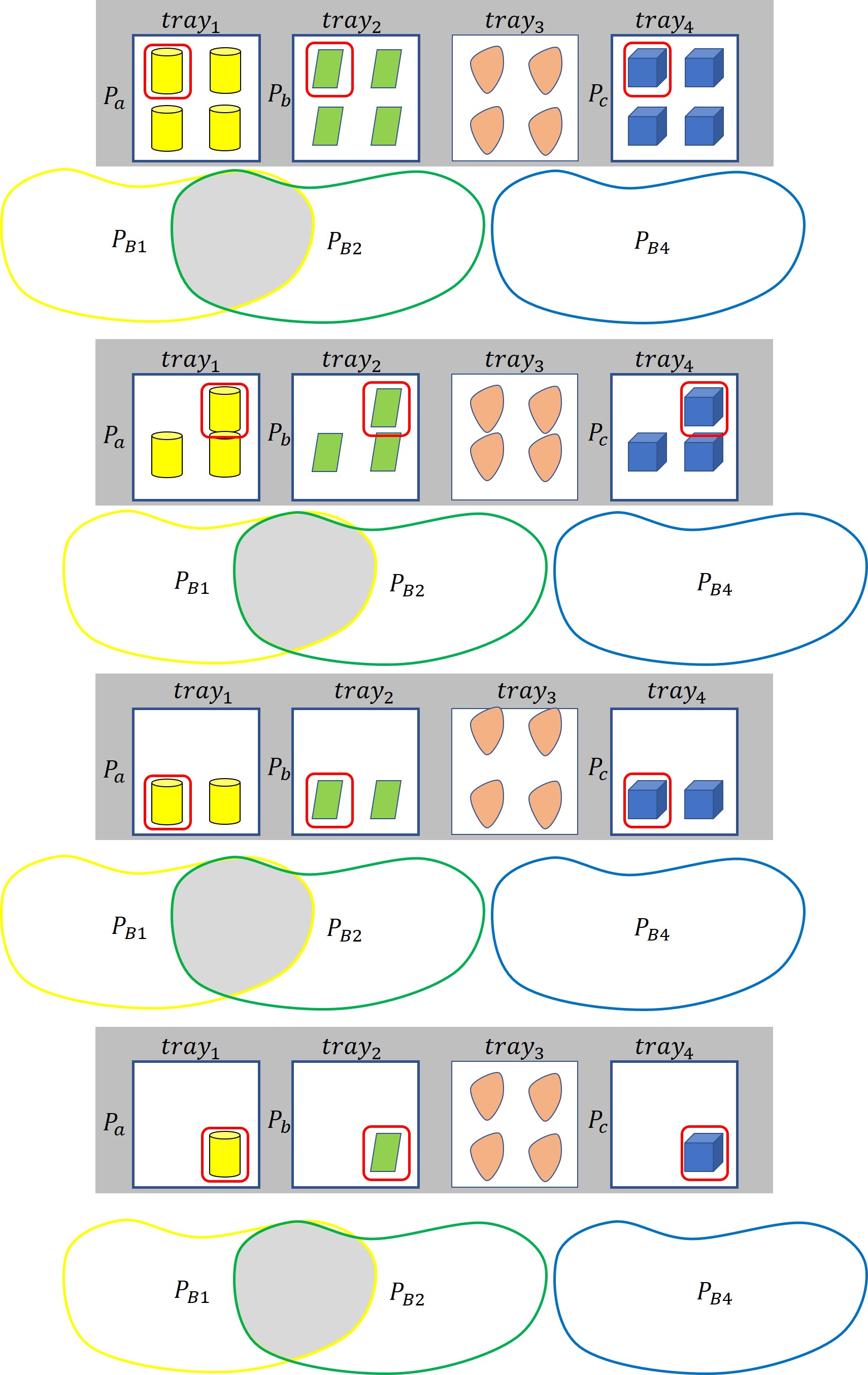}
        \caption{}
    \end{subfigure}
    \caption{Overview of the method (top view). Several types of parts are required in the task, one or more parts of each type will be collected. The sub-figures shows three types of schemes respectively: (a) Global static base positions for collecting all the objects in the tray. (b) The base positions update based on the remaining objects in the tray. (c) The base positions update based on the objects to be picked, which are surrounded by red lines, in a round of pick-and-place task.}
    \label{fig:method_overview}
\end{figure}

Fig. \ref{fig:method_overview}a shows the overview of the task. In order to assemble a product {\it P}, which consists of three assembly parts {\it $P_a$}, {\it $P_b$} and {\it $P_c$} stored in three different trays {\it $tray_1$}, {\it $tray_2$} and {\it $tray_4$}, we focus on planning the base sequence for collecting the target assembly parts from the containing trays, the following information is assumed to be known: (1) The types of parts to be collected and their associated trays. (2) The geometrical models of the trays and the potential obstacles in the environment. (3) The poses of the trays and obstacles. Since the target application scenario is in the manufacturing environment, the above information is readily available. The grasping poses for the objects can be obtained using either model-based or model-free method depending on whether the objects are regularly or randomly placed in the tray, in the first case, we need the geometric models of the objects, and in the later case, the grasping poses can be estimated by the model-free method described in section \MakeUppercase{\romannumeral 7}-B.

Taking advantage of the precomputed reachability database, for every sampled base position, we query the grasping poses from the database to obtain the approximated IK solutions, the base position is valid if there exist collision-free IK solutions for grasping the target objects in one tray. The base region of a tray is formulated as the collection of base positions, where the mobile manipulator is able to grasp the targets from a tray without self-collision and the collision with environment. Let $P_{Bi}$ denote the base region of $tray_{i}$, $P_{Bi} \cap P_{Bj}$ denotes the intersection of $P_{Bi}$ and $P_{Bj}$, the base movements can be reduced by moving the manipulator to the intersection of different base regions, for example, the mobile manipulator moves to $P_{B1} \cap P_{B2}$, and is able to pick up $P_a$ and $P_b$, then it moves to $P_{B4}$ to pick up $P_c$.

The intersection might be very small, such that the mobile manipulator will fail to collect all the targets if the actually arrived position significantly deviates from the planned base position. Considering base positioning uncertainty, we set a threshold for the size of intersections, such that an intersection is applied to reduce the number of base movements only when the radius of its inscribed circle is above the threshold, thus it is robust to a certain level of base positioning uncertainty. The threshold is determined based on base positioning performance in experiments. Then a set of robust base positions are selected from the base regions or intersections, and connected by the shortest path.

Fig. \ref{fig:method_overview}a, b and c illustrate three schemes that can be implemented in practical applications, they differ in the target objects used to calculated the base positions. In Fig. \ref{fig:method_overview}a, all the objects in a tray are used to plan a globally static base sequence. Fig. \ref{fig:method_overview}b shows the update of base regions based on the remaining objects in the tray, after every round of pick-and-place, the remaining objects decreases while the area of base region increases, which improves the overall robustness. Fig. \ref{fig:method_overview}c shows another policy for updating the base region according to the objects to be picked, a good picking order may reduce the variance of the robustness in different rounds of pick-and-place tasks. 

\section{Inverse Kinematics}
The inverse kinematics problem is to determine a set of joint angles that bring the end effector to a desired pose. In this study, we aim at obtaining a nearly complete set of base positions where there exists at least one collision-free IK solution that reaches desired grasping poses. Since solving IK and checking collision are performed separately, this leads to an inherently incomplete planner that cannot obtain all the feasible base positions. In terms of collision check between the mobile manipulator and the environment, it is helpful to find collision-free manipulator configurations by generating a variety of candidate IK solutions that cover all kinds of manipulator configurations, thus it is not likely to miss a feasible base position in which collision-free IK solutions can be found. For non-redundant manipulators, 
it is feasible to find out all the IK solutions and preform collision check between the manipulator and the environment. However, there might be an infinite number of IK solutions for a redundant manipulator, a common IK solver returns only one or multiple but not necessarily representative IK solutions, in that case, the IK solver might only find IK solutions that consequently fail in collision check, even if there exist collision-free solutions. Therefore, for redundant manipulators, it is important to find out representative IK solutions for further collision check. Parametrized IK approaches for 7-DOF redundant manipulator \cite{shimizu2008,singh2010} are able to find all the feasible IK solutions, while this method is usually manipulator-specific. We propose a manipulator independent method of obtaining approximated representative IK solutions, by querying the vicinity of a target pose in the reachability database, and it is proved to be a resolution complete method for solving IK.

\subsection{Reachability Database}
The reachability database is constructed by intensively sampling in the joint space of the manipulator, reachable poses are obtained by calculating forward kinematics (FK). This may introduce the preference of singular configurations, that large variation in joint angles only result in small difference in the grasping poses, the manipulability measure \cite{yoshikawa1985} can be applied to relieve the congestion of configurations near the singular configurations. However, we simply employ the uniform sampling scheme in the joint space, which does not affect the resolution completeness. 

For further query of poses comprised of both translational and rotational parts, 6 dimensional voxels are adopted, thus the workspace is discretized in both position ($x,y,z$) and orientation (represented by roll $\alpha$, pitch $\beta$ and yaw $\gamma$), the grid lengths are $\Delta x$, $\Delta y$, $\Delta z$, $\Delta \alpha$, $\Delta \beta$ and $\Delta \gamma$, respectively. All the resultant poses calculated by forward kinematics, together with their joint angles, are stored in the corresponding 6 dimensional voxels according to their positions and orientations. For example, a grasping pose $\mathcal{G}_i = [x_i, y_i, z_i, \alpha_i, \beta_i, \gamma_i]^T$, should be stored in the voxel indexed by ($x_i/\Delta x$, $y_i/\Delta y$, $z_i/\Delta z$, $\alpha_i/\Delta \alpha$, $\beta_i/\Delta \beta$, $\gamma_i/\Delta \gamma$), within the voxel is a set of poses with similar position and orientation, thus the vicinity of a target pose can be quickly found by querying the pose from such a data structure.

In our implementation, we use step size of 0.35$rad$ to uniformly step though all the valid joint values of the 7DoF arm, as a result, 92 million poses are obtained by solving forward kinematics. These poses are distributed into 2 million 6D voxels in the workspace, the voxels are discretized in 0.1$m$ along $x$, $y$ and $z$ axis, and 0.26$rad$ along $roll$, $pitch$ and $yaw$. The total memory consumption of the reachability database is about 9GB, and it is implemented as a ROS service, which can run on a workstation for IK queries.

\subsection{IK Query}
Instead of resolving the inverse kinematics by an IK solver, we obtain the IK solutions by querying the grasping pose in the database and access the corresponding joint angles. However, the reachability database is only a discrete representation of the continuously varying reachable poses. Probabilistically, we will fail to find an identical grasping pose in the database. Since solving IK and checking collision are treated separately, the IK solutions should be as diverse as possible in order to be resolution complete in finding collision-free IK solutions. Instead, we approximate the IK solutions of the target grasping pose $\mathcal{G}_t$ by querying a range of poses in the vicinity of $\mathcal{G}_t$, for $\mathcal{G}_i \in (\mathcal{G}_t - \Delta \mathcal{G}, \mathcal{G}_t + \Delta \mathcal{G})$, where $\Delta \mathcal{G}=[\Delta x,\Delta y,\Delta z,\Delta \alpha,\Delta \beta,\Delta \gamma]^T$, and the associated manipulator configurations of $\mathcal{G}_i$ are the approximated IK solutions of $\mathcal{G}_t$.

In the reachability database, every 6d voxel contains a small range of poses, firstly we find the voxel containing the target pose, and all the poses within the voxel are regarded as the vicinity of the target pose. For example, by querying pose $[0.6,0,0.8,0,0.5,0.5]^T$ from the database, the indexed voxel is found to contain 151 poses, and Fig. 4 shows a part of the manipulator configurations among them, these are the approximated IK solutions of the target pose\footnote{If exact IK solutions are desired, it is recommended to use the approximated IK solution as the initial seed in a numerical IK solver, then it can quickly converge to the exact solution after a few iterations}. As the database resolution goes to infinity, the approximation error approaches zero and the queried manipulator configurations include complete IK solutions of the target pose, such that, collision-free IK solutions can be found if there exist, regardless the location of the obstacles.

\begin{figure}
    \centering
    \begin{subfigure}[h]{0.15\columnwidth}
        \centering
        \includegraphics[width=1\textwidth]{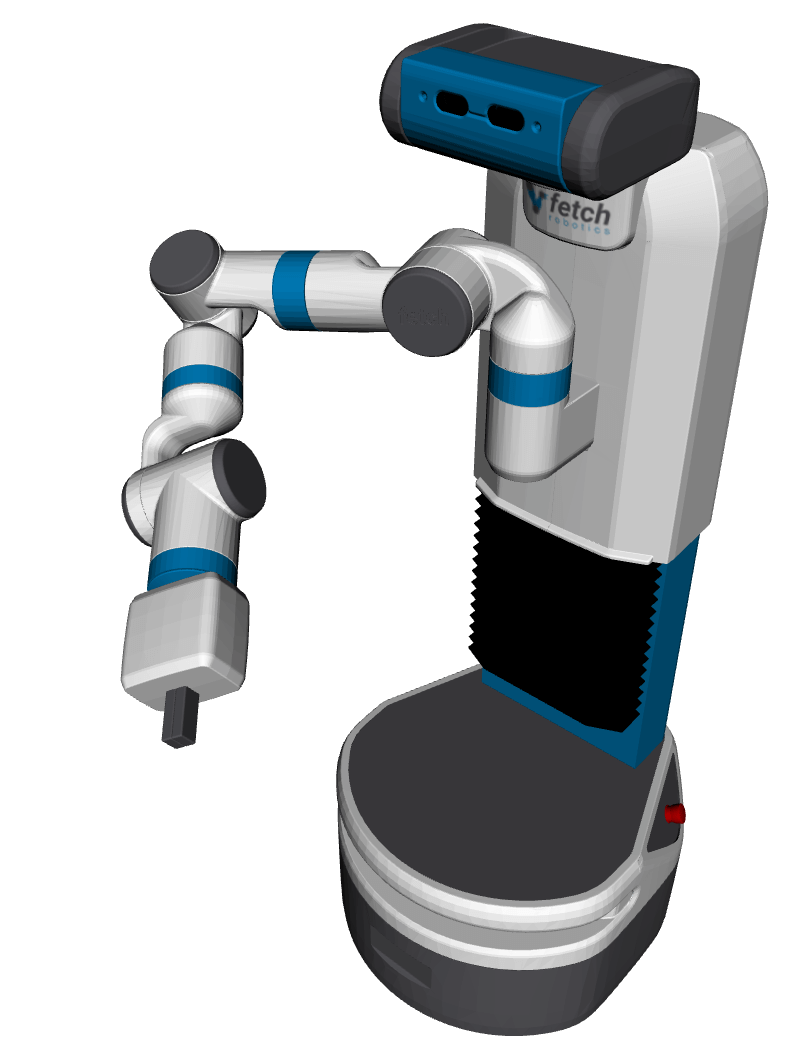}
    \end{subfigure}
    \begin{subfigure}[h]{0.15\columnwidth}
        \centering
        \includegraphics[width=1\textwidth]{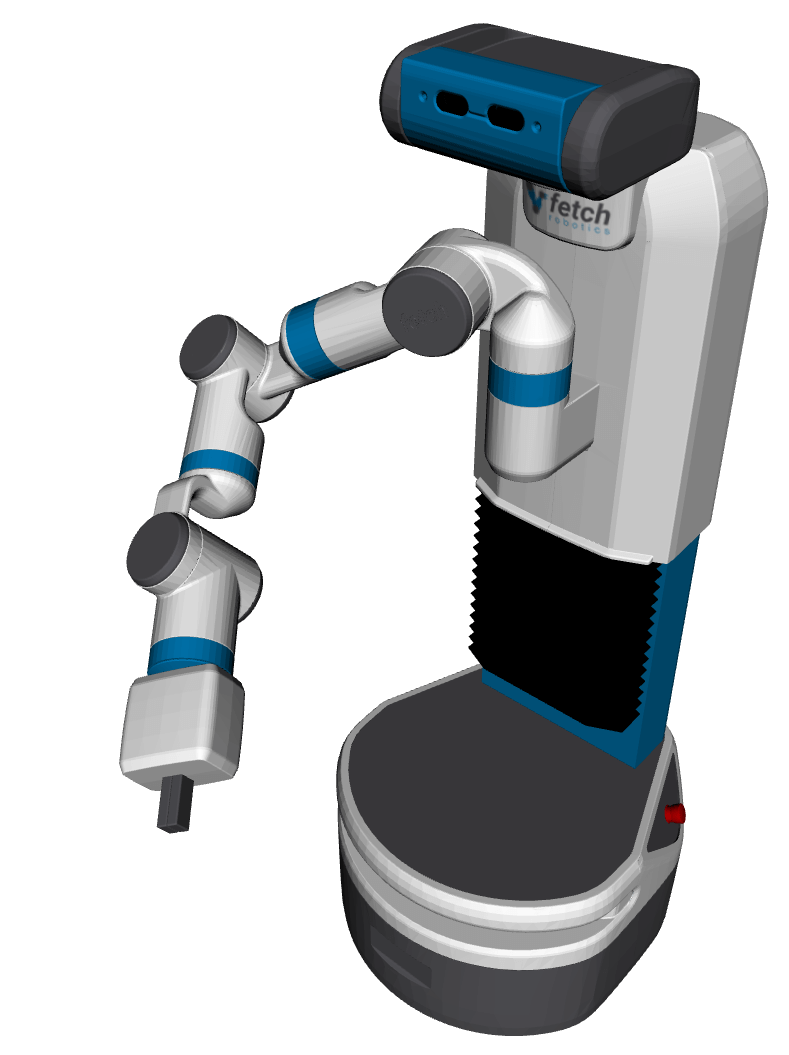}
    \end{subfigure}
    \begin{subfigure}[h]{0.15\columnwidth}
        \centering
        \includegraphics[width=1\textwidth]{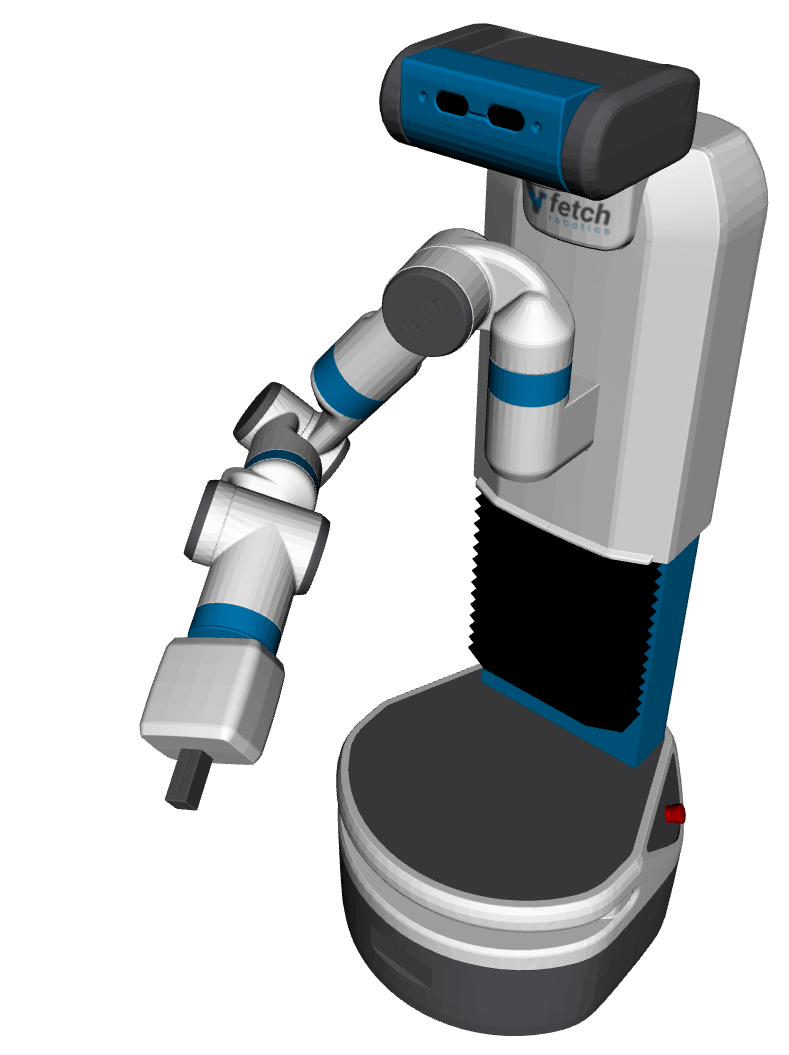}
    \end{subfigure}
    \begin{subfigure}[h]{0.15\columnwidth}
        \centering
        \includegraphics[width=1\textwidth]{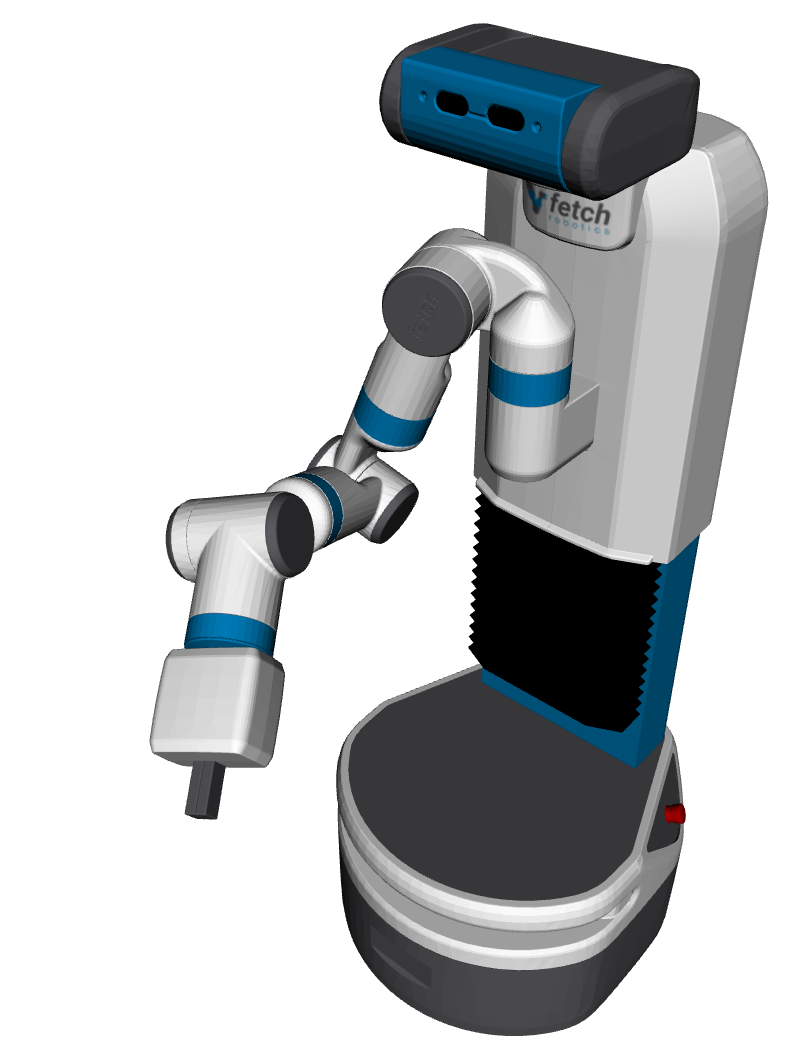}
    \end{subfigure}
    \begin{subfigure}[h]{0.15\columnwidth}
        \centering
        \includegraphics[width=1\textwidth]{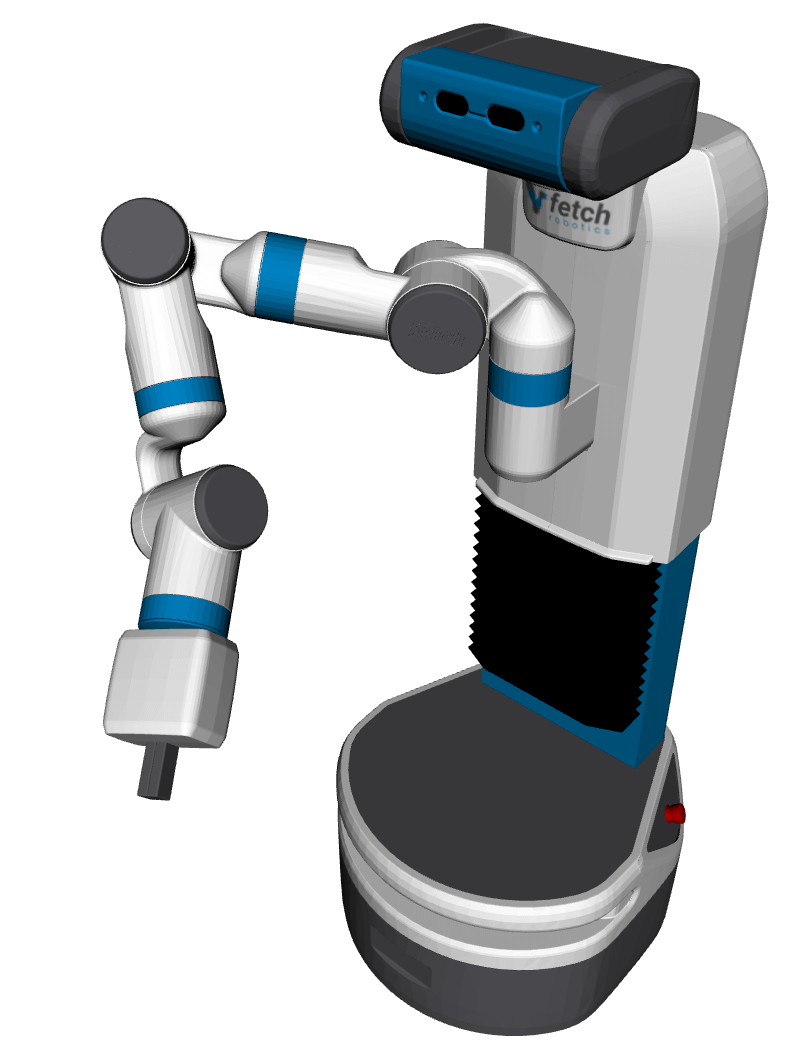}
    \end{subfigure}
    \begin{subfigure}[h]{0.15\columnwidth}
        \centering
        \includegraphics[width=1\textwidth]{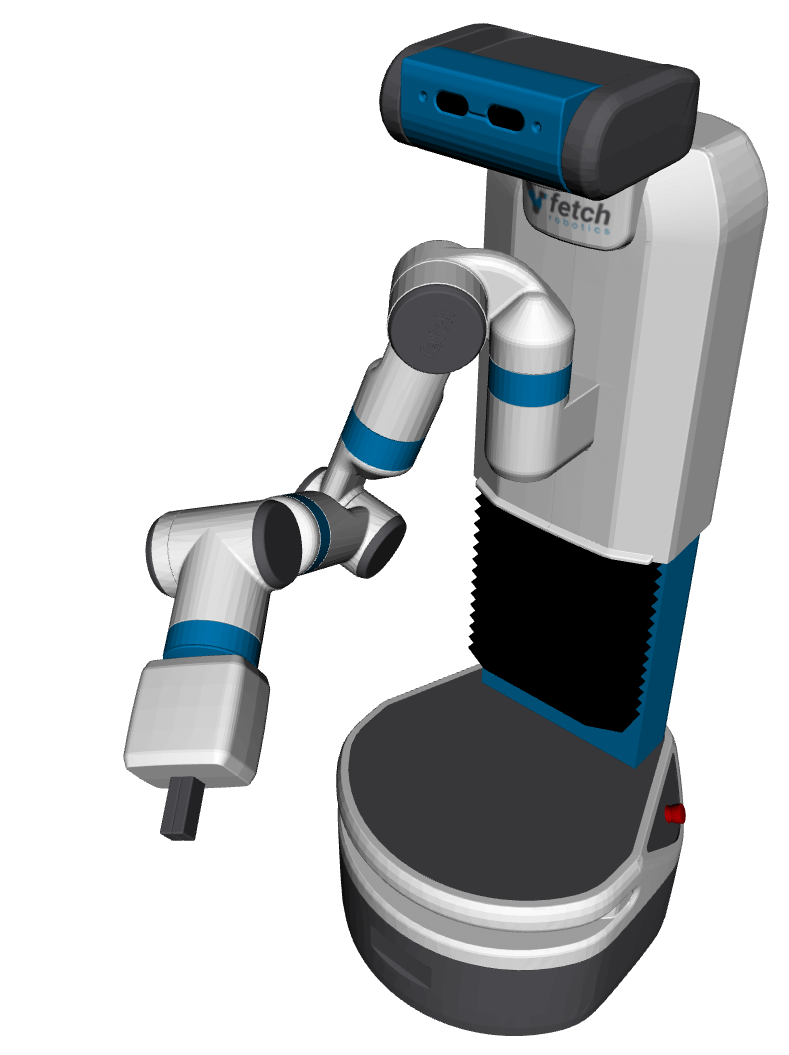}
    \end{subfigure}
    \begin{subfigure}[h]{0.15\columnwidth}
        \centering
        \includegraphics[width=1\textwidth]{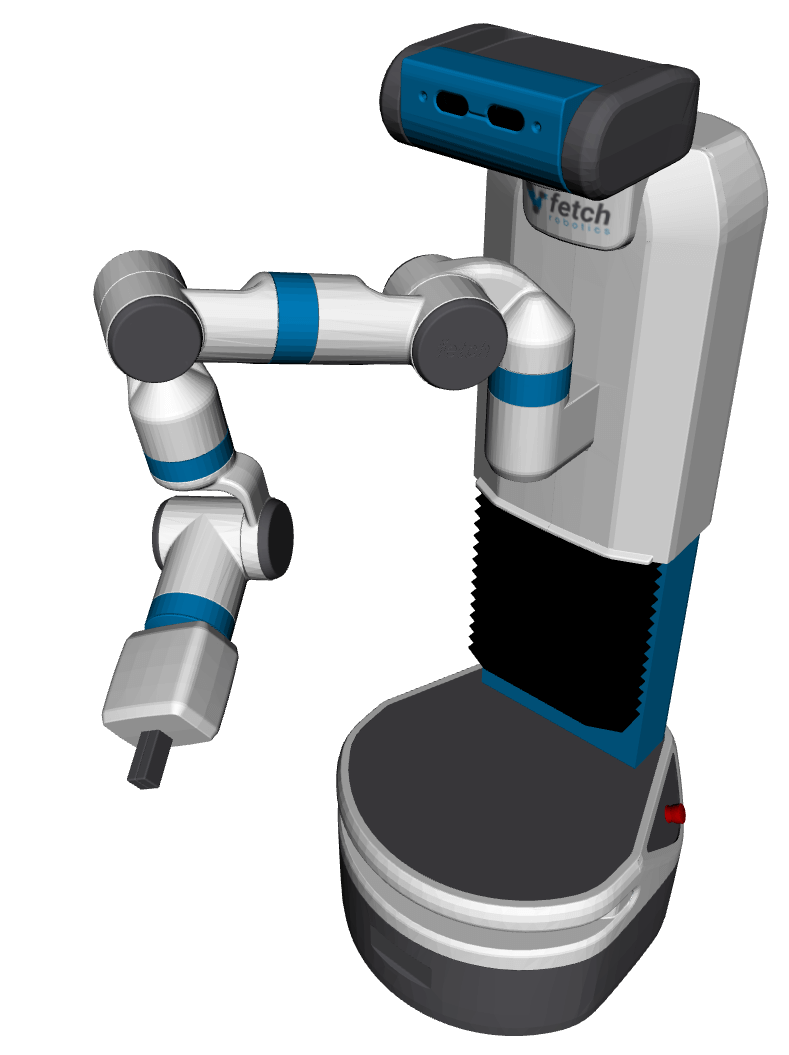}
    \end{subfigure}
    \begin{subfigure}[h]{0.15\columnwidth}
        \centering
        \includegraphics[width=1\textwidth]{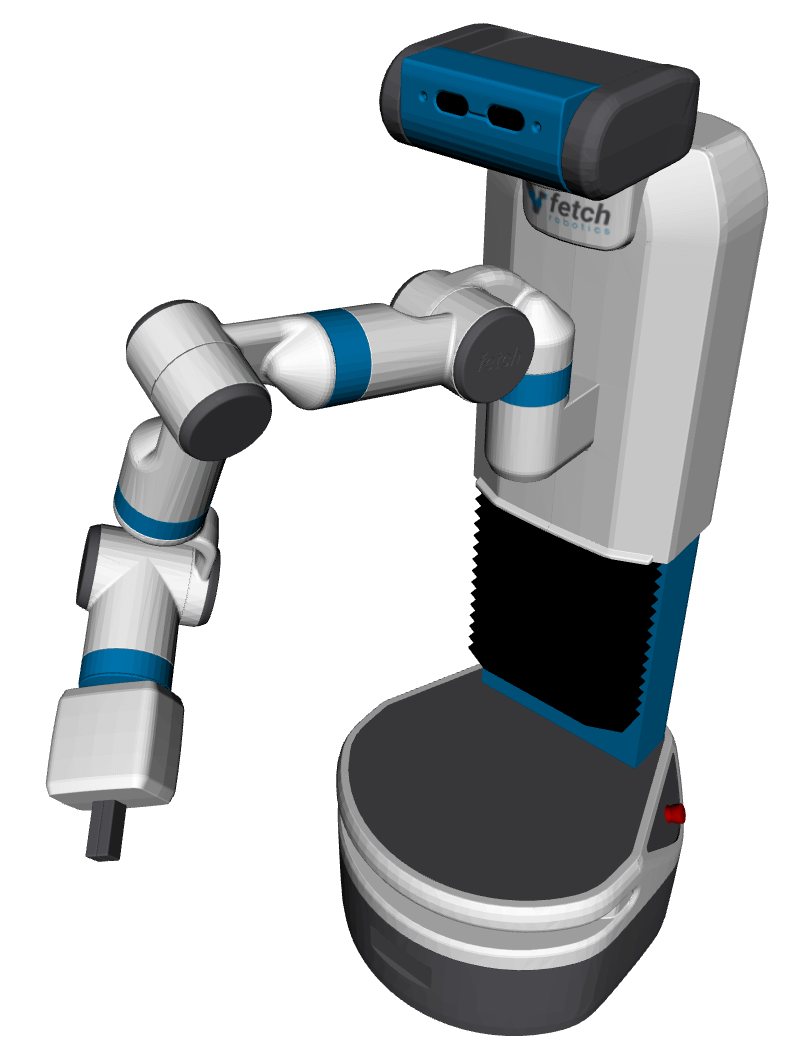}
    \end{subfigure}
    \begin{subfigure}[h]{0.15\columnwidth}
        \centering
        \includegraphics[width=1\textwidth]{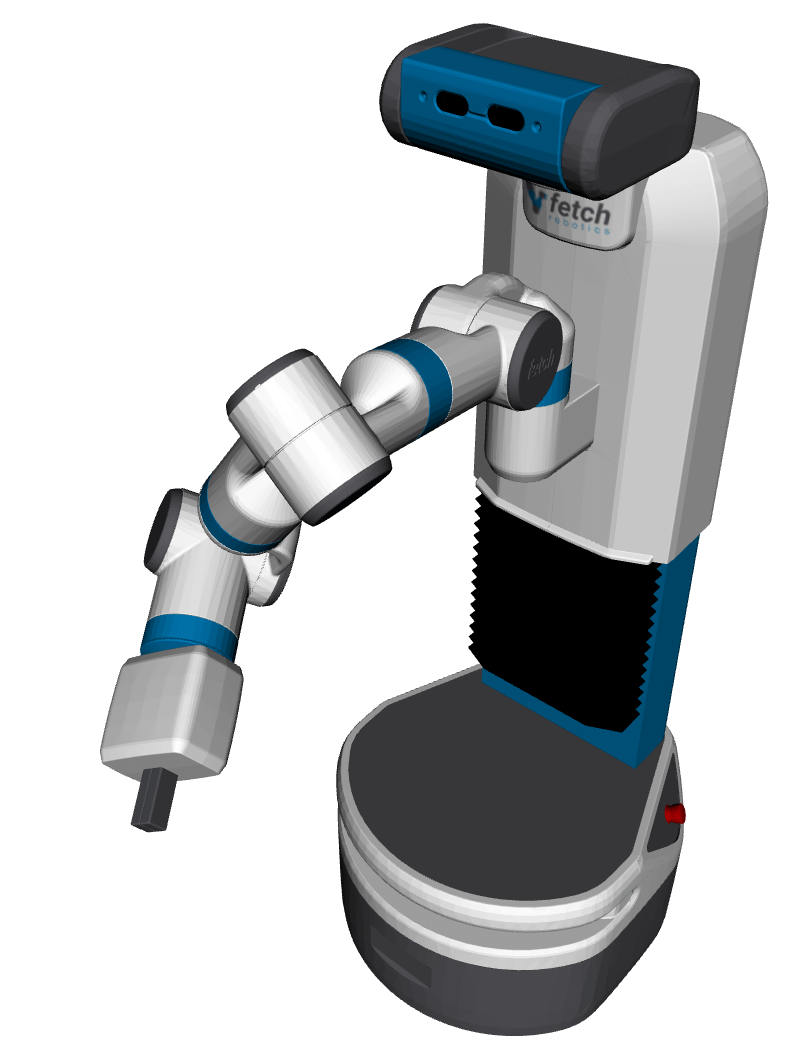}
    \end{subfigure}
    \begin{subfigure}[h]{0.15\columnwidth}
        \centering
        \includegraphics[width=1\textwidth]{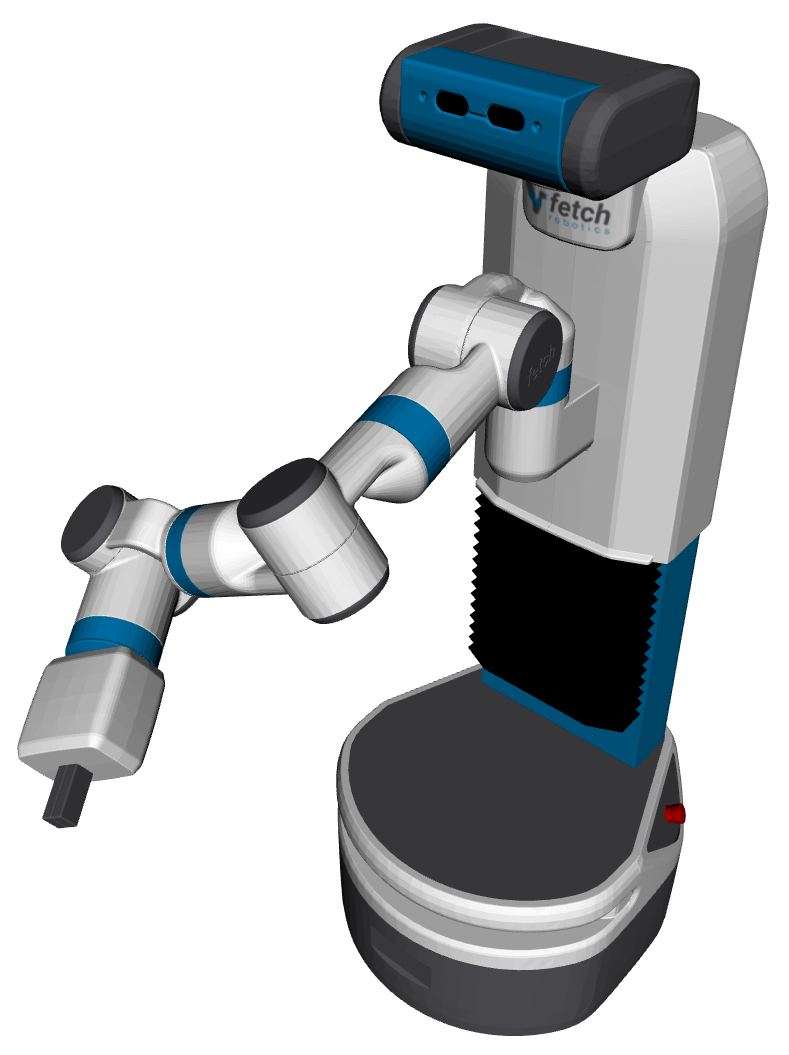}
    \end{subfigure}
    \begin{subfigure}[h]{0.15\columnwidth}
        \centering
        \includegraphics[width=1\textwidth]{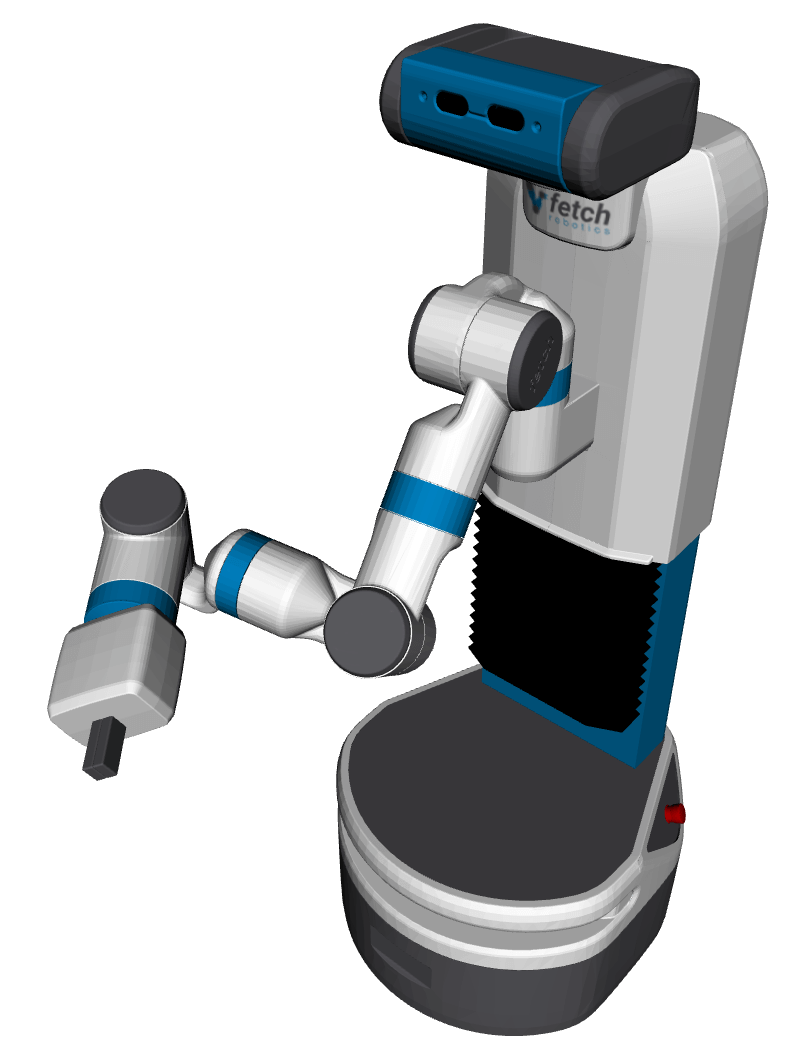}
    \end{subfigure}
    \begin{subfigure}[h]{0.15\columnwidth}
        \centering
        \includegraphics[width=1\textwidth]{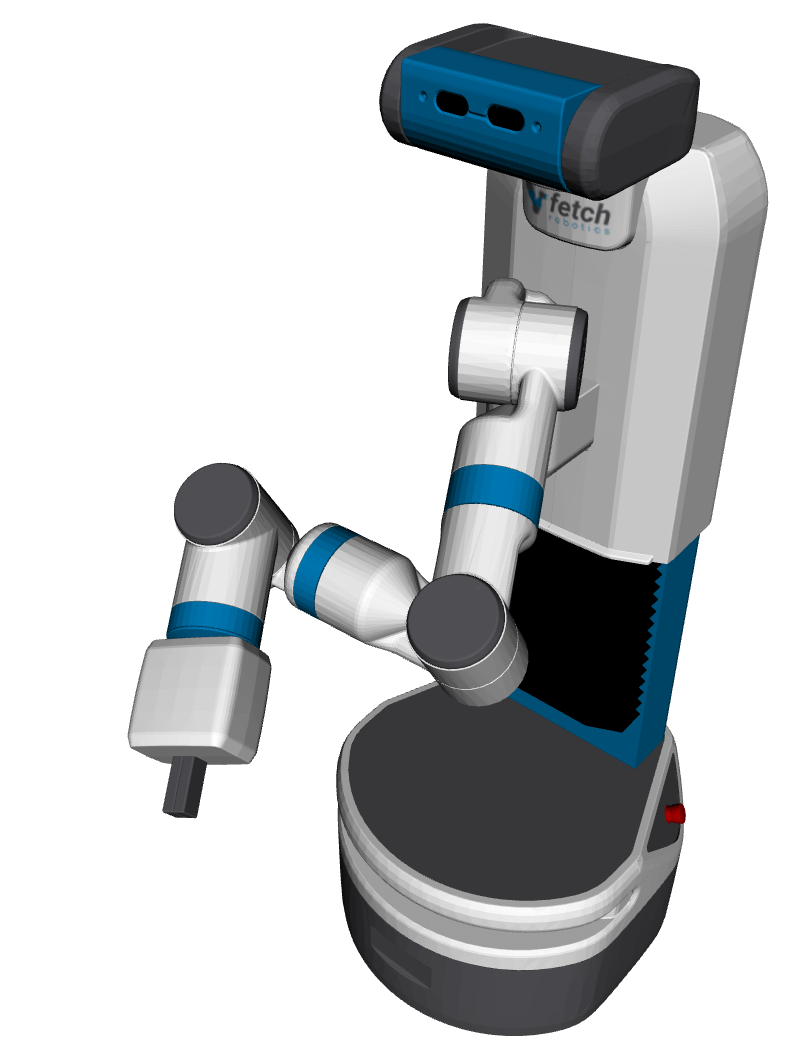}
    \end{subfigure}
    \caption{A part of obtained 151 representative manipulator configurations by querying pose $[0.6,0,0.8,0,0.5,0.5]^T$ from the reachability database, they are the approximation of the exact IK solutions of the target pose, and cover different possible manipulator configurations.}
    \label{fig:representative_IKs}
\end{figure}

The feasibility and completeness of approximating IK solutions of $\mathcal{G}_t$ by the associated manipulator configurations of $\mathcal{G}_i \in (\mathcal{G}_t - \Delta \mathcal{G}, \mathcal{G}_t + \Delta \mathcal{G})$, are proved by the following two lemmas, they are based on the differentiable mapping between configuration space and workspace, except for singular configurations. The first lemma is to prove that, as the sampling resolution goes to infinity, we can always find a manipulator configuration within the voxel, that approaches any one of the IK solutions of the target pose. The second lemma proves that, as the voxelization resolution goes to infinity, all the manipulator configurations within the voxel approach the IK solutions of the target pose. Note that the completeness of approximating all the IK solutions of $\mathcal{G}_t$ is already guaranteed by lemma 1, and lemma 2, together with lemma 1, is to guarantee the completeness of obtained base positions in the next section.

\textbf{Definition:} Let $\Theta = [\theta_1,\theta_2,\dots,\theta_n]^T$ be the manipulator configuration vector, $\Delta \theta$ and $\Delta \mathcal{G}$ be the sampling and voxelization resolution, $IK(\mathcal{G}_t)$ be the IK solution set of $\mathcal{G}_t$, $FK(\Theta)$ be resultant pose calculated by forward kinematics of $\Theta$, $IKRDB(\mathcal{G}_i)$ be corresponding manipulator configuration of $\mathcal{G}_i$ in the reachability database, $J^+$ be the pseudo-inverse of Jacobian matrix $J$, $voxel(\mathcal{G}_t)$ be the voxel containing pose $\mathcal{G}_t$.

\textbf{Lemma 1:} $\forall {\Theta_t = [\theta_1,\theta_2,\dots,\theta_n]^T} \in IK(\mathcal{G}_t)$, $\exists \mathcal{G}_i \in (\mathcal{G}_t - \Delta \mathcal{G}, \mathcal{G}_t + \Delta \mathcal{G})$, $\Theta_i=IKRDB(\mathcal{G}_i)$, such that $\lim\limits_{\Delta \theta \to 0} \Vert \Theta_i - \Theta_t \Vert = 0$.

\textbf{Proof:}
Set $\{\Theta \mid FK(\Theta) = \mathcal{G}_t \}$ is equivalent to set $\{\Theta \mid \Theta = IK(\mathcal{G}_t) \}$, if $\Delta \theta \to 0$, then $\{\Theta \mid FK(\Theta) = \mathcal{G}_t\} \subset \{\Theta \mid FK(\Theta) = \mathcal{G}_i \in (\mathcal{G}_t - \Delta \mathcal{G}, \mathcal{G}_t + \Delta \mathcal{G}), \Delta \theta \to 0\}$, thus $\forall{\Theta_t} \in IK(\mathcal{G}_t)$, $\exists \mathcal{G}_i \in (\mathcal{G}_t - \Delta \mathcal{G}, \mathcal{G}_t + \Delta \mathcal{G})$, $\Theta_i=IKRDB(\mathcal{G}_i)$, such that 
$\lim\limits_{\Delta \theta \to 0} \Vert \Theta_i - \Theta_t \Vert = 0$.

\textbf{Lemma 2:} $\forall{\mathcal{G}_i} \in (\mathcal{G}_t - \Delta \mathcal{G}, \mathcal{G}_t + \Delta \mathcal{G})$, $\Theta_i=IKRDB(\mathcal{G}_i)$, $\exists \Theta_t \in IK(\mathcal{G}_t)$, that $\lim\limits_{\Delta \mathcal{G} \to 0} \Vert \Theta_i - \Theta_t\Vert  = 0$.

\textbf{Proof:} $\dot{\mathcal{G}} = J(\Theta)\dot{\Theta}$, $\dot{\Theta} = J^{+}(\Theta)\dot{\mathcal{G}}+(I-J^{+}J)k $, where $k$ is an arbitrary vector denoting redundancy, integrate two sides of the formula by a small time step, $\Delta \Theta = J^{+}(\Theta)\Delta \mathcal{G}+(I-J^{+}J)k\Delta t$, then $\forall{\Delta \mathcal{G}}\to 0$, $\exists k =0$ such that $\Delta \Theta \to 0$. Because $\vert \mathcal{G}_i - \mathcal{G}_t \vert < \Delta \mathcal{G}$, $\Delta \mathcal{G} \to 0 \Rightarrow \vert \mathcal{G}_i - \mathcal{G}_t \vert \to 0$, thus $\lim\limits_{\vert \mathcal{G}_i - \mathcal{G}_t \vert \to 0} \Vert \Theta_i - \Theta_t\Vert  = 0$. (replace $J^+$ with $J^{-1}$ for non-redundant manipulators)

The above lemmas apply to both redundant and non-redundant manipulators, the only problem with this method is that the continuous mapping between joint space and workspace breaks down at singular configurations.


\section{Base Sequence Planning}
Regardless there are different definitions of "target" objects, they share the same algorithm for calculating the base region from the target objects, as introduced in this section.
\subsection{Grasp Planning}
In order to determine the base positions, the end effector poses have to be derived from the object pose and grasping poses. The mobile manipulator should be able to grasp every target object with at least one grasping poses. The grasp planning method varies for different object placement. For objects regularly placed in the tray, model-based grasp planner \cite{harada2008,wan2020planning} can be used to prepare a set of grasps for the target object $\mathcal{O}_j$ in the tray in the offline phase. For objects randomly placed in the tray, we treat the objects in the tray as a whole and use a model-free grasp planning method \cite{domae2014fast} to estimate a set of grasps from depth images, the details are described in in section \MakeUppercase{\romannumeral 7}-B.

\subsection{Base Region Calculation}
\begin{figure}
    \centering
    \includegraphics[width=0.9\columnwidth]{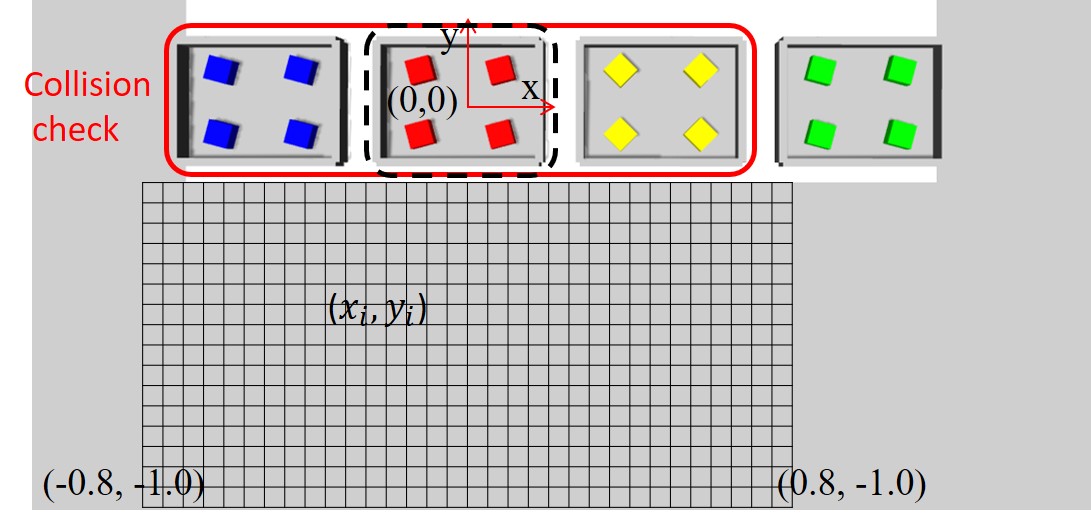}
    \caption{The sampled base positions for one of the trays, circled by black dashed line. The range of sampling is determined referring to reachable workspace in section \MakeUppercase{\romannumeral 4}-A. Collision check is performed between the mobile manipulator and the target tray, its neighboring trays and other obstacles in the environment.}
    \label{fig:base_sampling}
\end{figure}

The base region for a tray is a set of base positions where the mobile manipulator is able to reach all the targets in the tray, with avoiding self-collision and the collision with the environment. Firstly, we prepare stable grasping poses with respect to the object for every target object in the tray, using a grasp planner. Then sample base poses $(x_i, y_i, \phi)$ in front of the target tray, as illustrated in Fig. \ref{fig:base_sampling}, here we assume that the orientation $\phi$ of the mobile manipulator is constant and keep the mobile manipulator facing the tray, because for many mobile manipulators, the joint connecting the manipulator and mobile base rotates around a vertical axis, thus counteracts the rotation of the mobile base and contributes almost nothing new. Then the set of stable grasping poses $\{\mathcal{G}_{t1},\mathcal{G}_{t2},\dots,\mathcal{G}_{tn}\}_j$ with respect to the mobile base for object $\mathcal{O}_j$ is obtained for every target object in the tray. Finally, if there exists a grasping pose $\mathcal{G}_{ti} \in \{\mathcal{G}_{t1},\mathcal{G}_{t2},\dots,\mathcal{G}_{tn}\}_j$, such that we can find a collision-free manipulator configuration $IKRDB(\mathcal{G}_k)$ for a pose $\mathcal{G}_k \in voxel(\mathcal{G}_{ti})$, then $\mathcal{O}_j$ can be grasped from the base position, a base position for the tray is feasible if all the target objects in the tray can be grasped. All the feasible positions constitute the base region for grasping target objects from the tray. Considering the base region for a tray may update due to the change of target objects, it is preferable to calculate the base region for every object in the tray and save such information for further access, then the base region for a tray can be quickly solved, which is the intersection of the base regions of the target objects.

\begin{figure}
    \centering
    \includegraphics[width=\columnwidth]{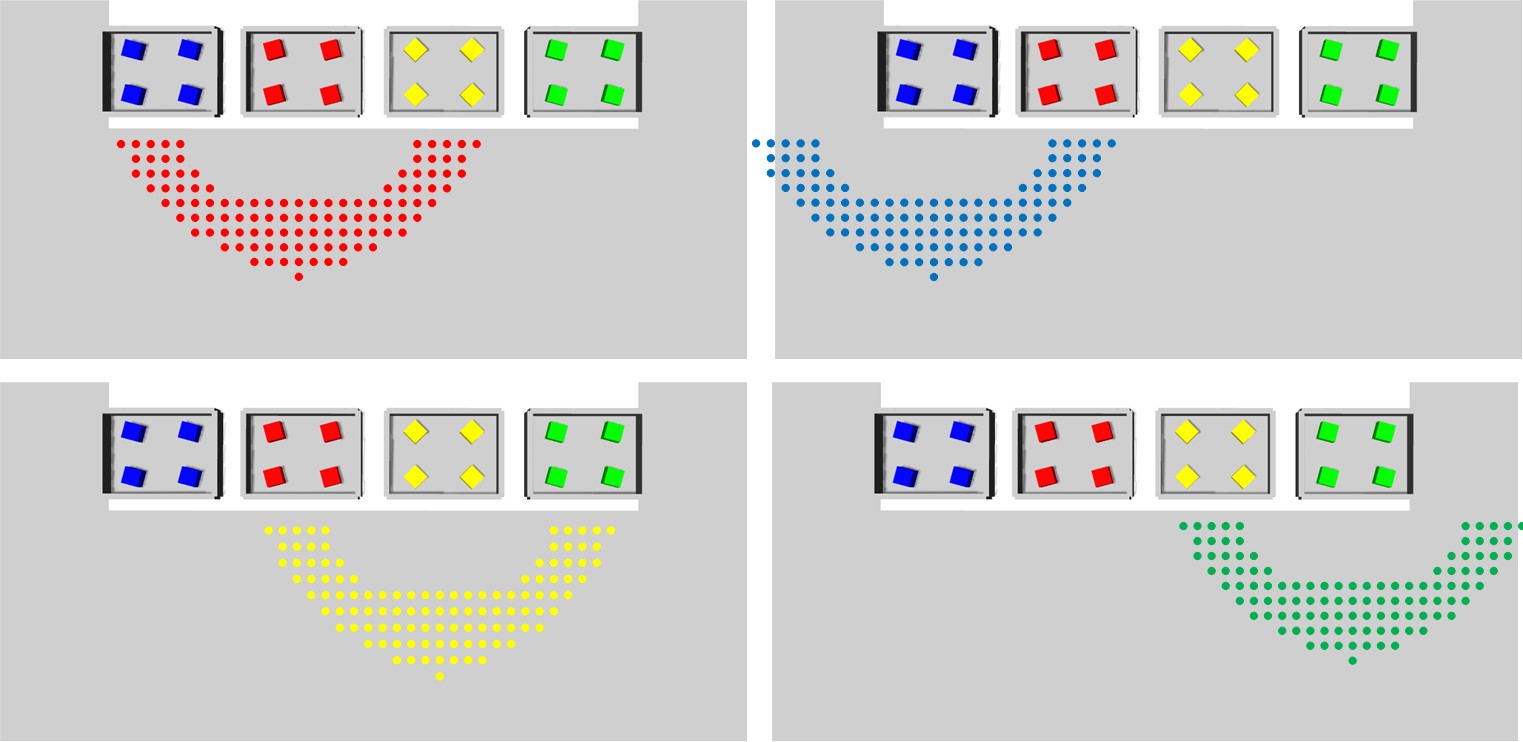}
    \caption{The base regions obtained by using the IKFast solver.}
    \label{fig:base_regions_IKsolver}
\end{figure}

\begin{figure}
    \centering
    \includegraphics[width=\columnwidth]{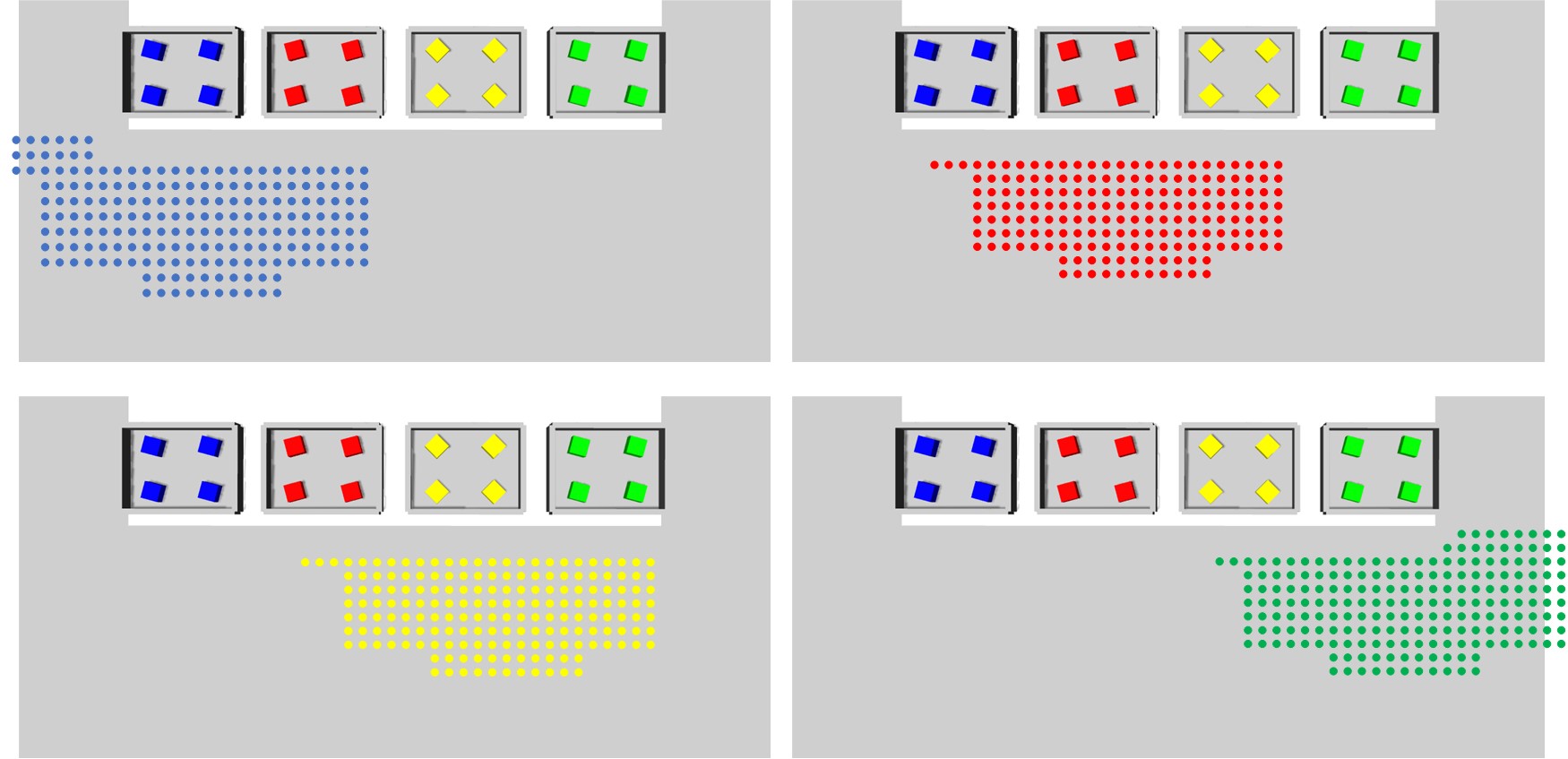}
    \caption{The base regions obtained by the proposed IK query method.}
    \label{fig:one_bin_result_IKRDB}
\end{figure}

We compare the base regions obtained by different IK approaches, notice that the presented examples in this section assume the target objects to be all the objects in the tray. The obtained base regions using IKFast plugin to find the IK solutions are shown in Fig. \ref{fig:base_regions_IKsolver}. Fig. \ref{fig:one_bin_result_IKRDB} shows the base regions calculated by IK query approach proposed in this paper, it is obvious the base region is larger. In the IK solver approach, IK solutions are not found within given attempts in some base positions, and some of the found IK solutions fail in the collision check. In the IK query approach, since diverse IK solutions are returned, the obtained base positions are closer to complete. However, due to finite resolution of the reachability database, the obtained base positions are not always feasible especially for the boundary positions, but fortunately, as described in the rest of this section, only the center of the base region or intersection is selected, which relaxes this limitation.

\subsection{Intersections of Base Regions}
\begin{figure}
    \centering
    \includegraphics[width=\columnwidth]{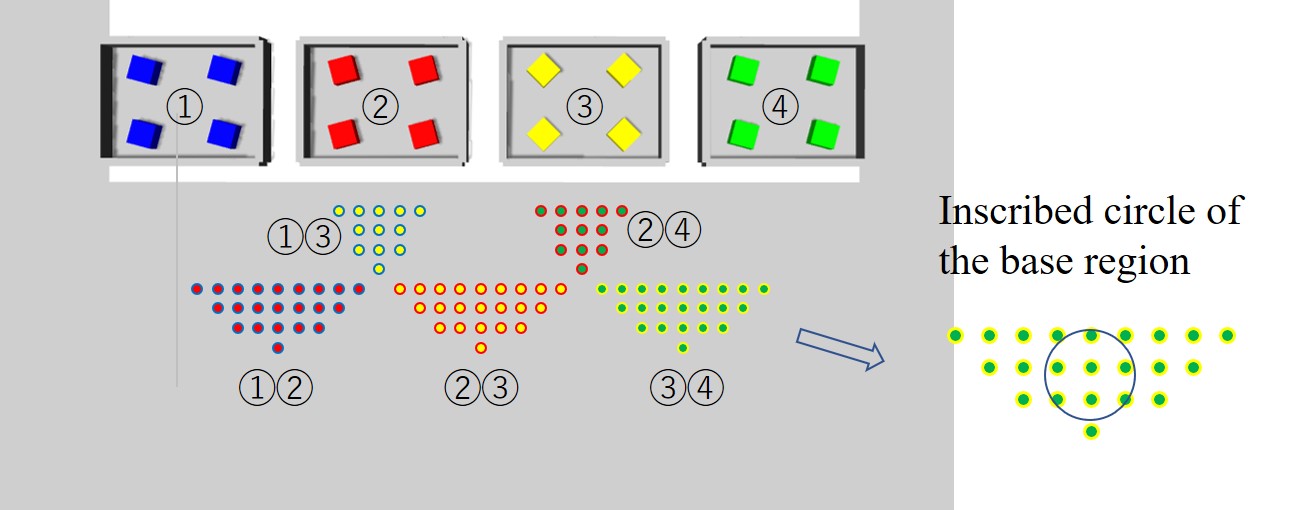}
    \caption{The intersections of the base regions in Fig. \ref{fig:base_regions_IKsolver}, the centers of their inscribed circles are the most robust base positions.}
    \label{fig:IK_intersections}
\end{figure}

\begin{figure}
    \centering
    \begin{subfigure}[h]{0.32\columnwidth}
        \centering
        \includegraphics[width=1\textwidth]{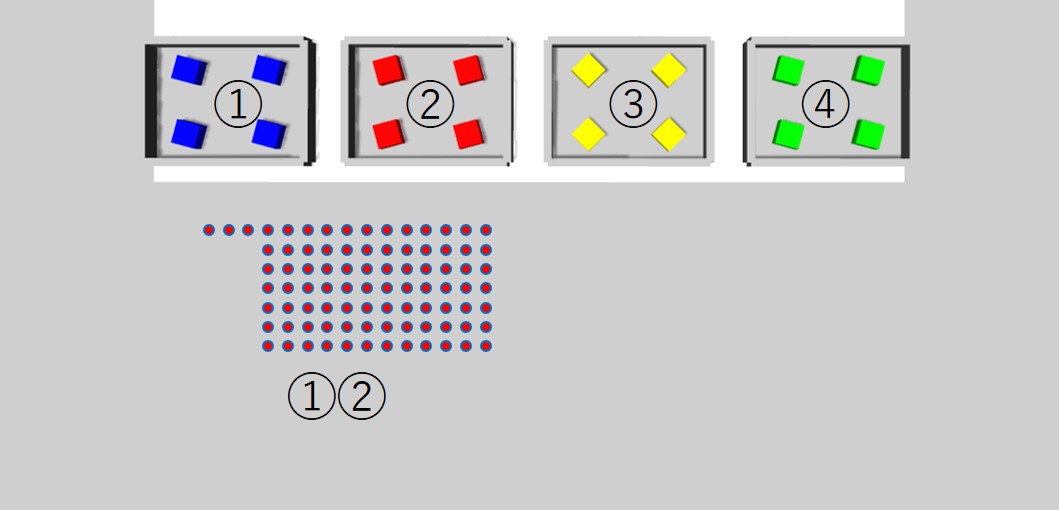}
        \caption{$P_{B1} \cap P_{B2}$}
    \end{subfigure}
    \begin{subfigure}[h]{0.32\columnwidth}
        \centering
        \includegraphics[width=1\textwidth]{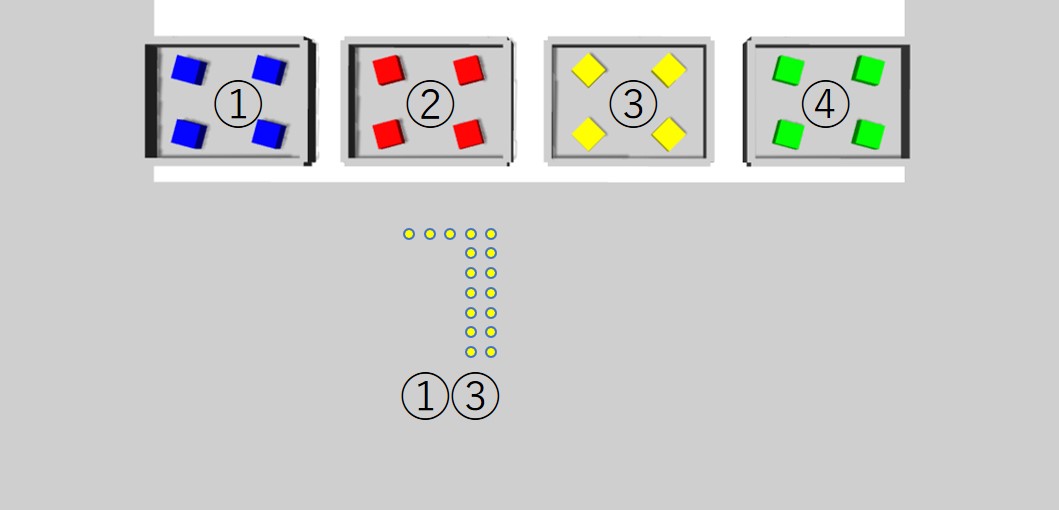}
        \caption{$P_{B1} \cap P_{B3}$}
    \end{subfigure}
    \begin{subfigure}[h]{0.32\columnwidth}
        \centering
        \includegraphics[width=1\textwidth]{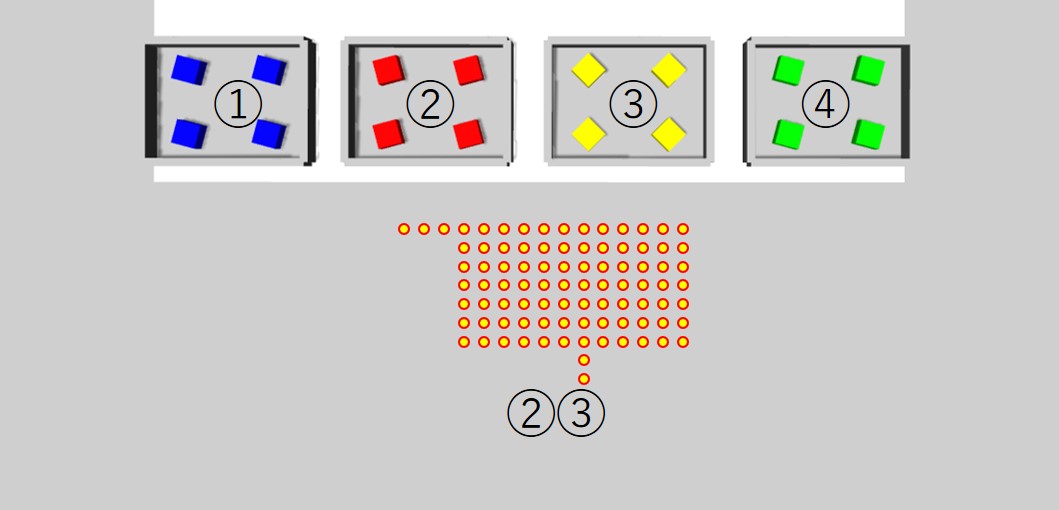}
        \caption{$P_{B2} \cap P_{B3}$}
    \end{subfigure}
    \begin{subfigure}[h]{0.32\columnwidth}
        \centering
        \includegraphics[width=1\textwidth]{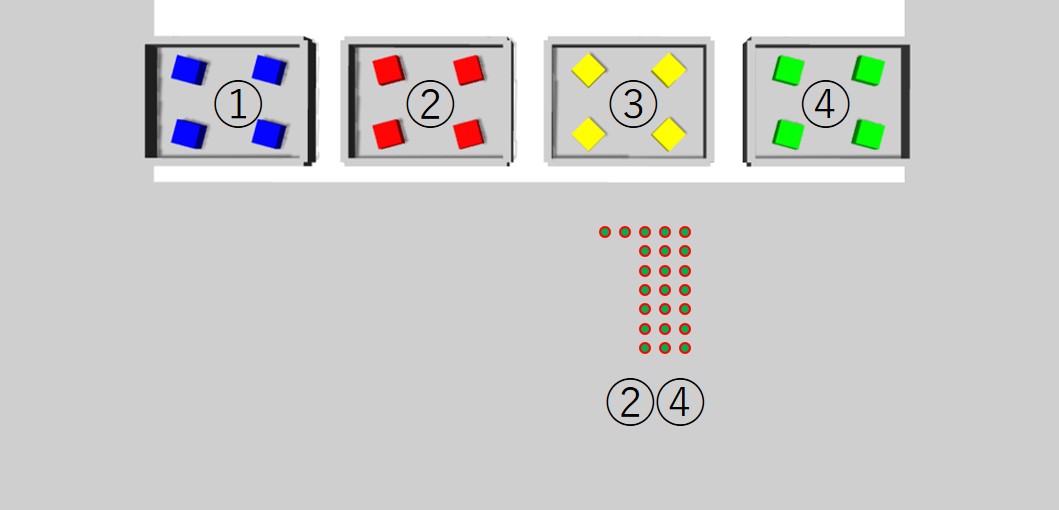}
        \caption{$P_{B2} \cap P_{B4}$}
    \end{subfigure}
    \begin{subfigure}[h]{0.32\columnwidth}
        \centering
        \includegraphics[width=1\textwidth]{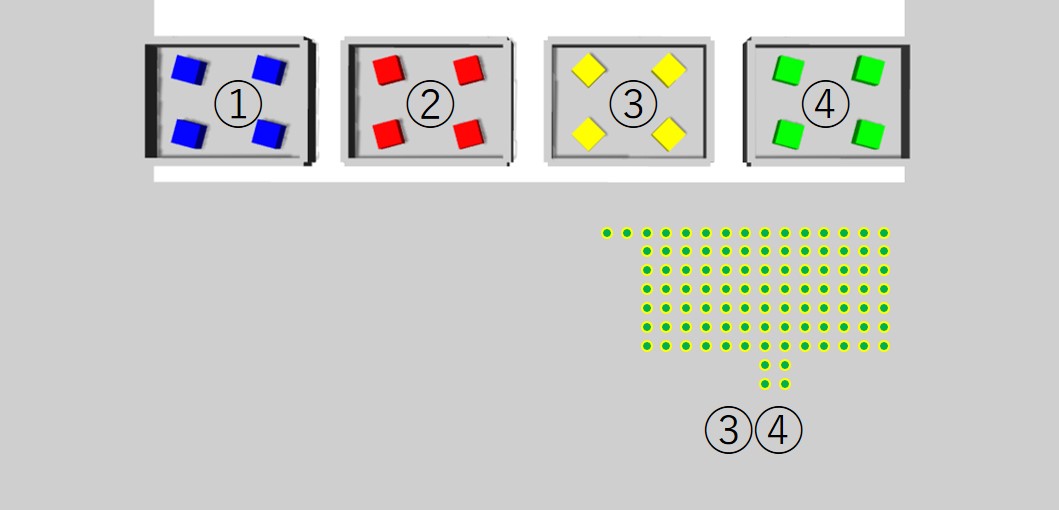}
        \caption{$P_{B3} \cap P_{B4}$}
    \end{subfigure}
    \begin{subfigure}[h]{0.32\columnwidth}
        \centering
        \includegraphics[width=1\textwidth]{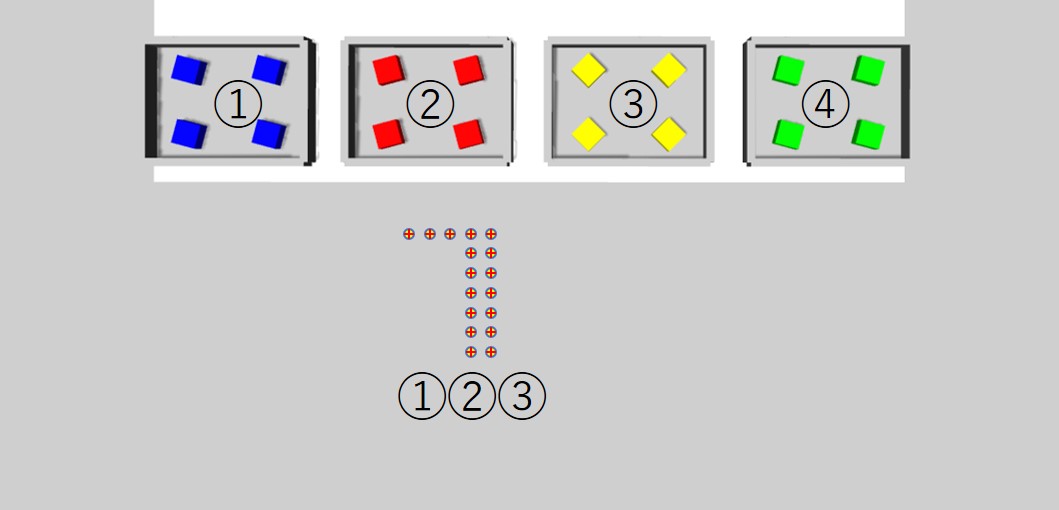}
        \caption{$P_{B1} \cap P_{B2} \cap P_{B3}$}
    \end{subfigure}
    \begin{subfigure}[h]{0.32\columnwidth}
        \centering
        \includegraphics[width=1\textwidth]{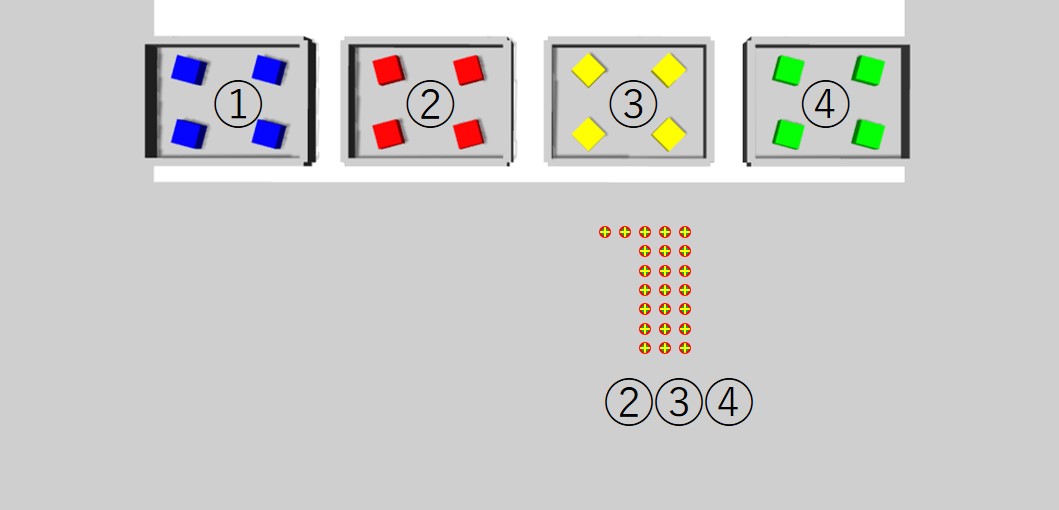}
        \caption{$P_{B2} \cap P_{B3} \cap P_{B4}$}
    \end{subfigure}
    \caption{All the intersections of the base regions in Fig. \ref{fig:one_bin_result_IKRDB}. (a)$\sim$(e) are all the intersections of two base regions, (f)$\sim$(g) are all the intersections of three base regions.}
    \label{fig:IKRDB_intersections}
\end{figure}

To illustrate different circumstances of intersections among the base regions, in this subsection, we consider $tray_1$, $tray_2$, $tray_3$ and $tray_4$ as the target trays. To reduce the number of base movements, the mobile manipulator had better move to the intersections where the mobile manipulator is able to pick up the objects in multiple trays. For the result of IK solver approach, there are 5 intersections for the obtained 4 base regions of 4 trays, all of them are the intersections of two base regions, while for the result of IK query approach, there are the intersections of two base regions and even the intersections of 3 base regions. The intersections and their associated trays are labeled with numbers in Fig. \ref{fig:IK_intersections} and Fig. \ref{fig:IKRDB_intersections}.
\subsection{Base Positioning Uncertainty}
Practically, the mobile manipulator is not able to accurately arrive at the planned position. It will be described in section \MakeUppercase{\romannumeral 7} that, a SLAM package is used to locate the mobile manipulator in the environment, the positioning error, as a result of numerous influencing factors, is assumed to be random and homogeneous in different directions. Let the average base positioning error be $\bar{\sigma}$(m), the mobile manipulator is most likely to arrive at a position $\bar{\sigma}$(m) away from the planned position. Therefore, in some base positions close to the boundary, the mobile manipulator may fail to reach the all the target objects in the tray when positioning error is imposed. The further the base position is from boundary, the more robust the point will be in terms of base positioning uncertainty. Therefore, the most robust base position is specified by the center of the inscribed circle of the intersection (Fig. \ref{fig:IK_intersections}). If the radius of the inscribed circle of an intersection is smaller than the base positioning uncertainty level, the intersection is regarded as unreliable, and should not be applied to reduce the base sequence size. Notice that, the overall operation time is scarcely influenced by choosing different positions within the base region or intersection, due to their limited area, thus the robustness is given much higher priority in this stage without sacrificing much performance.

\subsection{Path Planning}
\begin{figure}
    \centering
    \begin{subfigure}[h]{0.48\columnwidth}
        \centering
        \includegraphics[width=1\textwidth]{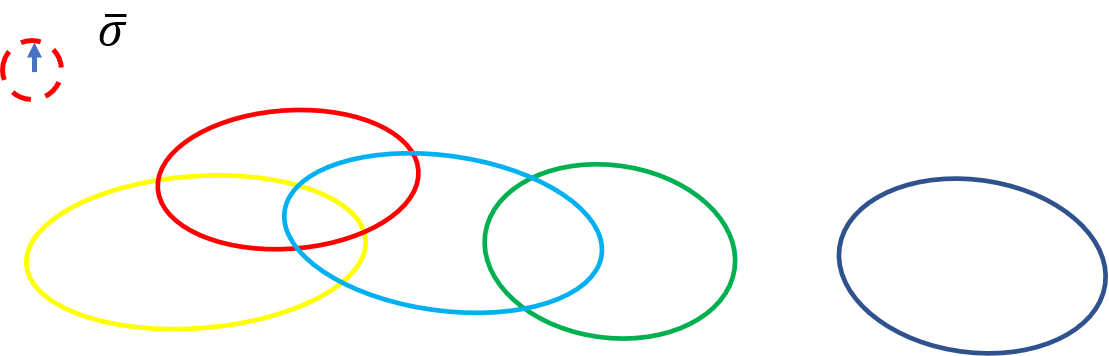}
        \caption{}
    \end{subfigure}
    \begin{subfigure}[h]{0.48\columnwidth}
        \centering
        \includegraphics[width=1\textwidth]{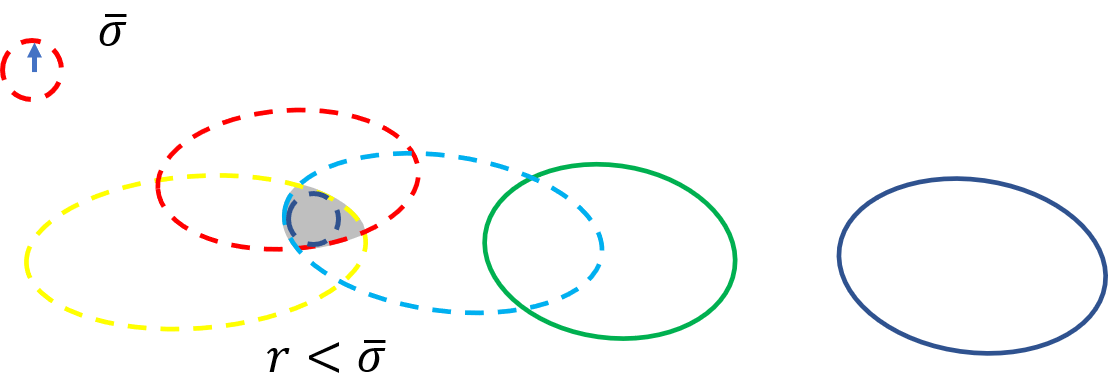}
        \caption{}
    \end{subfigure}
    \begin{subfigure}[h]{0.48\columnwidth}
        \centering
        \includegraphics[width=1\textwidth]{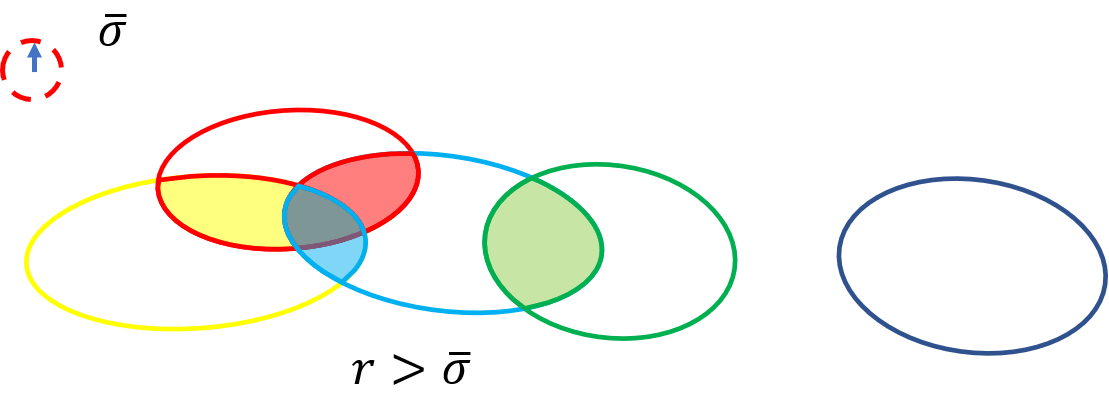}
        \caption{}
    \end{subfigure}
     \begin{subfigure}[h]{0.48\columnwidth}
        \centering
        \includegraphics[width=1\textwidth]{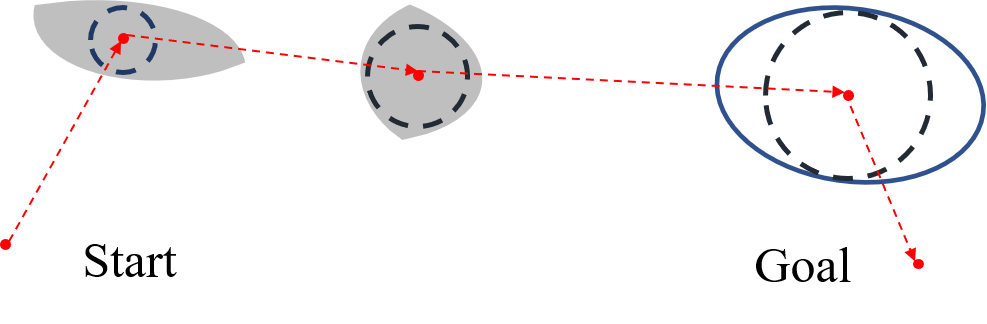}
        \caption{}
    \end{subfigure}
    \caption{The procedure of path planning: (a) From left to right are five base regions $P_{B1} \sim P_{B5}$ for 5 trays. (b) ${\it InscribedRadius}(P_{B1} \cap P_{B2} \cap P_{B3})<\bar{\sigma}$, thus discarded. (c) Four feasible second order intersections are filled with different colors. (d) The planned intersections $P_{B1} \cap P_{B2}$, $P_{B3} \cap P_{B4}$ and $P_{B5}$ are connected by the shortest path.}
    \label{fig:path_planning}
\end{figure}
Given an assembly task to collect the objects in trays \{$tray_1, tray_2,\dots,tray_n$\}, a set of base regions \{$P_{B1}, P_{B2}, \dots, P_{Bn}$\} and their intersections can be obtained following the proposed method. Let $\cap_{i}^{k}P_{Bi}$, $(1 \leq i \leq n)$, denote all the k-th order intersections, which are the intersections of k base regions, base regions themselves are regarded as first order intersections, and $\lambda$ be the largest k, then $\{\cap_{i}^{1}P_{Bi},\cap_{i}^{2}P_{Bi},\dots,\cap_{i}^{\lambda}P_{Bi}\}$ is the set of all the possible intersections. While considering the base positioning uncertainty, the intersection is removed from the set if the radius of its inscribed circle is smaller than $\bar{\sigma}$. Then from the remaining set with size $N$, we select $m$ intersections that cover all the target trays and $m$ is minimized, this is equivalent to the problem of assigning $(N-m)$ zeros and $m$ ones to base sequence vector $[x_1,x_2,\dots,x_N]^T$ and minimizing the sum of its elements, which is formulated as:

\begin{equation}
\begin{array}{ll@{}ll}
\text{minimize} & \sum_{i=1}^N x_i\\
\text{subject to} & x_i = \{0,1\}\\
 & \sum_{i=1}^N a_{1i}x_i=1\\
 & \sum_{i=1}^Na_{2i}x_i=1 \\
 & \dots \\
 & \sum_{i=1}^N a_{ni}x_i=1 
\end{array}
\end{equation}
Here, $a_{si}, s=\{1,2,\dots,n\}$, is 1 if $P_{Bs}$ is reached by the intersection, and $x_i=1$ if the robot moves to the corresponding intersection.
This is the 0-1 knapsack problem and can be solved by BB method \cite{kolesar1967}. Finally, take the centers of these $m$ intersections and connect them, together with start and goal positions, by the shortest path.

For example, in Fig. \ref{fig:path_planning}, there are 5 base regions of 5 trays, one of them is a third order intersection and four of them are second order intersections, however, the third order intersection $P_{B1} \cap P_{B2} \cap P_{B3} < \bar{\sigma}$, thus removed from the total set of intersections. From the remaining 9 intersections (4 second order intersections and 5 first order intersections), three of them are planned to reach all of the trays, then we search for the shortest path, that connects the starting and goal position, via the the centers of their inscribed circles. If the sequence size becomes too large for searching, the shortest path can be approximated by SA method \cite{cerny1985}.

\subsection{Guiding the Design of Tray and Object Configurations}

The base region can be calculated from the target objects to be picked, inversely, the tray and object configurations can be designed in order to achieve a better sequence of base positions. The area of the intersection of base regions depends on the robot kinematic model, the distance between neighboring trays, the size of the tray and the possible object poses in the tray. For instance, the area of intersection decreases with the increase of the distance between trays, for a given mobile manipulator, it is necessary to tune the distance between neighboring trays in order to have more chance of reaching multiple trays at one position, then the overall efficiency can be improved. As shown in Fig. \ref{fig:IKRDB_intersections}a, when the distance between the neighboring trays is 0.5$mm$, the radius of inscribed circle of intersection $P_{B1} \cap P_{B2}$ is larger than the base position uncertainty level, and the largest distance between the neighboring trays is 0.85$mm$ such that the resultant intersection is just enough to accommodate the base positioning uncertainty. In another word, the distance between the neighboring trays should be less than 0.85$mm$. Given other configurations fixed, we can also elaborately arrange the object poses that are easy to reach, such that there is a large base region. In addition, an appropriate size of the tray is important, otherwise, with objects stored in wide cross-section trays, the resultant base region for grasping all the objects is small or even empty. Therefore, it is necessary to carefully select these parameters to improve the overall robustness and efficiency.

\section{Dynamically Update the Base Positions}
In the presented examples in section \MakeUppercase{\romannumeral 5}, since we assume the target to be all the objects in the tray, the planned base positions are feasible for the mobile manipulator to grasp \textbf{all} the objects in the tray, even though the number of objects decreases as the pick-and-place tasks proceed, the base positions remain valid no matter how many objects left in the tray. The assumption is not necessarily appropriate as the objects are picked away gradually, it is possible to dynamically update the base positions according to the remaining objects, such that the base region becomes larger with the decreasing remaining objects, and the robustness with respect to base positioning uncertainty is improved. In this section, we discuss the feasibility and performance of dynamically updating the base regions, which depends on whether the update can be performed in the online phase, and the feasibility of online execution is dominated by the online availability of base regions of the target objects. To analyze the online availability, we have to consider the object placement styles in the tray, including the following two situations: (1) The objects are regularly placed in the trays; (2) The objects are randomly placed in the trays, both of the situations are common in manufacturing environment.

\subsection{Objects Regularly Placed in the Trays}
\begin{figure}
    \centering
    \includegraphics[width=\columnwidth]{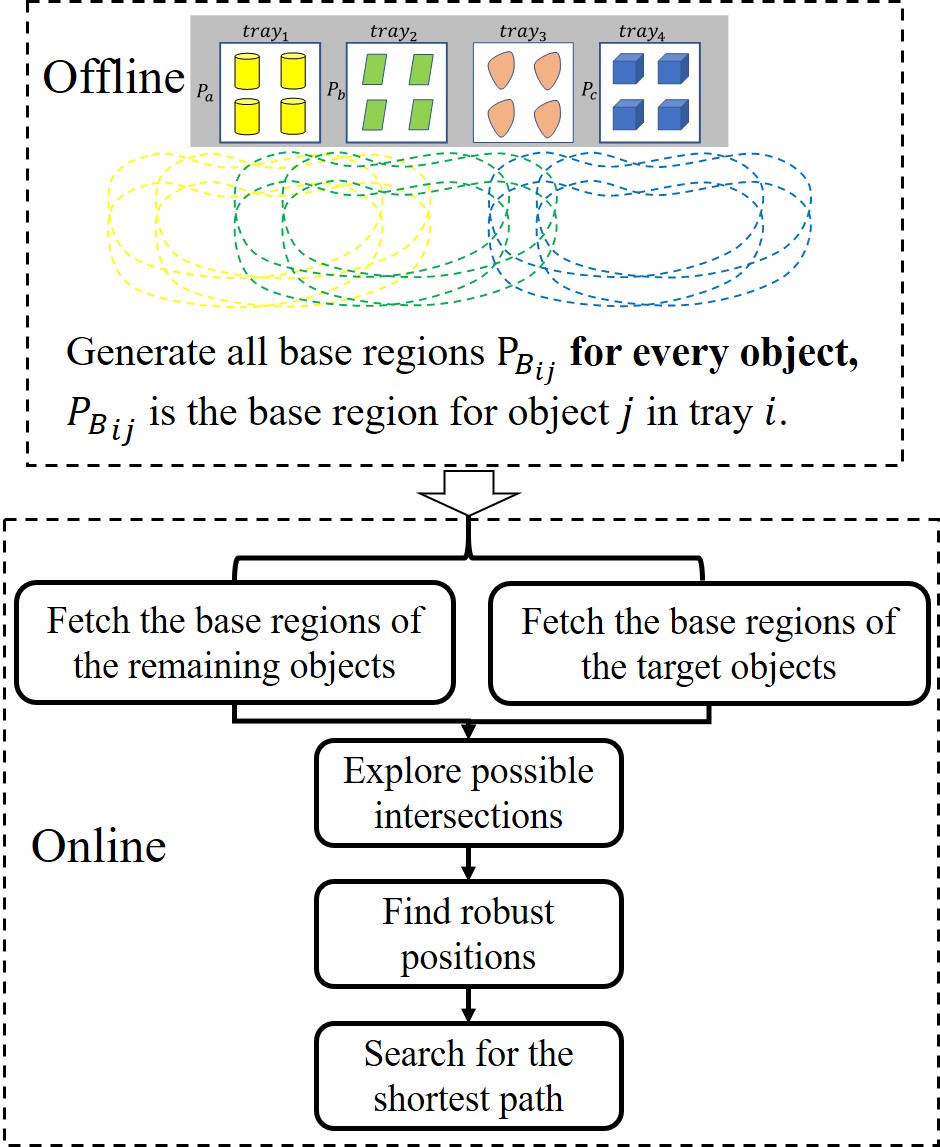}
    \caption{Workflow of two schemes for updating the base positions for picking regularly placed objects in the trays, two schemes differ in the base regions to be retrieved in the online phase.}
    \label{fig:workflow_regular}
\end{figure}

One policy for updating the base positions is based on the remaining objects, the update is performed after every round of pick-and-place task, using the current remaining objects in the tray as the target, intuitively, the size of the base region increases as the task proceeds. The workflow is described in Fig. \ref{fig:workflow_regular}, since the objects are regularly placed in the tray with known poses, the base region $P_{Bij}$ for object $\mathcal{O}_j$ in $tray_i$ can be calculated in the offline phase for all the objects. In the online phase, firstly we determine the remaining objects in the tray, by either remembering which object has been picked or using a camera to extract the configuration of remaining objects, then the base regions for the remaining objects can be retrieved from the offline database. The retrieved base regions are further processed to explore the possible intersections, iteratively find robust positions, as well as searching for the shortest path.

\begin{figure}
    \centering
    \includegraphics[width=0.7\columnwidth]{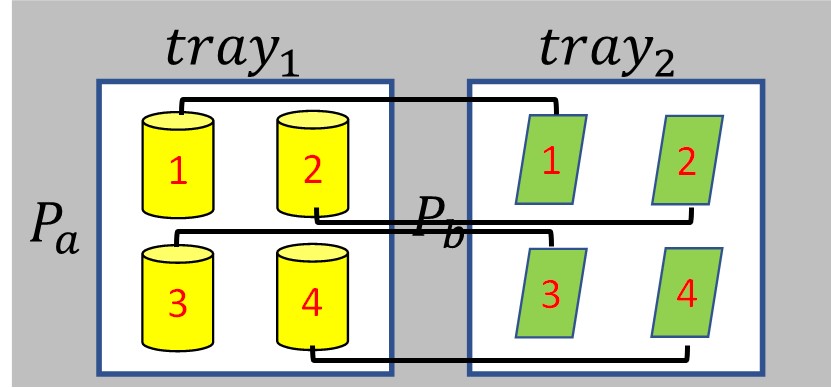}
    \caption{Pair the objects in neighboring trays and keep the distance between the objects in a pair nearly constant, a pair objects are labeled with the same number and connected by a black line. Picking objects following such pairing reduces the overall variance of the robustness in different rounds.}
    \label{fig:pick_order}
\end{figure}

Another policy for updating the base positions is based on the target objects to be picked. This is motivated by the scenario where the mobile manipulator is requested to pick up a certain amount of objects in every round of pick-and-place task, it is sufficient for the mobile manipulator to reach a certain number of objects in the target tray, and the base region is calculated from the target objects to be picked, instead of all the objects or all the remaining objects. For instance, if there are $m_a$ objects remaining in a tray with a specific configuration, in a round of pick-and-place task, $m_b$ objects, where $(m_b<m_a)$, should be picked from the tray. We can either exhaustively search for an optimal picking order of objects that achieves an efficient and robust sequence of base positions, or heuristically select objects from the tray to pick. Fig. \ref{fig:pick_order} illustrates a simple heuristic, firstly we pair the objects in neighboring trays, such that the distance between objects in every pair is nearly constant, then these pairs of objects take precedence to be picked when the mobile manipulator has to pick up objects from two neighboring trays. In every round of pick-and-place, since the distance between objects does not change much, so does the size of the intersection of their base regions, therefore, the robustness is more consistent.

Through dynamically re-planning the base positions for the remaining objects or the target objects to be picked, the robustness with respect to base positioning uncertainties is improved. The overall efficiency is also likely to be improved, since the base region becomes larger, it may result in more intersections or higher order intersections with other base regions.

\subsection{Objects Randomly Placed in the Trays}
If the objects are randomly placed in the tray, it becomes impractical to calculated the base region for every object, however, we can treat the objects in the tray as a whole and estimate the total grasps using the method described in section \MakeUppercase{\romannumeral 7}-B. Similarly, the grasps for situations where different amount of objects remain can be estimated. For instance, in the offline phase, we can estimate the base regions for the trays where there are 100\%, 75\%, 50\%, 25\% of the objects remain, during the online execution, the amount of remaining objects can be measured by a weighing device, then the base region with similar amount of remaining objects can be retrieved from the offline database. 

If the base region updates according to the target objects to be picked, since the base region for an individual object is not available, we have to select the target objects to be grasped and calculate the base regions online, which is time-consuming.

\subsection{Analysis of Different Schemes}
We use some numerical examples to evaluate the performance of different schemes. The calculation is implemented in C++ and runs on a laptop with Intel 2.5GHz processors and 16GB of RAM. Typical calculation for the sub-tasks are listed in Table \ref{tab:calculation_time}, all the sub-tasks except the calculation of base regions are feasible for online execution. Notice that the complexity of brute-force search for the shortest path is $O(n!)$, but the calculation can be done within one second if the base sequence size is less than 10 (excluding the start and goal positions).

\begin{table}[]
    \centering
    \begin{tabular}{c c}
    \hline
        Sub-tasks &  Calculation time (s) \\
    \hline
        Generate base regions for all objects & 288\\ 
        Explore intersections & 0.16 \\
        Find robust positions & 0.009\\
        Search for the short path & 0.6\\
    \hline
    \end{tabular}
    \caption{Typical calculation time for different sub-tasks. The result is conducted on 9 neighboring trays, where there are 12 objects in every tray, every object is provide with one grasp. Except the calculation of base regions, all the other tasks can be calculated online.}
    \label{tab:calculation_time}
\end{table}

\begin{table}
    \centering
    \begin{tabular}{c c c c c}
    \hline
    Schemes & Base region & Average robustness & Online & Feasibility \\ 
    \hline
    Reg/ALL & Reliable & 0.1 mm & OK & \textcolor{red}{Yes} \\
    Reg/updateR & Reliable & 0.155 mm & OK & \textcolor{red}{Yes} \\  
    Reg/updateT & Reliable &  0.186 mm & OK & \textcolor{red}{Yes} \\
    \hline
    Rand/ALL & Reliable & 0.1 mm & OK & \textcolor{red}{Yes} \\
    Rand/updateR & Unreliable & 0.143 mm & OK & No \\ 
    Rand/updateT & Reliable & 0.229 mm & No & No \\
    \hline
    \end{tabular}
    \caption{Comparison of different schemes. In the column of schemes, Reg and Rand are the abbreviations of regular and random placement, respectively. updateR and updateT represent update the base regions based on remaining objects and target objects to be picked, respectively. ALL represents all the objects that fill up a tray, respectively.}
    \label{tab:schemes_comparison}
\end{table}
For regularly placed objects, we assume there are 9 consecutive trays in a row, every tray contains 12 objects regularly placed at a 3 by 4 grid, every object is provided with one candidate grasp. The first scheme to calculate the base positions using the grasps of all the objects. For the other two schemes which update the base positions, we assume only one object is picked from a tray in every round of pick-and-place task, and after every round the base positions will be updated. The calculated base regions for all the three schemes are reliable since the grasps are the same as the grasps used in offline calculation, and they are all feasible for online execution since base regions are readily available from the offline database. The robustness of different schemes are evaluated by the average robustness of all the rounds of the tasks, $\sum_{i=1}^m(robustness_i)/m$, where $robustness_i$ denotes the average robustness of base positions in i-th round and $m$ is the total number of rounds. As a result, the schemes that update the base region according to the target objects to be picked has the highest average robustness score.


For randomly placed objects, it is possible to estimate all the grasps of all the objects that fill up a tray, and perform the offline calculation of the globally static base positions, which are valid for different rounds of pick-and-place tasks, and the mobile manipulator can pick any object from the tray in the online phase. However, updating the base positions for randomly placed objects is difficult. In the case of updating based on the remaining objects, the base region with a similar amount of remaining objects is retrieved from the offline database, but the retrieved base region is not reliable. Because the base regions calculated offline assume the randomness of the poses of the placed objects, but the actual picking is usually not performing randomly in terms of the poses of the remaining objects, instead the robot picks according to some metrics, such as grasp quality metrics, which may favor some specific poses. Therefore, the poses of the remaining objects are not guaranteed to be random, in another word, there are some discrepancies between the actual base regions and the offline-generated base regions. Furthermore, if the remaining objects are assumed to be randomly distributed in the tray, then the base regions for the tray with different amount of remaining objects are theoretically the same, in this sense, the base region is not necessary to be updated.

On the other hand, updating the base region according to the target objects cannot be implemented online. Because the offline-generated base regions are estimated by treating the objects as a whole, while the base region for an individual object is not available, thus the base regions for the target objects have to be calculated online. However, Table \ref{tab:calculation_time} shows that the calculation of base region is the most time-consuming sub-task, which involves many IK queries and collision checks, it takes about 3 seconds to obtained the base region of one object labeled with only one grasp, and the total calculation time grows linearly with respect to the number of objects and grasps. Therefore, dynamically update the base region for target objects to be picked is not practical for randomly placed objects.

Table \ref{tab:schemes_comparison} summarizes all the 6 schemes, from the above analysis, 4 of them are feasible for practical application. For randomly placed objects, we conclude that a globally static sequence of base positions should be used, without further update. For regularly placed objects, both static and dynamically updated base sequences are feasible. Updating the base positions improves the overall robustness, and the update scheme based on the target objects to be picked has the highest average robustness score, however, one of the disadvantages is that, if the picking fails, the robot has to re-plan and move to another position to try the picking once again. Furthermore, updating the base positions cannot be completed until the mobile manipulator finishes one round of the task, this obstructs the efficient use of multiple mobile manipulators, one has to wait until another mobile manipulator finishes a round of pick-and-place task, and then update the base positions and perform the pick-and-place using the updated base positions. But for the globally static base sequence calculated from all the objects, multiple mobile manipulators can be involved in the tasks efficiently, for instance, when a mobile manipulator finishes picking objects from $tray_1$ and $tray_2$ and is ready to move to the next base position, another mobile manipulator can immediately move to pick objects from $tray_1$ and $tray_2$. Therefore, despite the overall robustness is outperformed by the schemes that update the base positions, it still makes sense to use the offline planning schemes without further update.

\section{Experiments}
We present three sets of experiment to demonstrate the 4 feasible schemes, which cover different object placement styles and whether the base positions update or not. The Fetch robot \cite{wise2016fetch}, a single arm mobile manipulator equipped with a parallel-jaw gripper, is used to pick objects from multiple trays, there is a Primesense Carmine 1.09 short-range RGBD sensor mounted on the head of the robot. ROS \cite{quigley2009ros} navigation packages and Moveit! are used to plan and control the motion of the robot. The size of the tray used to store objects is 0.4 m by 0.3 m by 0.1 m. The recorded videos of all the experiments are provided in the supplementary material.

\begin{figure}
    \centering
    \begin{subfigure}[h]{0.48\columnwidth}
        \centering
        \includegraphics[width=\textwidth]{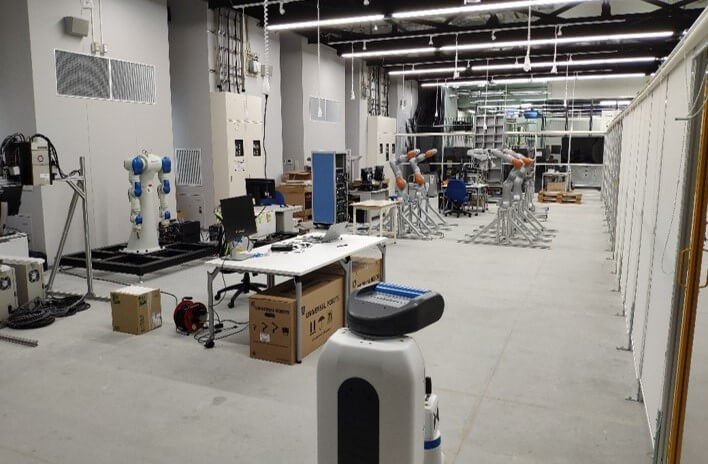}
    \end{subfigure}
    \begin{subfigure}[h]{0.48\columnwidth}
        \centering
        \includegraphics[width=\textwidth]{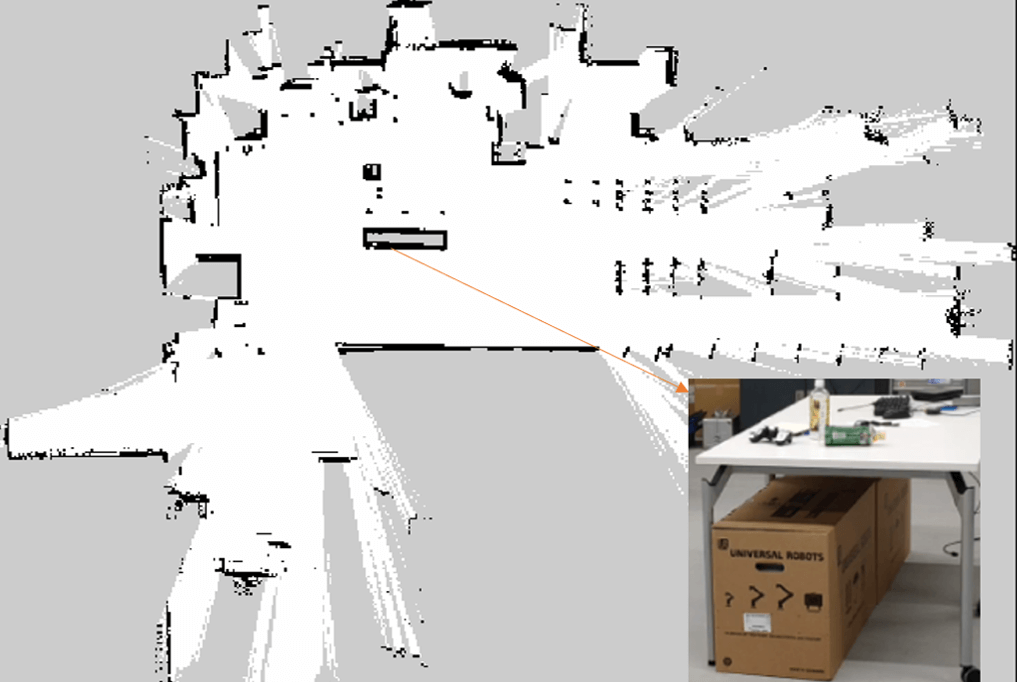}
    \end{subfigure}
    \begin{subfigure}[h]{0.32\columnwidth}
    \centering
    \includegraphics[width=\textwidth]{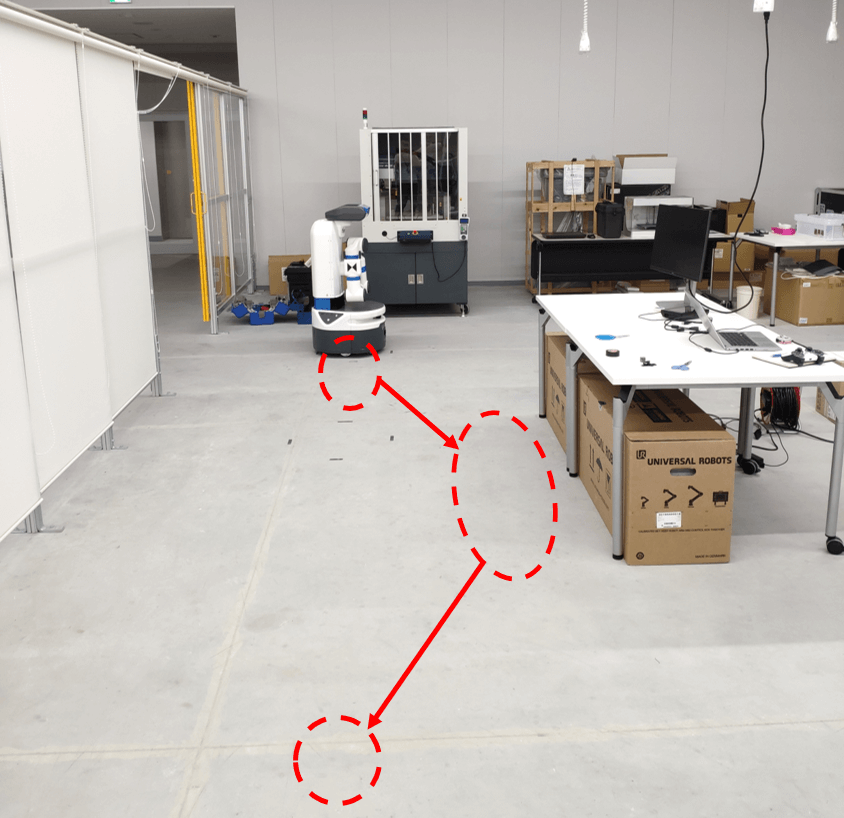}
    \end{subfigure}
    \begin{subfigure}[h]{0.65\columnwidth}
    \centering
    \includegraphics[width=\textwidth]{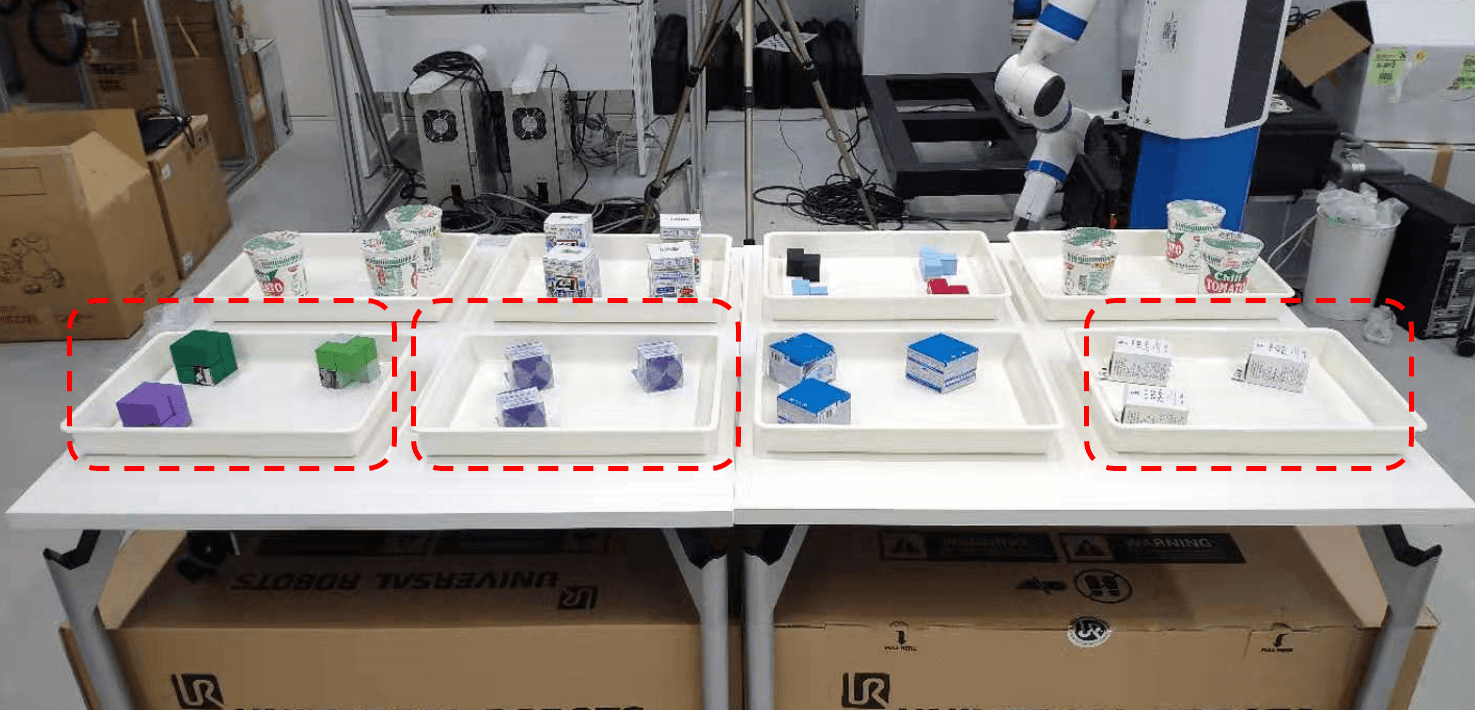}
    \end{subfigure}
    \caption{Experiment setup: (Top left) Indoor experimental environment. (Top right) A 2D map build by the laser scanner. (Bottom left) Task overview. (Bottom right) Target objects and their poses in three trays.}
    \label{fig:experiment_setup}
\end{figure}

\subsection{Regularly Placed, Globally Static Base Sequence}
The mobile manipulator navigates in an indoor environment as shown in Fig. \ref{fig:experiment_setup}, it starts from a predefined position in the environment and moves to pick up 3 objects stored in 3 different trays (as circled by red dashed line), whose locations in the environment are known, then the mobile manipulator carries the collected objects to the goal position. The base regions and intersections are calculated by the proposed method, and the results are already presented in Fig. \ref{fig:one_bin_result_IKRDB} and Fig. \ref{fig:IKRDB_intersections}. In order to obtain a robust base sequence, the base positioning uncertainty and repeatability are tested by looping the mobile manipulator between two fixed positions, the actually arrived positions are observed to deviate about 10 cm in average from the planned positions. Since the base positioning error is the result of map accuracy, sensor accuracy, environmental complexity and the performance of the mechanical system, with so many factors involved, it is assumed to be random and homogeneous in all directions, therefore, the base positioning uncertainty level $\bar{\sigma}$ is set as 10 cm, then a sequence of base positions can be planned by algorithm described in section \MakeUppercase{\romannumeral 5}-E. As a result, the mobile manipulator should successively move to the center of $P_{B1} \cap P_{B2}$ and $P_{B4}$ to collect all the required parts. 

When the Fetch robot moves to the calculated position, its head-mounted camera points to a rough direction such that the target tray is in the scope of the camera. We remove the point cloud segments of the table and tray to get objects' point cloud, for the simple box-shaped objects used in this experiment, the remaining point cloud is fitted with cuboids, such that the object pose can be determined, then the grasping poses are retrieved from the offline planned grasps. A more straightforward way which is used in \MakeUppercase{\romannumeral 7}-C is to attach a marker in front of the tray, then the object poses, as well as the grasps, with respect to the robot are easily obtained.

\begin{figure}
    \centering
    \begin{subfigure}[h]{0.15\textwidth}
    \centering
    \includegraphics[width=\textwidth]{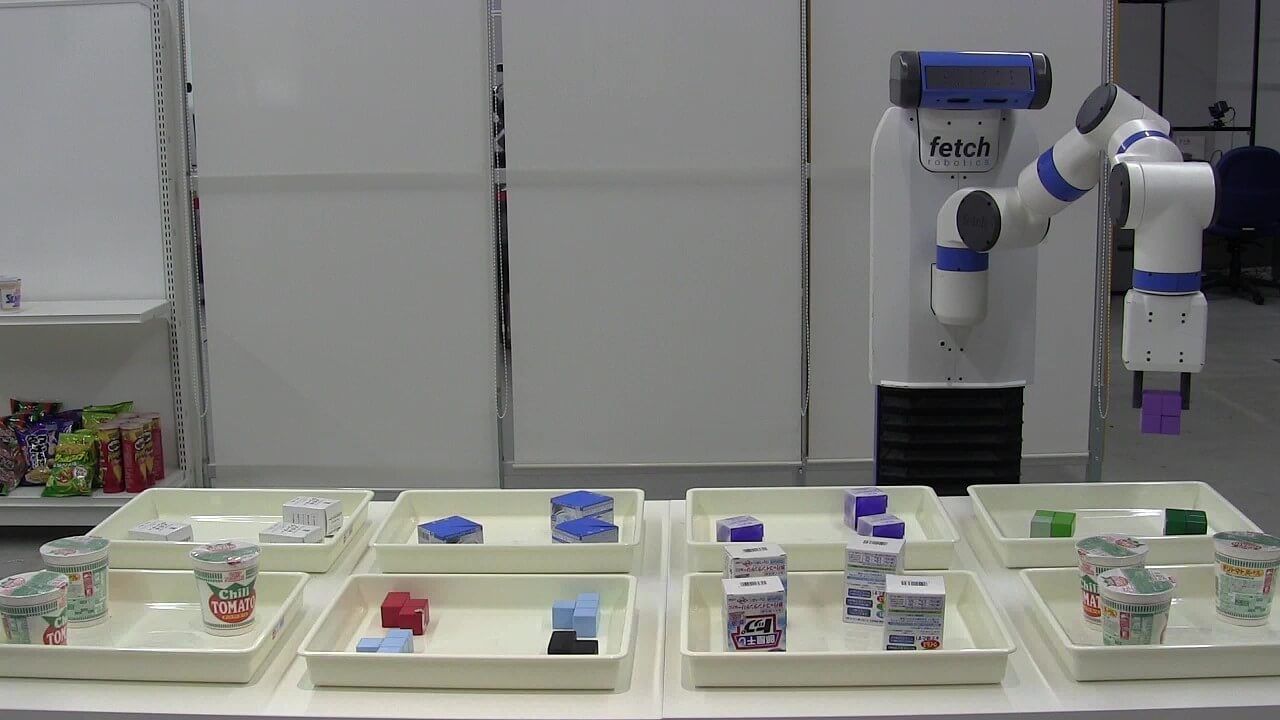}
    \end{subfigure}
    \begin{subfigure}[h]{0.15\textwidth}
    \centering
    \includegraphics[width=\textwidth]{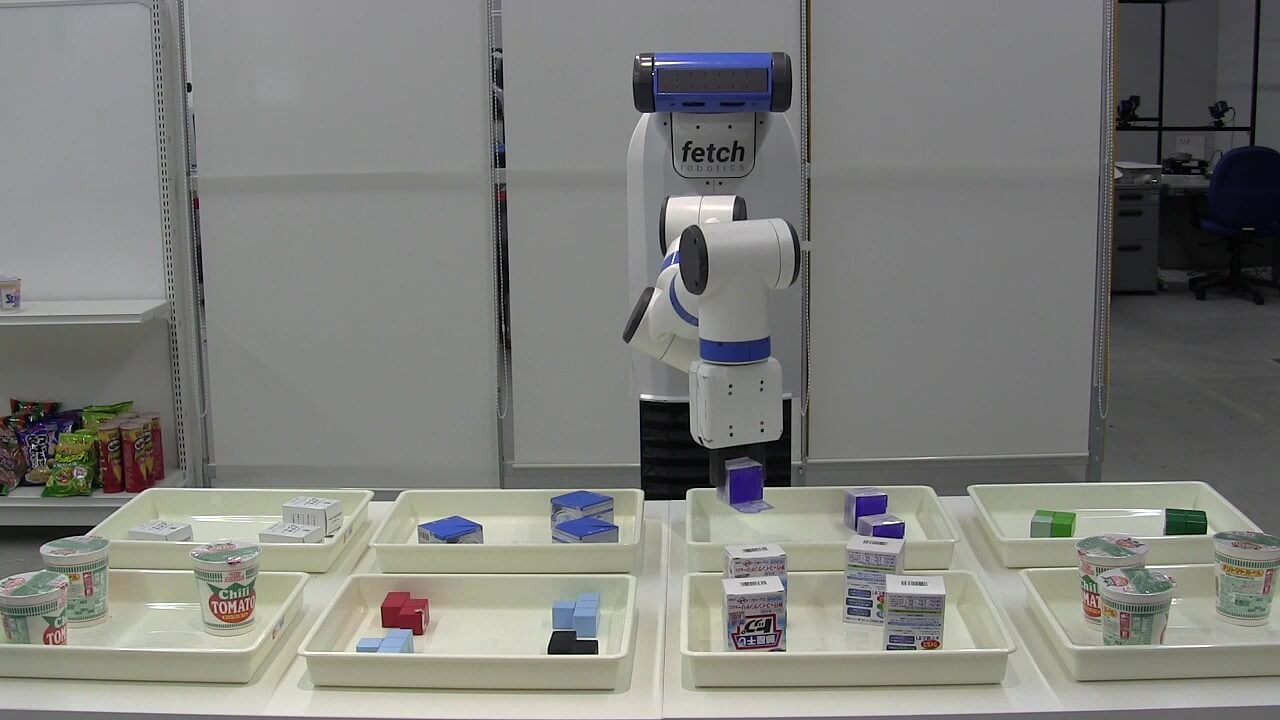}
    \end{subfigure}
    \begin{subfigure}[h]{0.15\textwidth}
    \centering
    \includegraphics[width=\textwidth]{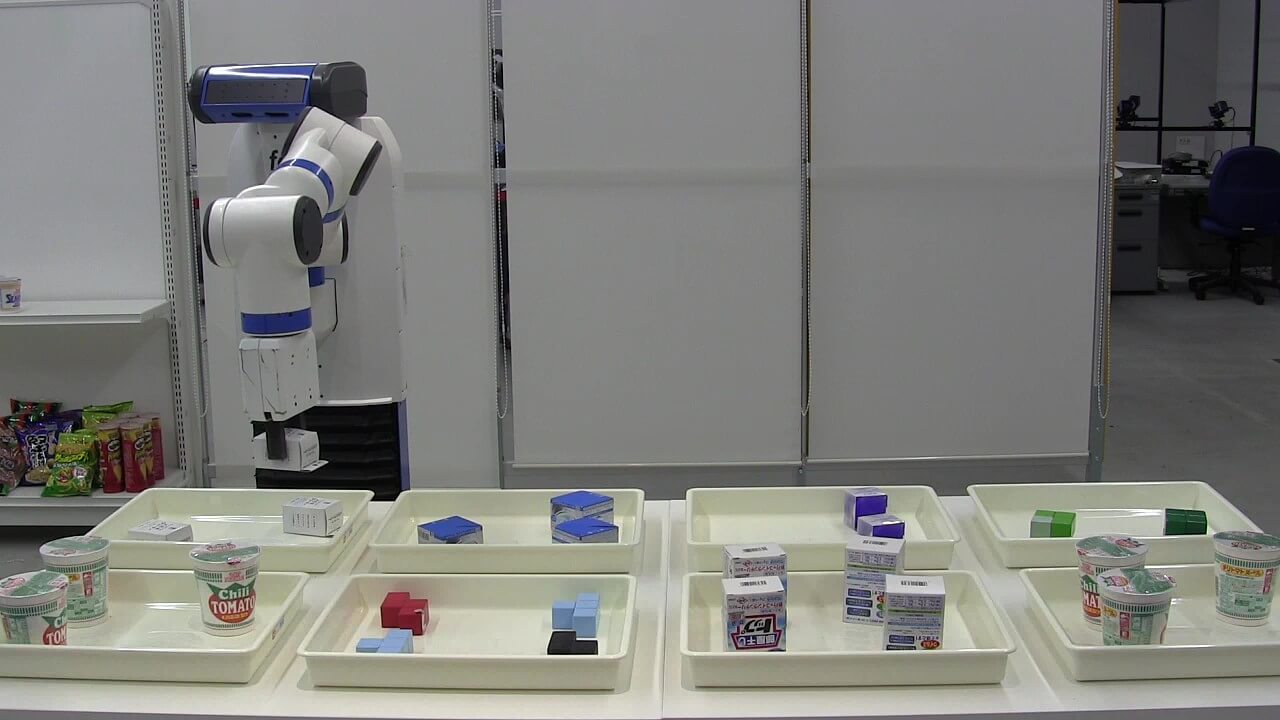}
    \end{subfigure}
    \begin{subfigure}[h]{0.15\textwidth}
    \centering
    \includegraphics[width=\textwidth]{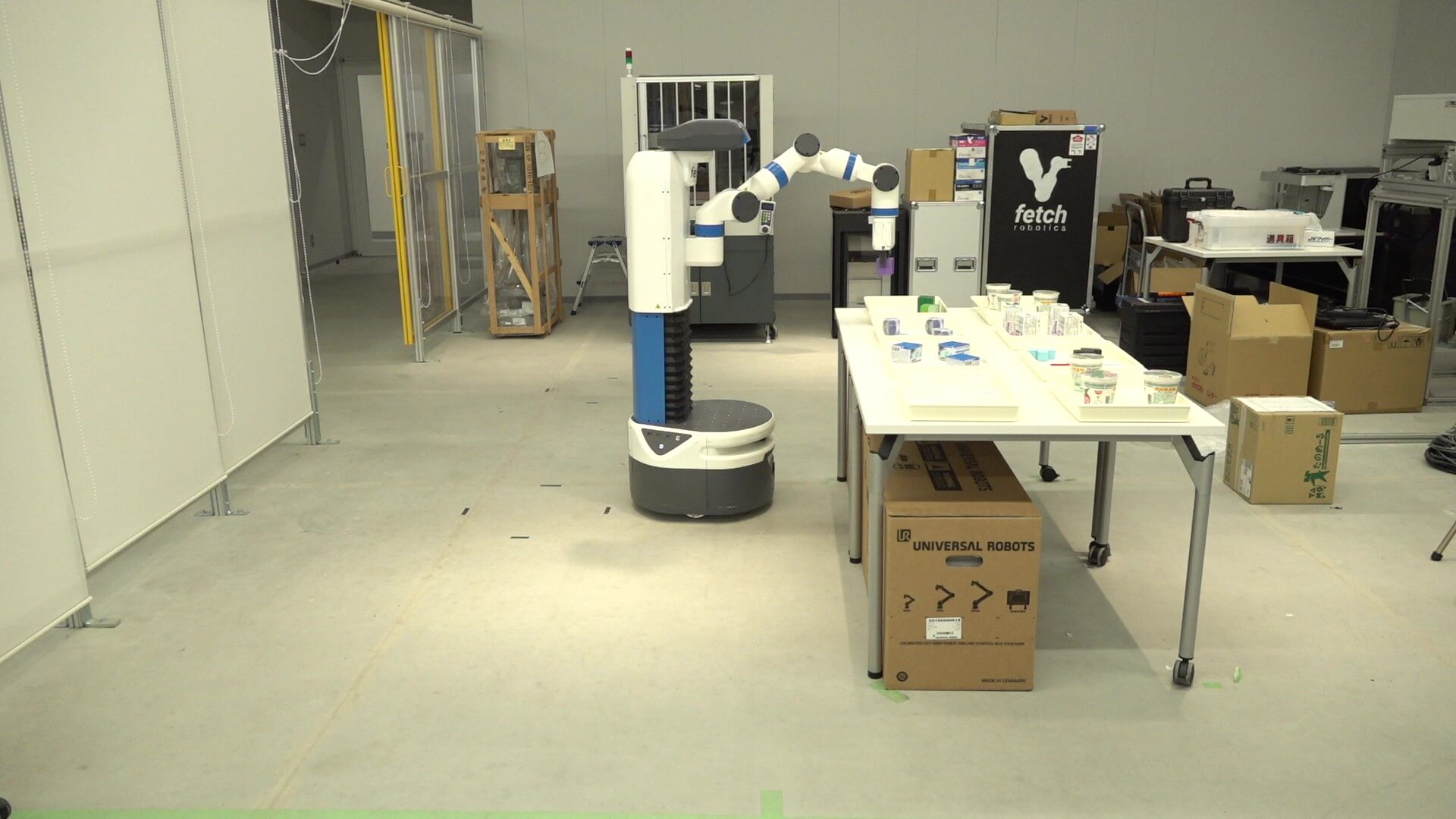}
    \end{subfigure}
    \begin{subfigure}[h]{0.15\textwidth}
    \centering
    \includegraphics[width=\textwidth]{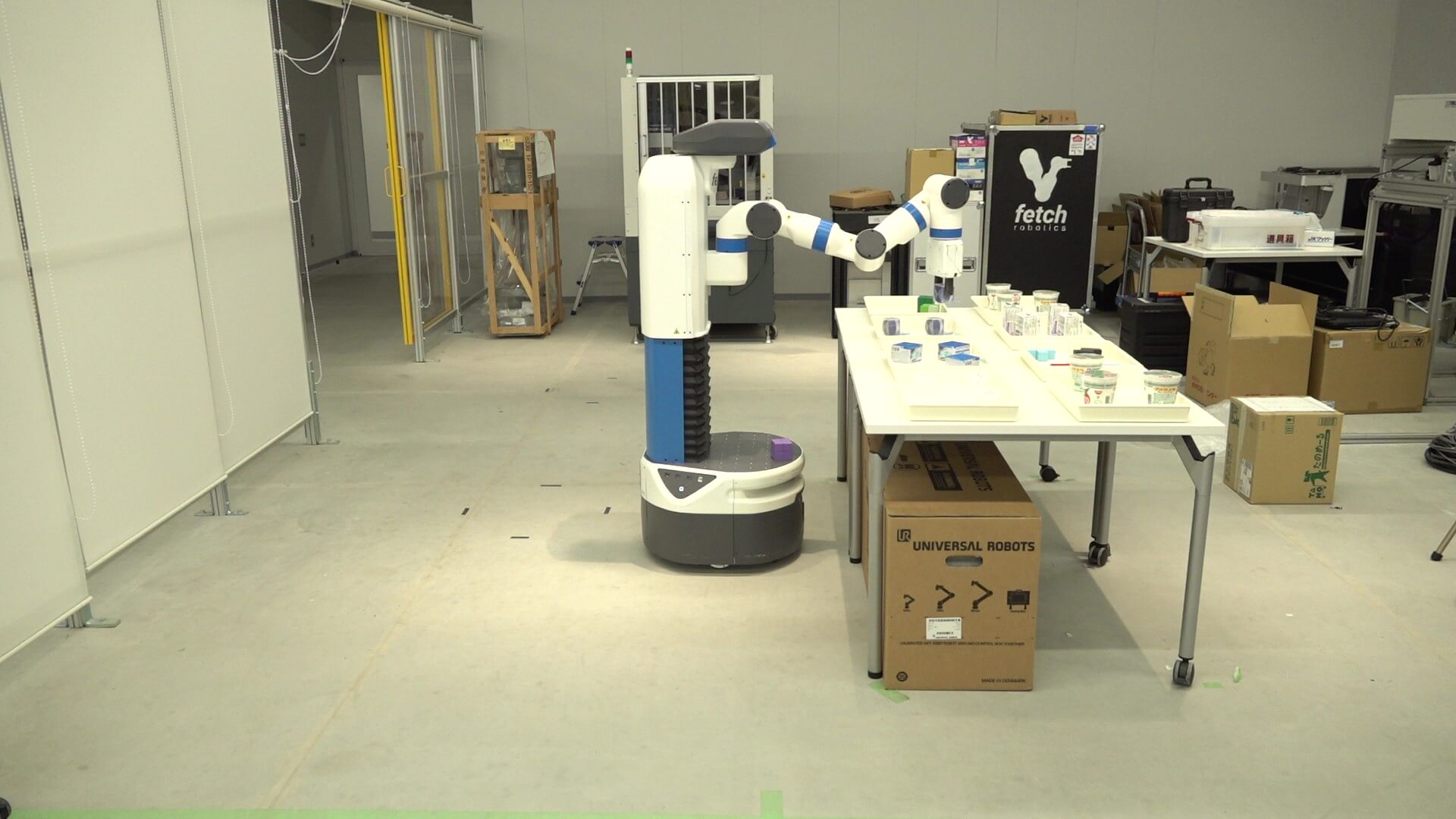}
    \end{subfigure}
    \begin{subfigure}[h]{0.15\textwidth}
    \centering
    \includegraphics[width=\textwidth]{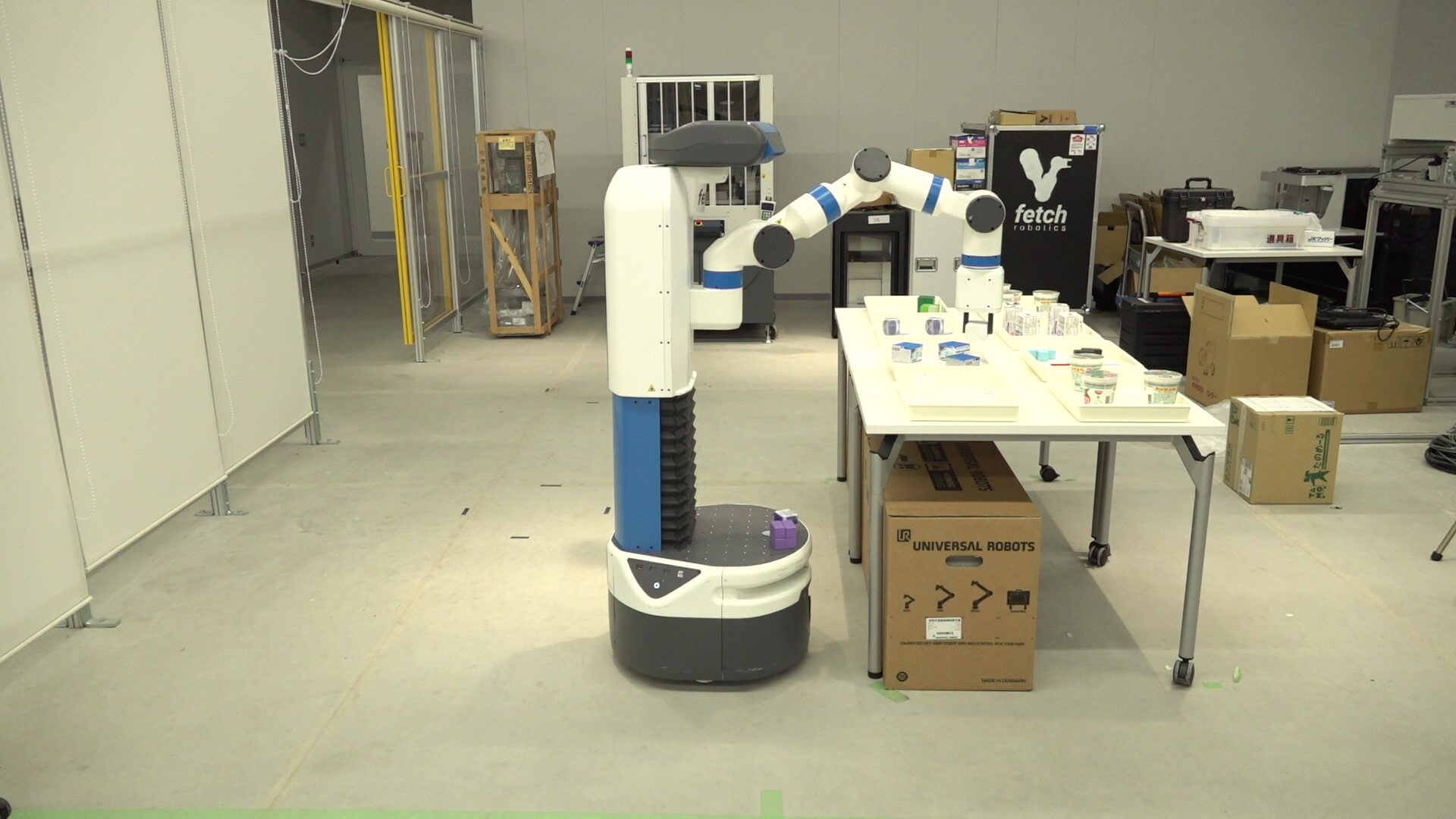}
    \end{subfigure}
    \caption{(Left) Move to $P_{B1}$ to pick up a part from $tray_1$. (Middle) Move to $P_{B2}$ to pick up a part from $tray_2$. (Right) Move to $P_{B4}$ to pick up a part from $tray_4$.}
    \label{fig:exper1}
\end{figure}
\begin{figure}[t]
    \centering
    \begin{subfigure}[h]{0.15\textwidth}
    \centering
    \includegraphics[width=\textwidth]{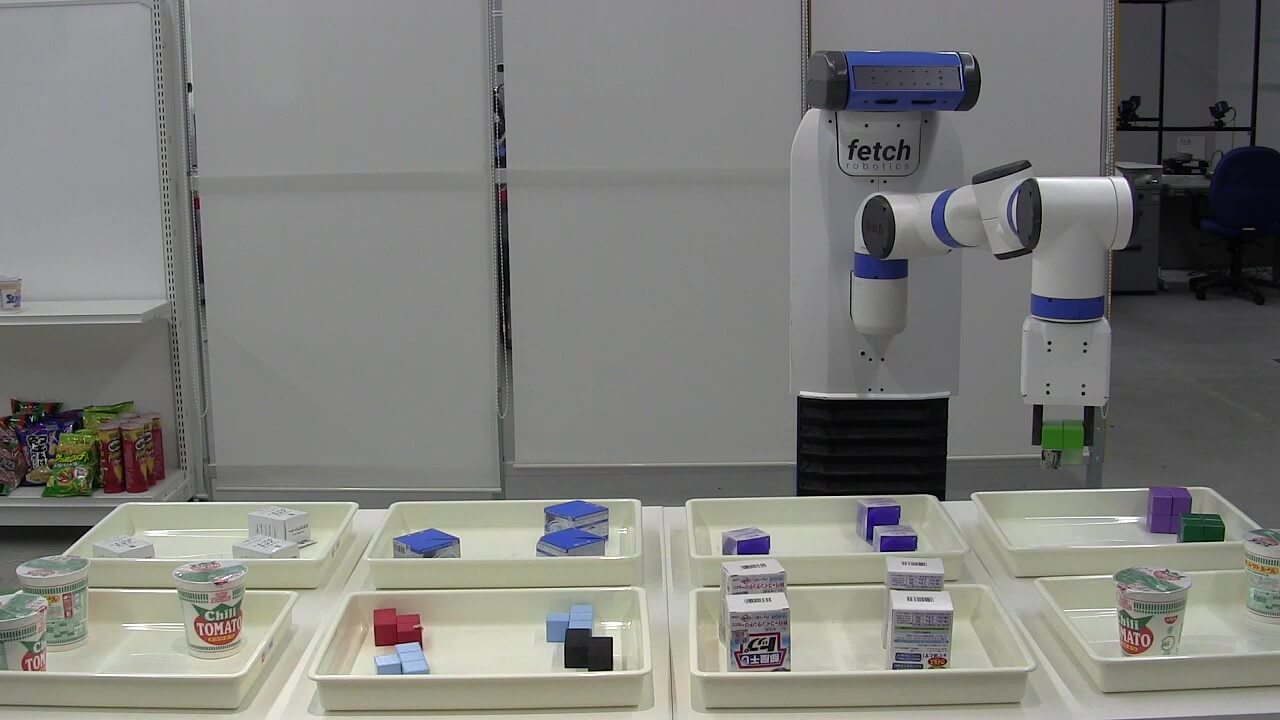}
    \end{subfigure}
    \begin{subfigure}[h]{0.15\textwidth}
    \centering
    \includegraphics[width=\textwidth]{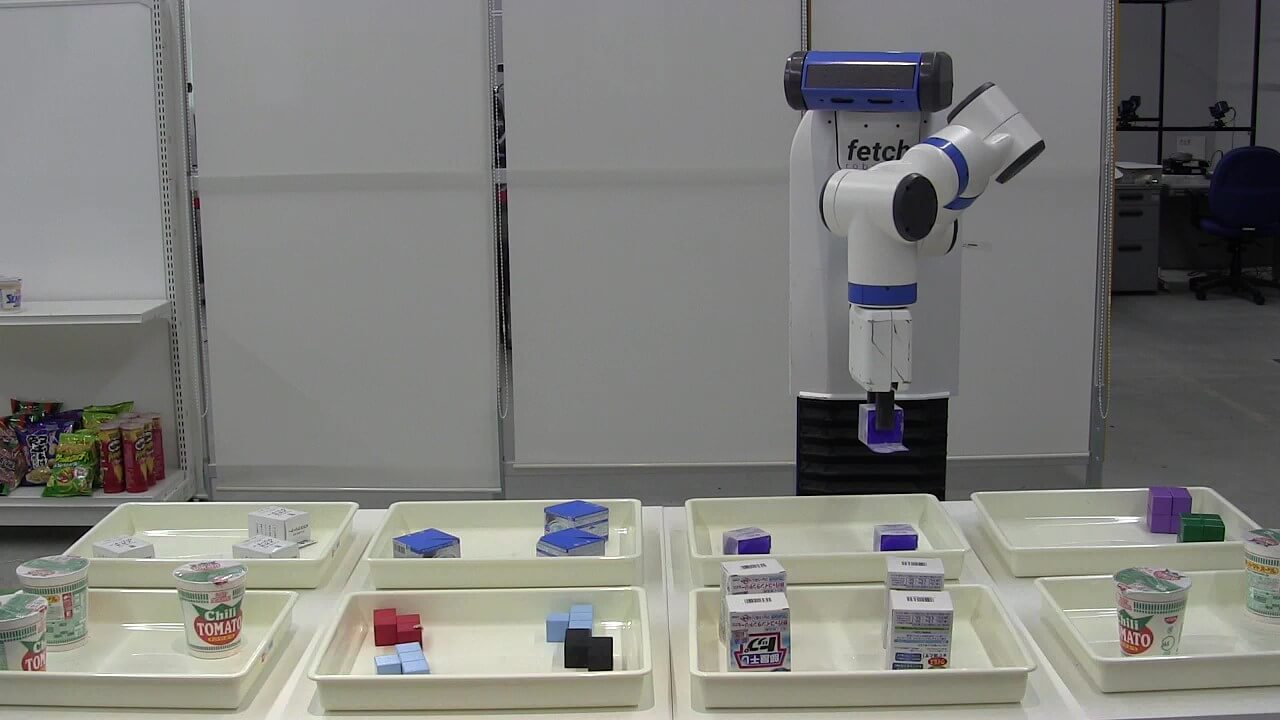}
    \end{subfigure}
    \begin{subfigure}[h]{0.15\textwidth}
    \centering
    \includegraphics[width=\textwidth]{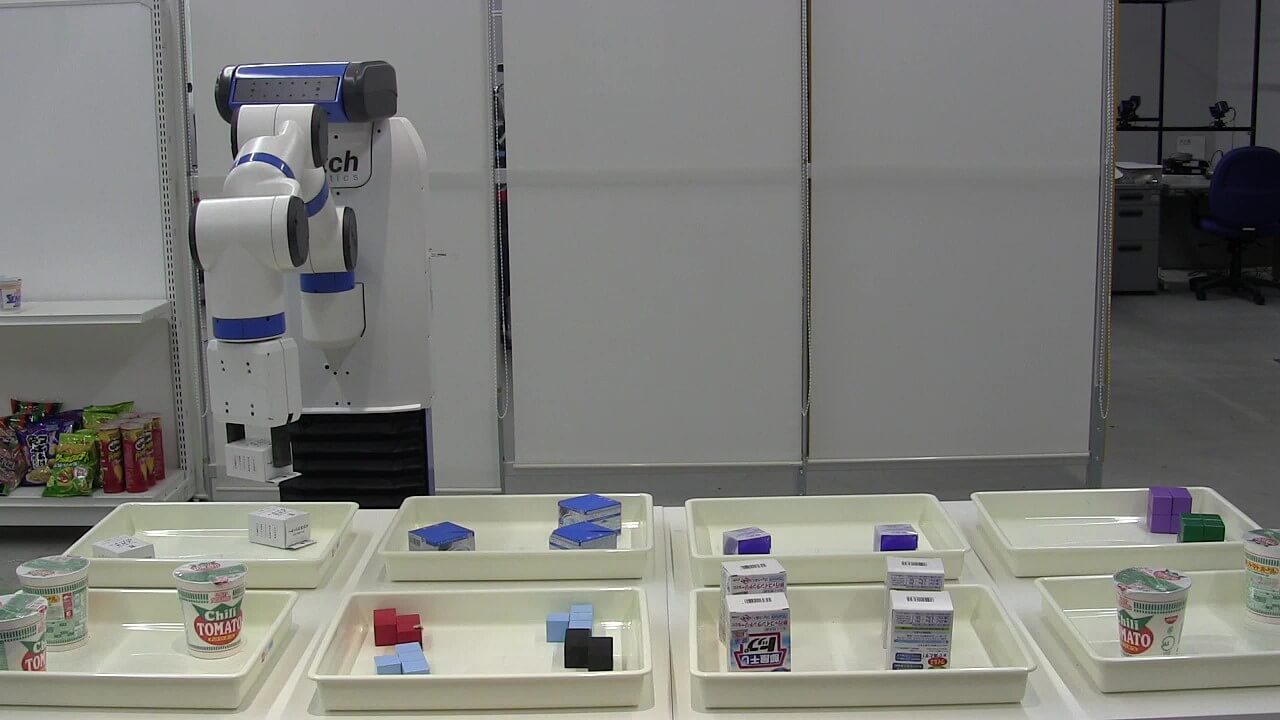}
    \end{subfigure}
    \begin{subfigure}[h]{0.15\textwidth}
    \centering
    \includegraphics[width=\textwidth]{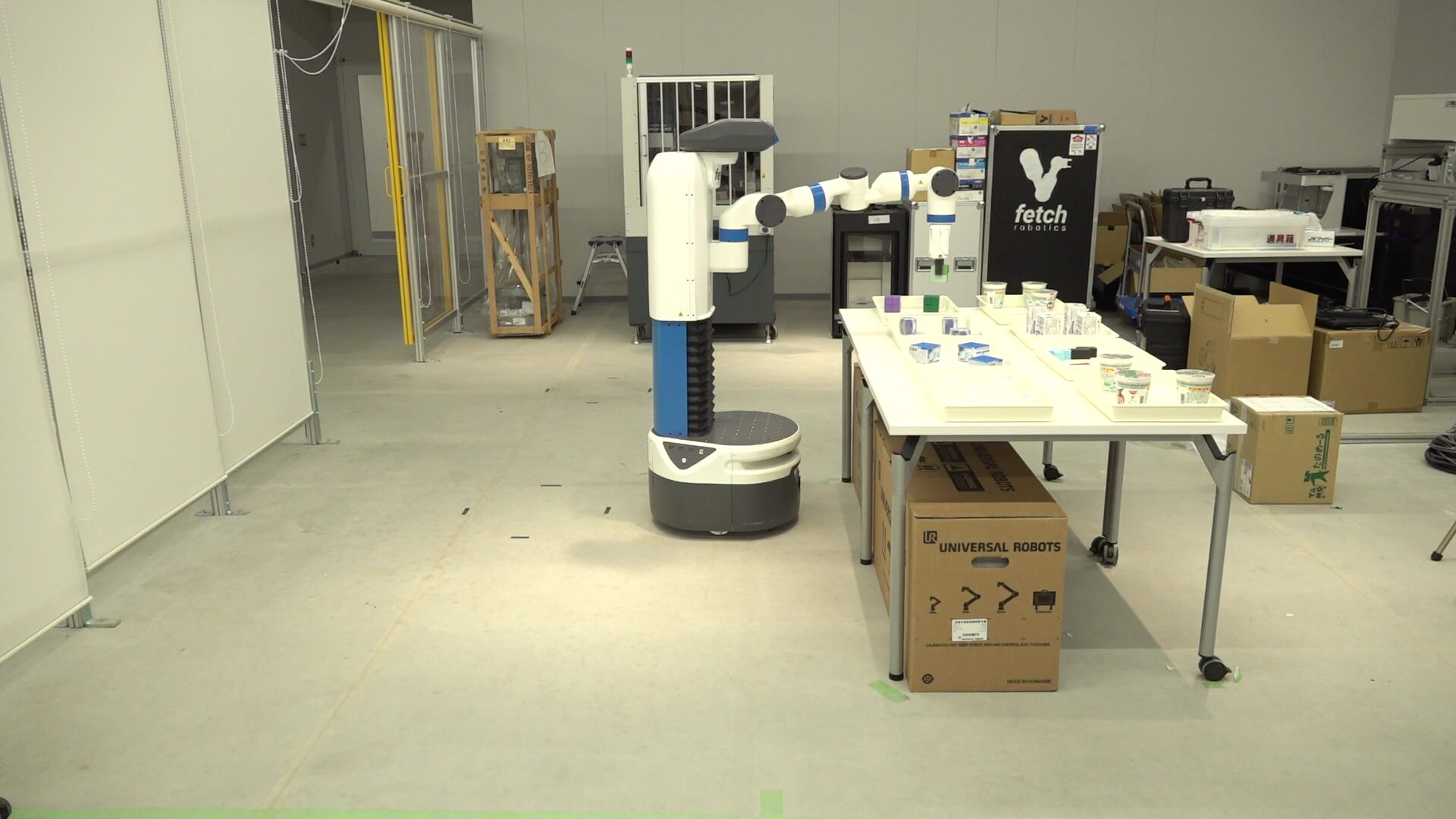}
    \end{subfigure}
    \begin{subfigure}[h]{0.15\textwidth}
    \centering
    \includegraphics[width=\textwidth]{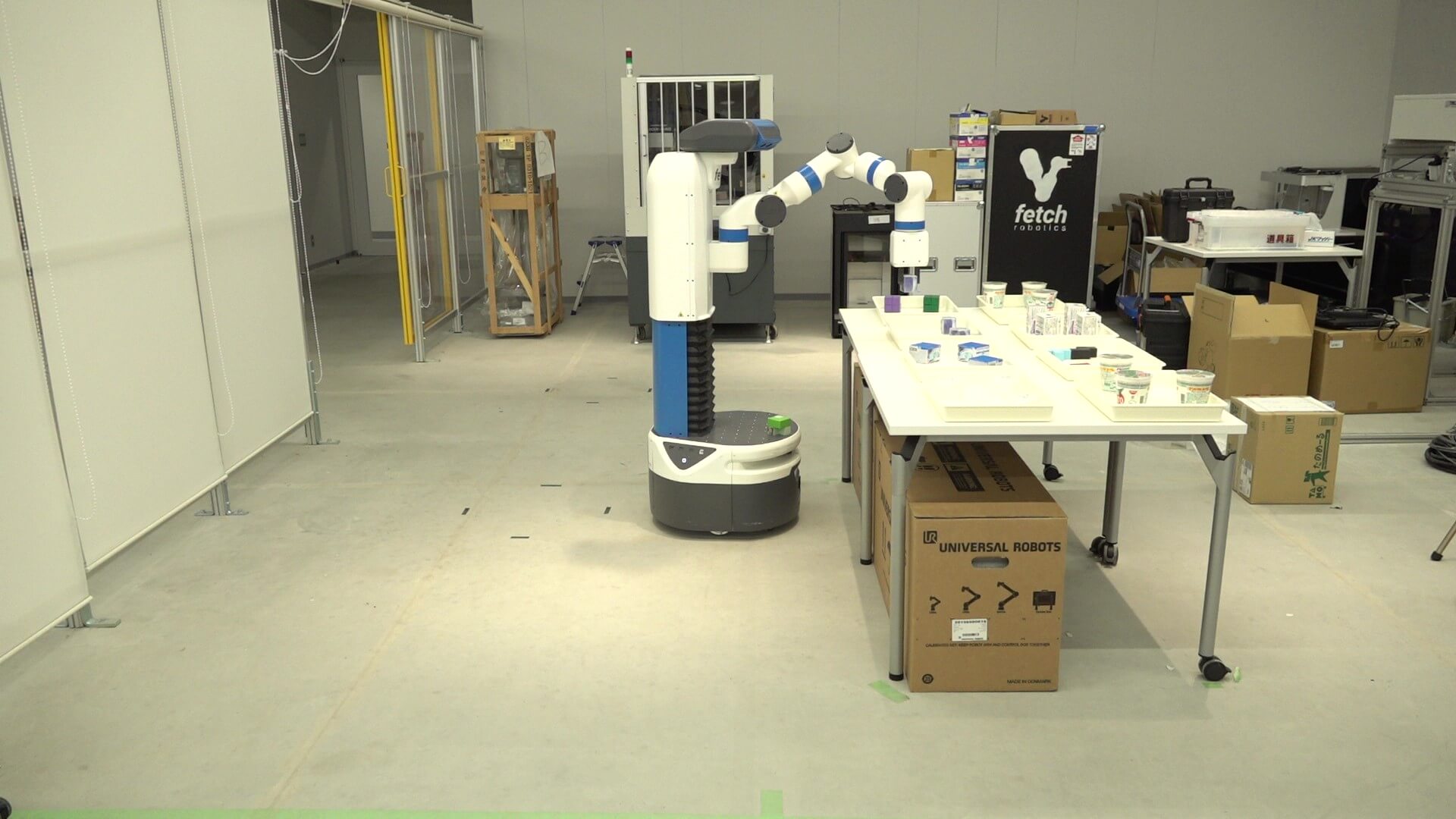}
    \end{subfigure}
    \begin{subfigure}[h]{0.15\textwidth}
    \centering
    \includegraphics[width=\textwidth]{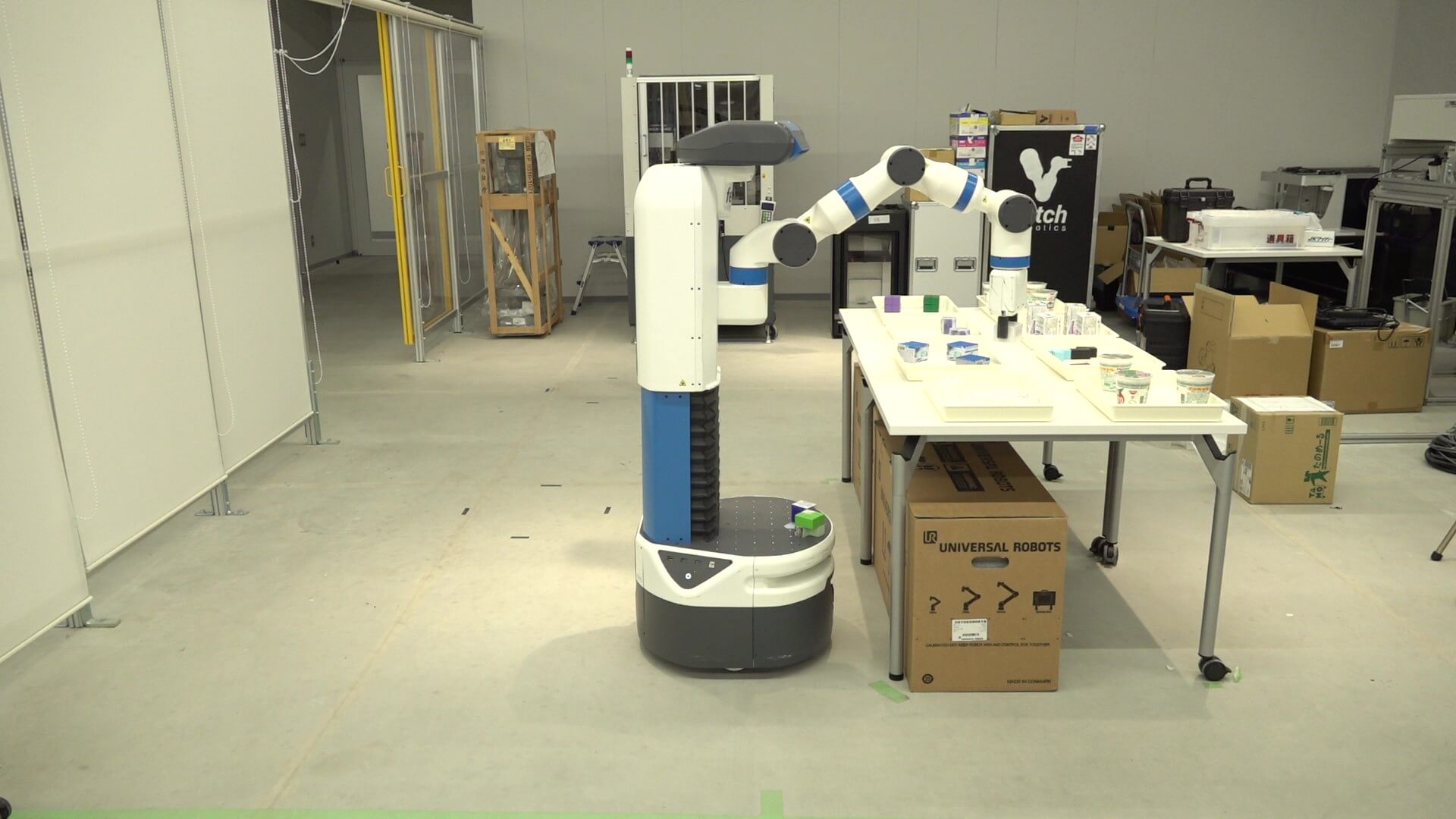}
    \end{subfigure}
    \caption{(Left and Middle) Move to $P_{B1} \cap P_{B2}$ to pick up parts from $tray_1$ and $tray_2$. (Right) Move to $P_{B4}$ to pick up a part from $tray_4$.}
    \label{fig:exper2}
\end{figure}
For comparison, we move the robot to the center of $P_{B1}, P_{B2}$ and $P_{B4}$ to collect the parts from three trays, respectively. As shown in Fig. \ref{fig:exper1}, in each base position, the mobile manipulator picks up one object from the associated tray. Fig. \ref{fig:exper2} shows the robot motion following the planned base sequence, the mobile manipulator moves to the center of $P_{B1} \cap P_{B2}$ to pick up parts from $tray_1$ and $tray_2$, then moves to the center of $P_{B4}$ to pick up part from $tray_4$. This experiment shows that the total operation time is reduced by 17 seconds due to reduced one base movement, the efficiency of the proposed method becomes more significant when the number of trays get larger.

\subsection{Randomly Placed, Globally Static Base Sequence}
\begin{figure}
    \centering
    \includegraphics[width=\columnwidth]{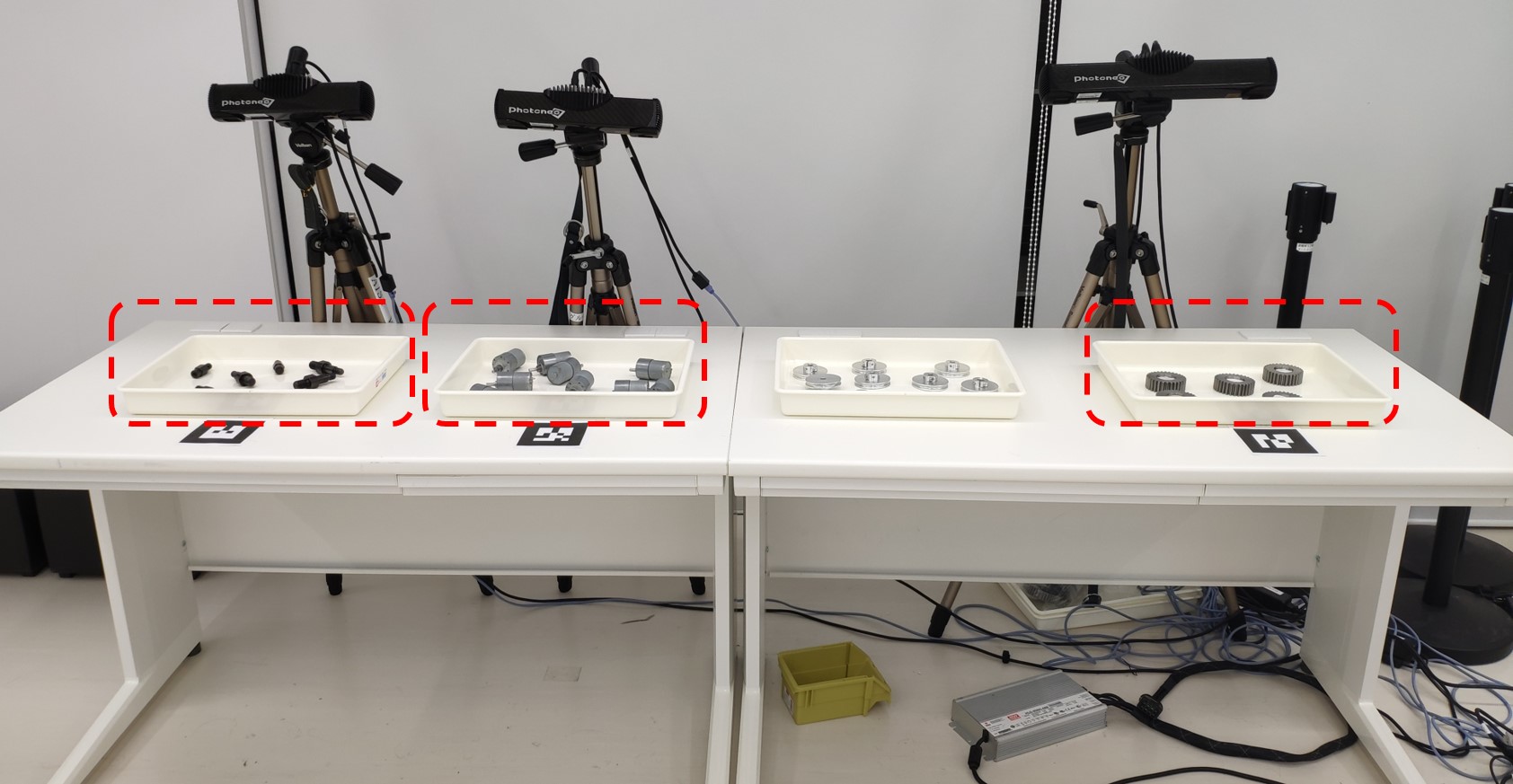}
    \caption{Experiment setup: The mechanical components to be grasped are marked with red dashed lines, PhoXi scanners are configured above the target components, the markers in front of the tray are for obtaining the transform from the tray to the mobile manipulator.}
    \label{fig:experiment2_overview}
\end{figure}

\begin{figure}
    \centering
    \begin{subfigure}[h]{0.28\columnwidth}
    \centering
    \includegraphics[width=\textwidth]{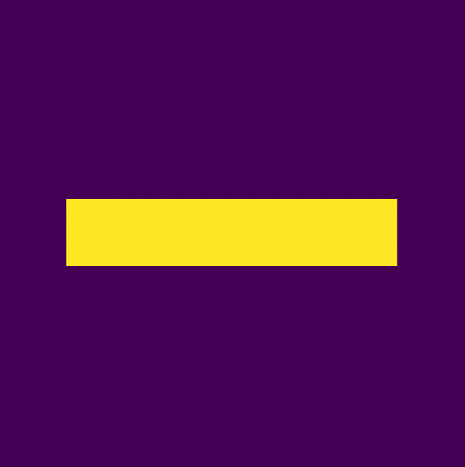}
    \caption{}
    \label{contact_region}
    \end{subfigure}
    \begin{subfigure}[h]{0.28\columnwidth}
    \centering
    \includegraphics[width=\textwidth]{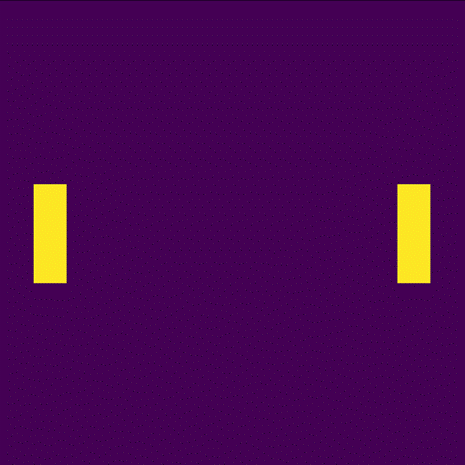}
    \caption{}
    \label{collision_region}
    \end{subfigure}
    \begin{subfigure}[h]{0.38\columnwidth}
    \centering
    \includegraphics[width=\textwidth]{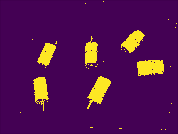}
    \caption{}
    \end{subfigure}
    \begin{subfigure}[h]{0.4\columnwidth}
    \centering
    \includegraphics[width=\textwidth]{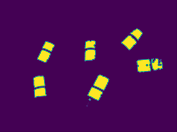}
    \caption{}
    \end{subfigure}
    \begin{subfigure}[h]{0.4\columnwidth}
    \centering
    \includegraphics[width=\textwidth]{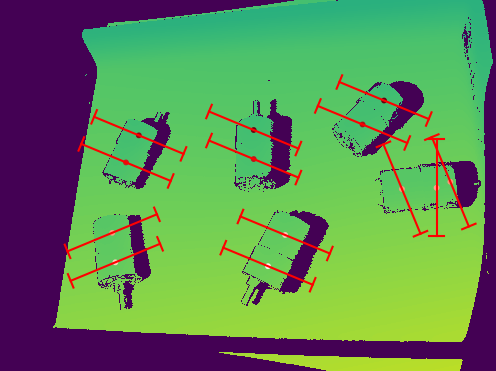}
    \caption{}
    \end{subfigure}
    \caption{(a) Contact region of the gripper model. (b) Collision region of the gripper model. (c) Mask image of the object. (d) Graspability map. (e) Planned grasps.}
    \label{fig:graspability_calculation}
\end{figure}

In the manufacturing environment, the mechanical components are often randomly placed in the tray. In this experiment, we use a mobile manipulator to pick up multiple mechanical components randomly placed in different trays, the overall experimental configuration is shown in Fig. \ref{fig:experiment2_overview}. Compared to the first experiment, we used a different grasp planning method for the mechanical components, which usually have complex shapes and reflective surfaces. The head-mounted camera on the Fetch robot can not capture high-quality point cloud of the object, instead we use fixed PhoXi 3D scanners to scan the mechanical components to obtain depth images. Fast graspability evaluation \cite{domae2014fast} is used to plan grasps from a single depth image, the gripper is represented by two mask images as shown in Fig. \ref{fig:graspability_calculation}a and b, Fig. \ref{fig:graspability_calculation}a represents the contact region where the gripper should contact the object and Fig. \ref{fig:graspability_calculation}b represents the collision region where the gripper should avoid collision with the environment. The mask image of contact region is used to convolve with the object's mask image to find the centroids of grasps, and the mask image of collision region is applied to find collision-free orientations around the centroid normal.

\begin{figure}
    \centering
    \begin{subfigure}{0.23\textwidth}
    \centering
    \includegraphics[width=\textwidth]{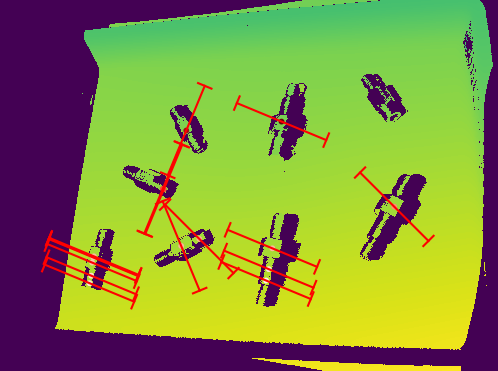}
    \caption{}
    \end{subfigure}
    \begin{subfigure}{0.23\textwidth}
    \centering
    \includegraphics[width=\textwidth]{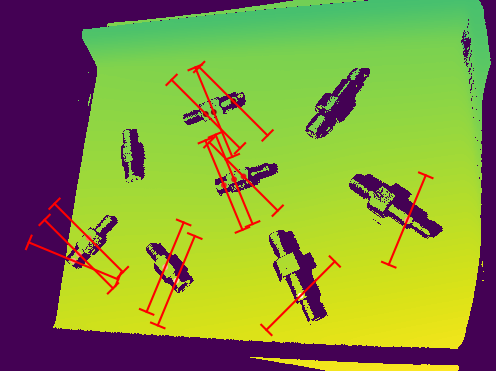}
    \caption{}
    \end{subfigure}
    \begin{subfigure}{0.23\textwidth}
    \centering
    \includegraphics[width=\textwidth]{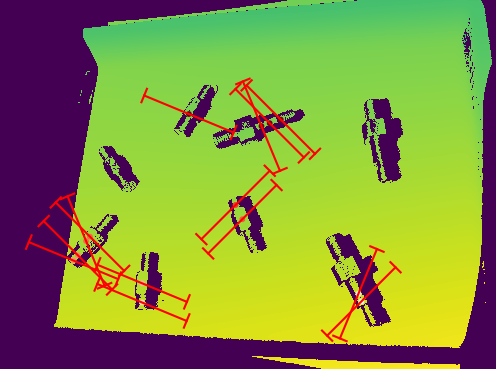}
    \caption{}
    \end{subfigure}
    \begin{subfigure}{0.23\textwidth}
    \centering
    \includegraphics[width=\textwidth]{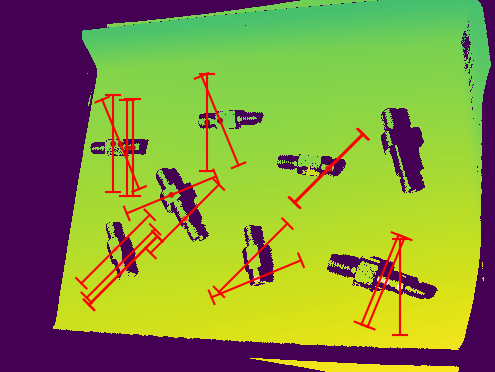}
    \caption{}
    \end{subfigure}
    \caption{Planned grasps at different object configurations.}
    \label{fig:grasps_part1}
\end{figure}

\begin{figure}
    \centering
    \begin{subfigure}{0.23\textwidth}
    \centering
    \includegraphics[width=\textwidth]{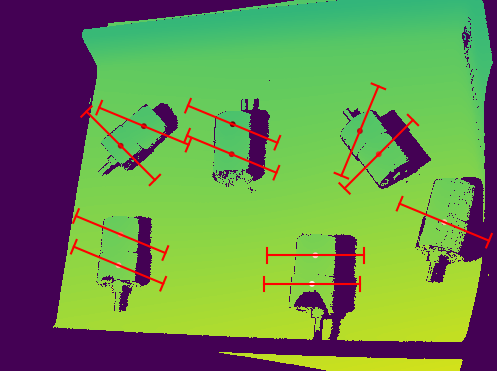}
    \caption{}
    \end{subfigure}
    \begin{subfigure}{0.23\textwidth}
    \centering
    \includegraphics[width=\textwidth]{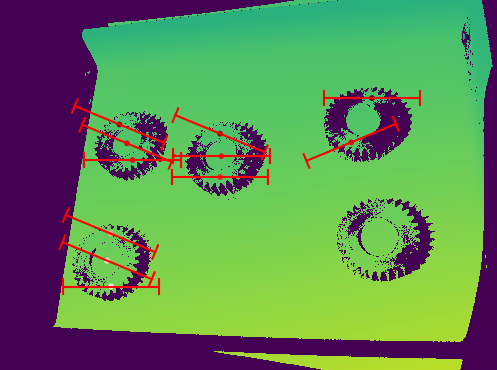}
    \caption{}
    \end{subfigure}
    \caption{Planned grasps for different objects.}
    \label{fig:grasps_part23}
\end{figure}

In order to obtain the base region where the mobile manipulator is able to grasp all the randomly placed objects, we have to find a set of object poses that nearly cover all the possible poses and plan grasps for these poses. Therefore, we randomly place objects in the tray and scan the objects in the tray, repeat this process for a couple of times, then it is assumed that these recorded object poses nearly represent all the possible objects poses. Fig. \ref{fig:grasps_part1} shows 4 different object placement, from the corresponding depth image grasp planning is performed to obtain grasps $\mathcal{G}_1$, $\mathcal{G}_2$, $\mathcal{G}_3$ and $\mathcal{G}_4$ for each of the object placement, respectively. The union of the grasps $\mathcal{G} = \mathcal{G}_1 \cup \mathcal{G}_2 \cup \mathcal{G}_3 \cup \mathcal{G}_4$ roughly represents the required grasps for grasping all the objects randomly placed in a tray, and $\mathcal{G}$ is used to calculate the base region of the tray following the method in section \MakeUppercase{\romannumeral 5}. The grasp planning examples for other objects are shown in Fig. \ref{fig:grasps_part23}.

\begin{figure}
    \centering
    \includegraphics[width = \columnwidth]{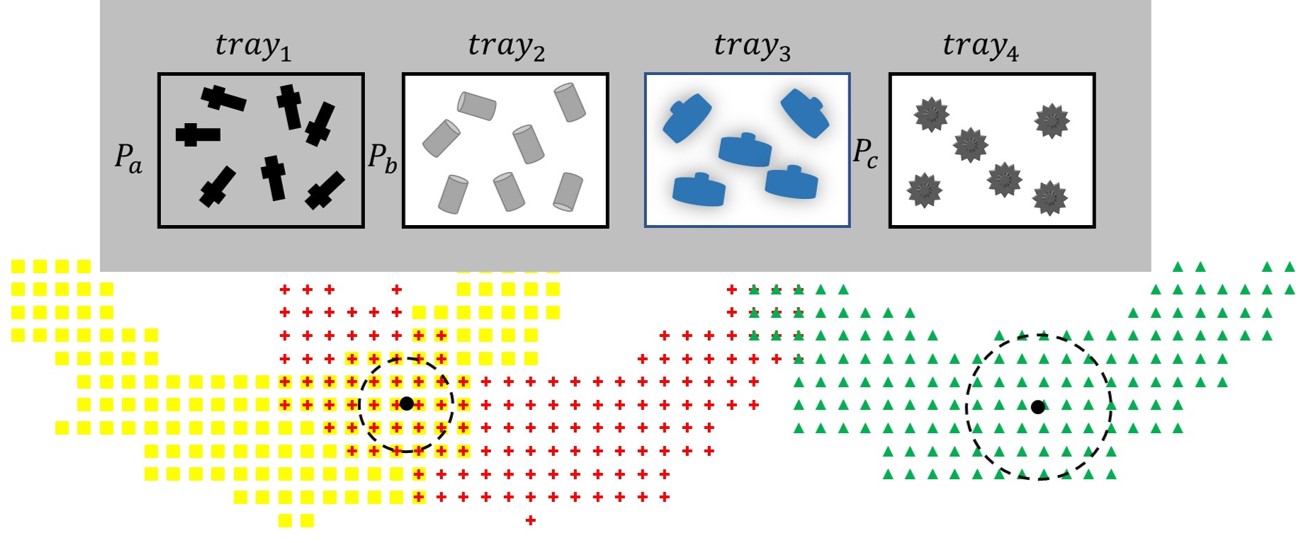}
    \caption{Schematic diagram of randomly placed objects in the trays (view from top), the dots of one color represent the base regions of the corresponding tray. The robot moves to the center of the first intersection to pick up the objects in $tray_1$ and $tray_2$, then it moves to the center of the third base region to pick up objects in $tray_4$.}
    \label{fig:base_positions_results2}
\end{figure}

\begin{figure}
    \centering
    \begin{subfigure}[h]{0.15\textwidth}
    \centering
    \includegraphics[width=\textwidth]{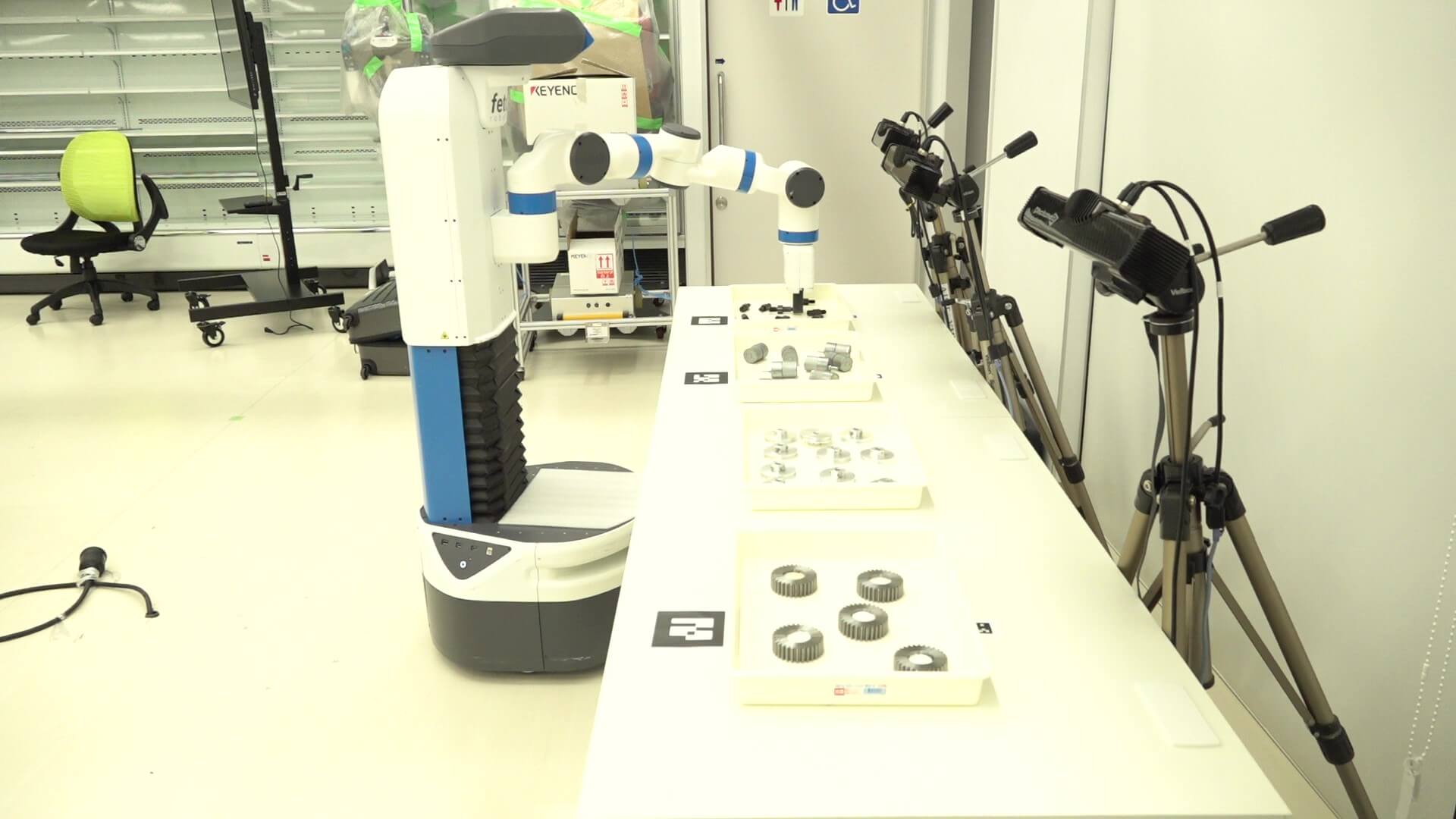}
    \end{subfigure}
    \begin{subfigure}[h]{0.15\textwidth}
    \centering
    \includegraphics[width=\textwidth]{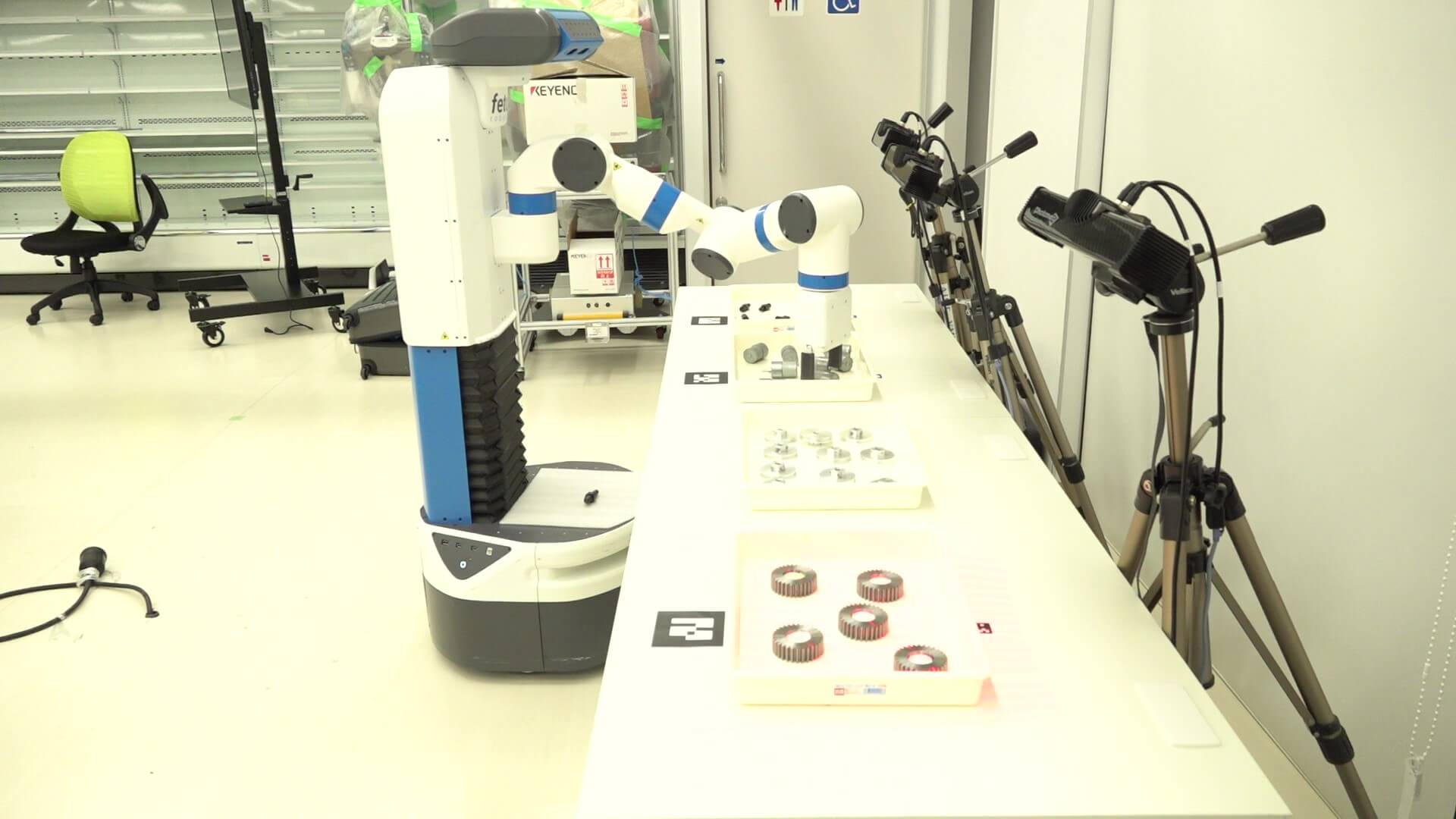}
    \end{subfigure}
    \begin{subfigure}[h]{0.15\textwidth}
    \centering
    \includegraphics[width=\textwidth]{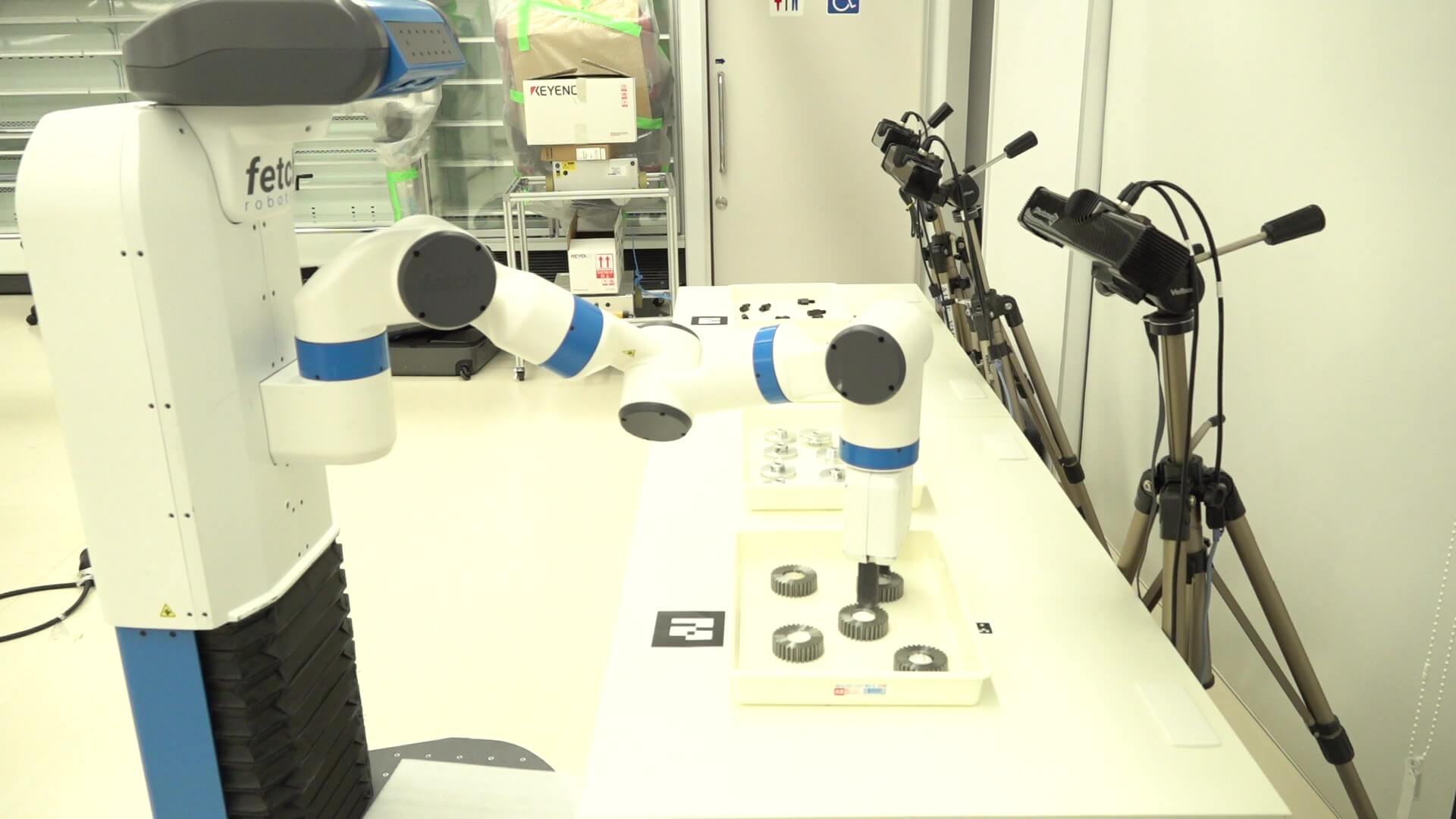}
    \end{subfigure}
    \caption{Pick up randomly placed objects following the planned globally static positions in Fig. \ref{fig:base_positions_results2}.}
    \label{fig:experiment_phoxi}
\end{figure}

During the online execution, the mobile manipulator moves to the calculated positions in the global map, the planned base positions are the black dots in Fig. \ref{fig:base_positions_results2}. Once the mobile manipulator arrives at the calculated position, its head-mounted camera detected the marker in front of the tray to get the accuracy pose of the tray relative to the robot. The online grasp planning functionality is implemented as a ROS action server, once it is requested, it returns a set of grasps in the frame of the PhoXi scanner, since the pose of PhoXi scanner is also calibrated with respect to the tray, the poses of planned grasps in the robot's base frame are derived, among all the returned grasps, the grasp with the highest graspability score will be executed. Fig. \ref{fig:experiment_phoxi} shows the grasping of randomly placed objects during the experiment.

\subsection{Regularly Placed, Dynamically Update The Base Sequence}
For regularly placed objects in the tray, it is feasible to update the base positions according to either the remaining objects or the target objects to be picked, both of the two cases are demonstrated by the experiment in this section. The experimental configuration is similar to that of the first experiment, the difference is that the base positions are updated after every round of pick-and-place task. We assume one object is picked from every target tray in every round of the task, then there are totally 4 rounds since every tray contains 4 objects. For randomly placed objects, it is impractical to determine the picking order of objects in advance, but for regularly placed objects, we can either determine which object to pick in the online phase or specify the picking order in the offline phase. In this experiment, we choose to specify the picking order before the pick-and-place tasks begin, and all the base positions in different round of tasks can also be calculated in advance. In the online execution, the base positions are updated after every round of the task.

\begin{figure}
    \centering
    \includegraphics[width=\columnwidth]{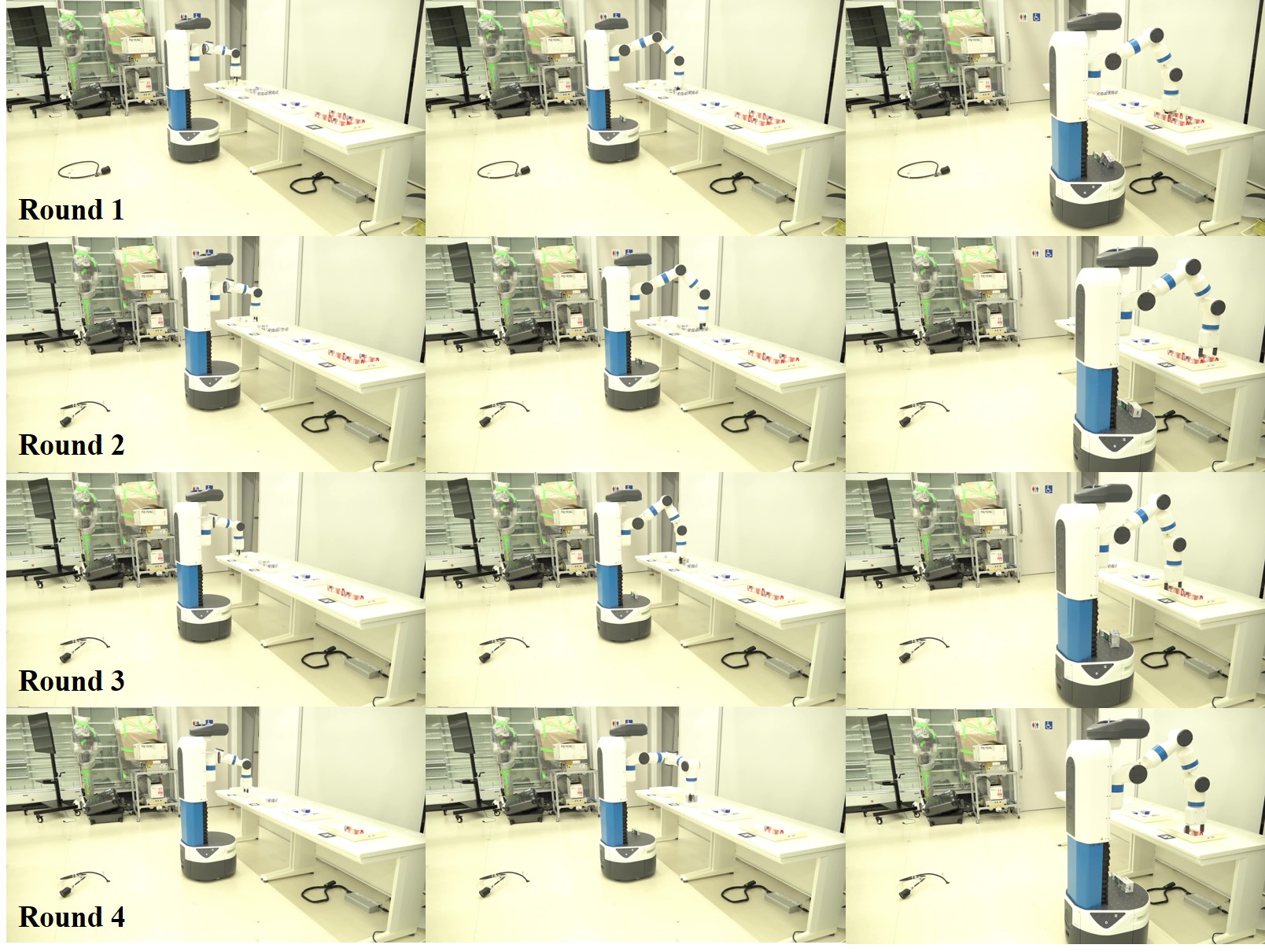}
    \caption{The base positions update based on the target objects to be picked, the update is performed after every round of pick-and-place task.}
    \label{fig:experiment_update_target}
\end{figure}

Fig. \ref{fig:experiment_update_target} shows the experiment on updating the base positions based on the target objects to be picked. The picking order is specified using a simple heuristic as shown in Fig. \ref{fig:pick_order}, referring to the same object index, that is we pick up object $\mathcal{O}_i$ in $tray_1$ and $tray_2$ and $tray_4$ in i-th round of the task. From the snapshots of the experiment, though not very obvious, the change of base positions in different rounds of task can be observed, following the dynamically updated base positions, the mobile manipulator can robustly collect the target objects in different round of the task. The experiment on updating the base positions based on the remaining objects is also successfully performed, which can be seen in the supplementary video.

\section{Conclusions}
Multiple pick-and-place tasks involved in the part-supply tasks in the assembly factory are considered in this paper. Both the efficiency and the robustness with respect to base positioning uncertainty are of major concern for practical applications. The proposed IK query method is especially helpful for finding collision-free IK solutions in complex environment. It is resolution complete for generating all the feasible base positions in a part-supply task. By incorporating the base positioning uncertainty into base sequence planning, the mobile manipulator is able to robustly finish the task even if the actual arrived position deviate from the planned position. We considered different object placement styles and discussed the feasible schemes for practical implementation. For regularly placed objects, both globally static and dynamically updated base positions are feasible, but for randomly placed objects, it is impractical and unnecessary to update the base positions. The feasible schemes are demonstrated by three sets of experiments, the first experiment shows that, following the planned base sequence, the operation time is reduced and the pick-and-place tasks are completed under the positioning uncertainty. In the second experiment, we proposed the method for estimating the base region for randomly placed objects in the tray, and demonstrate our method in more realistic settings. The third experiment illustrates that we can dynamically update the base positions for regularly placed objects based on either the remaining objects or the target objects to be picked, the overall robustness can be improved. One of the limitation observed in the experiment is that the motion of the manipulator is not optimized, which will be considered in our future work.


\bibliographystyle{ieeetr}
\bibliography{reference.bib}

\end{document}